\documentclass[twoside,compsoc]{IEEEtran}
\ifCLASSOPTIONcompsoc
  \usepackage[nocompress]{cite}
\else
  \usepackage{cite}
\fi
\ifCLASSINFOpdf
   \usepackage[pdftex]{graphicx}
\else
  \usepackage[dvips]{graphicx}
\fi
\ifCLASSOPTIONcompsoc
  \usepackage[caption=false,font=footnotesize,labelfont=sf,textfont=sf]{subfig}
\else
  \usepackage[caption=false,font=footnotesize]{subfig}
\fi
\usepackage{amsmath,amssymb,bm,booktabs,multirow,url}
\usepackage[ruled,linesnumbered]{algorithm2e}
\usepackage{hyperref}
\usepackage{makecell,xcolor}
\usepackage{ragged2e}
\usepackage{threeparttable}
\usepackage[usestackEOL]{stackengine}
\edef\tmp{\the\baselineskip}
\setstackgap{L}{\tmp}
\usepackage{wasysym}
\usepackage{utfsym}

\begin{document}
\title{On Uniform Scalar Quantization for Learned Image Compression}
\author{Haotian Zhang, Li Li, \IEEEmembership{Member, IEEE,} and Dong Liu, \IEEEmembership{Senior Member, IEEE}
\thanks{Date of current version \today. The authors are with the CAS Key Laboratory of Technology in Geo-Spatial Information Processing and Application System, University of Science and Technology of China, Hefei 230027, China (e-mail: zhanghaotian@mail.ustc.edu.cn; lil1@ustc.edu.cn; dongeliu@ustc.edu.cn). (\textit{Corresponding author: Dong Liu})}
}

\markboth{Under review}
{Zhang \MakeLowercase{\textit{et al.}}: On Uniform Scalar Quantization for Learned Image Compression}

\IEEEtitleabstractindextext{
\begin{abstract}
\justifying
Learned image compression possesses a unique challenge when incorporating non-differentiable quantization into the gradient-based training of the networks. Several quantization surrogates have been proposed to fulfill the training, but they were not systematically justified from a theoretical perspective. We fill this gap by contrasting uniform scalar quantization, the most widely used category with rounding being its simplest case, and its training surrogates. In principle, we find two factors crucial: one is the discrepancy between the surrogate and rounding, leading to train-test mismatch; the other is gradient estimation risk due to the surrogate, which consists of bias and variance of the gradient estimation. 
Our analyses and simulations imply that there is a tradeoff between the train-test mismatch and the gradient estimation risk, and the tradeoff varies across different network structures. Motivated by these analyses, we present a method based on stochastic uniform annealing, which has an adjustable temperature coefficient to control the tradeoff. Moreover, our analyses enlighten us as to two subtle tricks: one is to set an appropriate lower bound for the variance parameter of the estimated quantized latent distribution, which effectively reduces the train-test mismatch; the other is to use zero-center quantization with partial stop-gradient, which reduces the gradient estimation variance and thus stabilize the training. Our method with the tricks is verified to outperform the existing practices of quantization surrogates on a variety of representative image compression networks.

\end{abstract}
}
\maketitle
\IEEEdisplaynontitleabstractindextext
\IEEEpeerreviewmaketitle

\IEEEraisesectionheading{\section{Introduction}\label{Introduction}}
\vspace{-0.5em}
Lossy compression is a data compression technique that compresses with imperfect reconstruction quality. 
It is widely used in digital multimedia applications, such as image and video compression. 
Traditional image compression algorithms like JPEG \cite{wallace1992jpeg}, JPEG2000 \cite{skodras2001jpeg}, and BPG \cite{bpg} follow a general pipeline, where lossless entropy coders are used after transforms and quantization. Practical lossy image compression schemes rely heavily on quantization to reduce the information in images and compute discrete representations that can be transmitted digitally. For example, JPEG applies hand-crafted quantization lookup tables on coefficients obtained after transforms to reduce bits.

In contrast to traditional methods, learned image compression \cite{balle2016end} optimizes different modules in an end-to-end manner by exploiting the advantages of deep neural networks. Typically, a learned image compression scheme contains analysis transform, synthesis transform, quantizer, and entropy model. An image is transformed into a latent through analysis transform, and then the latent is quantized for digital transmission. The discrete latent is losslessly coded by the entropy model to further reduce the number of bits. Finally, the synthesis transform reconstructs an image from the discrete latent. These modules are jointly optimized to minimize the rate-distortion cost during training. 

 Quantization is a unique challenge during training in learned image compression. Since the gradient of the quantizer is zero almost everywhere, standard gradient back-propagation is inapplicable, which prevents us from optimizing the analysis transform. Although some studies have tried to forgo the quantization process \cite{agustsson2020universally,theis2022lossy}, the compression efficiency is worse than compressing with quantization.
 Therefore, incorporating quantization into the training remains a difficult issue for designing a neural image codec. 
 It is worth noting that the quantization brings troubles for training the analysis transform, but not for the entropy model and the synthesis transform. Thus, we may divide the training into two categories: joint-training and post-training. During joint-training, the analysis transform, synthesis transform, and entropy model are jointly optimized. During post-training (usually after the joint-training), the synthesis transform and entropy model are optimized with the analysis transform and quantizer being fixed. This two-stage training strategy was presented in \cite{guo2021soft}.

Uniform scalar quantization is the most widely used category in learned image compression, with rounding being its simplest case. To enable joint-training, many surrogates have been proposed to replace rounding. 
Training with adding uniform noise (AUN) \cite{balle2016end} is a popular approach to approximate rounding, which allows gradient propagation but introduces train-test mismatch. Despite AUN, some studies proposed to use stochastic rounding \cite{theis2017lossy} with straight-through estimator (STE) \cite{bengio2013estimating}, which considers rounding as an identity function during gradient propagation. STE performs well as a workaround, but in principle, it definitely introduces gradient estimation errors for training the analysis transform. In \cite{minnen2020channel}, a mixed quantization surrogate (MIX) is empirically proposed, using AUN and rounding with STE, respectively, for rate estimation and reconstruction. An empirical comparison of several surrogates is conducted in \cite{tsubota2023comprehensive}, which suggests that the best quantization surrogate varies across different architectures. Though encouraging progress has been made, there is still little insight into how these quantization surrogates work. The lack of a deeper theoretical understanding may prevent us from designing a better quantization surrogate and obtaining more effective training.  

In this paper, we introduce systematic theoretical analyses of the existing surrogates for rounding. Our analyses focus on two crucial factors: train-test mismatch and gradient estimation risk\footnote{We use the term "risk" from the statistical learning literature to denote the "expected error" of gradient estimation. Risk usually consists of bias and variance. Training a neural network is highly stochastic, which enables us to define bias and variance, in Sec. \ref{sec:gradient}.}, which correspond to the forward calculation and backward calculation across the quantization surrogate, respectively. The backward calculation estimates the gradient of the loss obtained by the forward calculation. We study them separately since one forward calculation is not limited to a single backward calculation method. The train-test mismatch is caused by the discrepancy between the forward calculation and rounding, resulting in a mismatch in the loss function. In our study, we attribute this mismatch to the rate and distortion estimation errors (compared to rounding) of different forward calculations. For backward calculation, we study the gradient estimation risk, which consists of bias and variance of gradient estimation. Unbiasedness is usually preferred as it ensures the convergence of stochastic optimization procedures \cite{mohamed2020monte}, and a low variance gradient promotes the efficiency of optimization. In addition to exploring the train-test mismatch and gradient estimation risk caused by different surrogates, we also explore how they hurt performance through carefully designed experiments. 

\begin{table}[t]
\centering
\begin{threeparttable}    
\tabcolsep=8.5pt
\caption{Abbreviations of different quantization surrogates\tnote{1}.}
\begin{tabular}{ c|c } 
    \hline
    AUN& Additive uniform noise \cite{balle2016end}. \\
    STE& Straight-through estimator \cite{bengio2013estimating}. \\
    MIX& Mixed quantization strategy \cite{minnen2020channel}. \\
    SGA & Stochastic Gumbel annealing \cite{yang2020improving}. \\
    STH & Soft-then-hard \cite{guo2021soft}. \\
    SHA & Soft-to-hard annealing \cite{agustsson2017soft}. \\
    UQ & Universal quantization \cite{ziv1985universal}. \\
    UQ-s & UQ with same noise \cite{choi2019variable}. \\
    UQ-i & UQ with independent noise.  \\    
    SUA & Stochastic uniform annealing \cite{agustsson2020universally}. \\
    SUA-n & SUA without denoising function. \\
    SR & Stochastic rounding. \\
    SRA & Stochastic rounding annealing. \\
    PGE& Pathwise gradient estimator \cite{kingma2013auto}. \\
    \hline
\end{tabular}
\begin{tablenotes}
    \footnotesize
    \item[1] Please refer to Sec. \ref{para:sth} and Sec. \ref{sec:quantization} for definitions of these methods.
\end{tablenotes}
\end{threeparttable}
\label{tab:notation}
\vspace{-0.8em}
\end{table}  

Our analyses imply that there is a tradeoff between the train-test mismatch and the gradient estimation risk, and the tradeoff varies across different network structures. Moreover, our analyses show that annealing-based methods enable the control of the tradeoff by adjusting the temperature coefficient. Motivated by this, we present a stochastic uniform annealing-based method to control the tradeoff. Additionally, the tolerance to the bias and variance of gradient estimation may differ across different network structures, as well. Therefore, we offer a biased gradient estimator with lower variance and an unbiased one with higher variance to be chosen for the backward calculation.

Moreover, our analyses enlighten us as to two subtle tricks. One is the lower bound for the variance parameter of the estimated quantized latent distribution.
Our analyses show that the train-test mismatch in rate is huge when the estimated variance is small, for almost all of the studied surrogates. Therefore, setting an appropriate lower bound to avoid such cases can reduce the train-test mismatch, but may lead to inaccurate latent distribution estimation. We then propose setting a relatively larger bound to effectively reduce train-test mismatch during joint-training, but setting a tiny bound to achieve accurate distribution estimation during post-training. 
The other is zero-center quantization, which applies the estimated mean as quantization offsets. Applying zero-center quantization to the MIX \cite{minnen2020channel} method helps decrease train-test mismatch, resulting in better performance.
When utilizing annealing-based methods with zero-center quantization, the gradient of the estimated mean is affected by high variance. To reduce the gradient estimation variance and stabilize training, we propose a partial stop-gradient method for the estimated mean in zero-center quantization. 

Our main contributions can be summarized as follows:
\begin{itemize}
    \item We perform systematic theoretical analyses of the existing quantization surrogates, focusing on two crucial factors: train-test mismatch and gradient estimation risk.
    Our analyses imply that there is a tradeoff between the train-test mismatch and the gradient estimation risk, which varies across different network structures.
    \item We present a stochastic uniform annealing-based method to control the tradeoff by adjusting the temperature coefficient.
    \item We study the effect of the lower bound for the estimated variance parameter and propose setting appropriate bounds during joint-training and post-training respectively.
    \item We present a partial stop-gradient method to reduce gradient estimation variance and stabilize training for zero-center quantization.
\end{itemize}

\vspace{-1em}
\section{Related Work}
\vspace{-0.2em}
\subsection{Learned Image Compression}
Learned image compression has made significant progress and demonstrated impressive performance. In the early years, some studies addressed learned image compression in terms of distortion and rate separately \cite{toderici2015variable, toderici2017full,johnston2018improved}. Balle \textit{et al.} \cite{balle2016end} formulated learned image compression as a joint rate-distortion optimization problem. The scheme in \cite{balle2016end} contains four modules: analysis transform, synthesis transform, quantizer, and entropy model. Most recent studies follow this joint rate-distortion optimization. 

The entropy model estimates the distribution of quantized latent. A factorized prior is employed in \cite{balle2016end}. Balle \textit{et al.} \cite{balle2018variational} proposed a hyperprior model that parameterizes the distribution as a zero-mean Gaussian model conditioned on side information. Minnen \textit{et al.} \cite{minnen2018joint} extended \cite{balle2018variational} to a mean-scale Gaussian model, and proposed a more accurate entropy model, jointly utilizing an autoregressive context model and hyperprior. Cheng \textit{et al.} \cite{cheng2020learned} extended \cite{minnen2018joint} by introducing a Gaussian mixture model. In order to balance compression performance and running efficiency, He \textit{et al.} \cite{He_2021_CVPR} proposed a checkerboard context model for parallel computation. Minnen and Singh \cite{minnen2020channel} proposed a channel-wise autoregressive context model. Qian \textit{et al.} \cite{qian2022entroformer} proposed a transformer-based context model to capture long-range dependencies. Some recent studies \cite{lee2018context,chen2021end,guo2021causal,hu2021learning,he2022elic} also focus on improving the efficiency of entropy models.

In addition to the design of entropy models, the architecture of the neural networks in transforms is also important. Recurrent neural networks are used in \cite{toderici2015variable,toderici2017full,johnston2018improved, lin2020spatial}. Balle \textit{et al.} \cite{balle2016end} proposed a convolution neural network-based image compression model. Chen \textit{et al.} \cite{chen2021end} introduced a non-local attention module to capture global receptive field. Cheng \textit{et al.} \cite{cheng2020learned} simplified it by removing the non-local block. 
Attention mechanism is also investigated in \cite{guo2021causal,zou2022devil}. Some studies \cite{zhu2022transformer,zou2022devil} tried to construct transformer-based transforms for image compression. In addition to these non-invertible transforms, several studies utilized invertible ones. Ma \textit{et al.} \cite{ma2020end} proposed a trained wavelet-like transform. Helminger \textit{et al.} \cite{helminger2020lossy} tried to leverage normalizing flows for learned image compression. Xie \textit{et al.} \cite{xie2021enhanced} combined non-invertible and invertible networks.
\vspace{-0.6em}
\subsection{Quantization in Learned Image Compression}
The quantization methods in the field of learned image compression can be grouped into two categories: scalar quantization and vector quantization. Up to now, the scalar quantizers can be roughly grouped into three categories: uniform, non-uniform, and binary quantization. Uniform scalar quantization is the most widely adopted method in learned image compression, with rounding being its simplest case.
\vspace{-0.6em}
\subsubsection{Uniform Scalar Quantization}
Since the gradient of rounding is zero almost everywhere, the standard back-propagation is inapplicable. To enable end-to-end optimization, lots of quantization surrogates are proposed. Training with additive uniform noise (AUN) proposed in \cite{balle2016end} is a widely used approach for approximating rounding. 
Theis \textit{et al.} \cite{theis2017lossy} proposed to use straight-through estimator (STE) \cite{bengio2013estimating} during training, which applies stochastic rounding in the forward pass, but uses modified gradient in the backward pass. Choi \textit{et al.} \cite{choi2019variable} employed universal quantization as another alternative for AUN and achieved better performance.

\label{para:sth} 
Training with AUN introduces stochasticity during training, leading to the train-test mismatch. Stochastic Gumbel annealing (SGA) is proposed in \cite{yang2020improving} to close the discretization gap through iterative inference. Agustsson and Theis \cite{agustsson2020universally} also applied an annealing-based method during training. 
Sun \textit{et al.} \cite{10008823} exploited the gradient scaling method to take quantization error into consideration. Guo \textit{et al.} \cite{guo2021soft} analyzed the train-test mismatch of different quantization surrogates, and presented a two-stage soft-then-hard (STH) training strategy to eliminate the train-test mismatch. This two-stage training strategy firstly trains with AUN, then fine-tunes the synthesis transform and entropy model with rounded latent.

\label{para:mix} 
Usually, the quantization surrogate is the same for reconstruction and rate estimation. But they can be different. Minnen and Singh \cite{minnen2020channel} empirically proposed a mixed quantization surrogate. The mixed surrogate (MIX) uses noisy latent for rate estimation but uses rounded latent and STE when passing a synthesis transform. This mixed surrogate outperforms AUN and is widely adopted in recent studies.
Tsubota \textit{et al.} \cite{tsubota2023comprehensive} empirically compared the combinations of several surrogates for a decoder and an entropy model. The results show that using noise-based methods for the entropy model and rounding-based methods for the decoder is better.

\vspace{-0.6em}
\subsubsection{Other Quantization Methods}
Non-uniform scalar quantization uses a non-uniform quantization interval. 
Cai \textit{et al.} \cite{cai2018deep} alternatively optimized the quantization intervals and the transforms. 
Some studies \cite{mentzer2018conditional,zhong2020channel} 
learned quantization intervals and transforms jointly based on clustering.

Some of the earliest studies \cite{toderici2015variable,toderici2017full} in the field of learned image compression used binary scalar quantization. During training, stochastic binarization is applied in the forward pass, and STE is used in the backward pass.

In addition to scalar quantization, vector quantization is also explored. Agustsson \textit{et al.} \cite{agustsson2017soft} proposed a soft relation of vector quantization
for end-to-end optimization. A probabilistic vector quantizer is proposed in \cite{zhu2022unified} to estimate the prior parameters. Feng \textit{et al.} \cite{feng2023nvtc} proposed nonlinear vector transform and applied entropy-constrained vector quantization. Zhang \textit{et al.} \cite{zhang2023lvqac} proposed a lattice vector quantization scheme.
\vspace{-0.5em}
\subsubsection{Compression without Quantization}
One way to avoid quantization is using a stochastic encoder \cite{agustsson2020universally} that sends continuous but stochastic samples with finite rate \cite{theis2022algorithms}. This approach is under active research. Till now, it has been reported that the efficiency of the stochastic encoder is worse than that of the deterministic one.

\vspace{-0.7em}
\section{Preliminaries}\label{Related_work}
\vspace{-0.1em}
\subsection{Lossy Compression}
Rate-distortion (R-D) theory \cite{shannon1948mathematical} established a foundation for lossy compression. Let us define $X \sim p_X(x),~x\in \mathcal{X}$, where $X$ is a stochastic data source, $\mathcal{X}$ is the source alphabet, and $x$ is a source symbol. The reconstruction $\hat{X}\sim p_{\hat{X}}(\hat{x}),~\hat{x}\in \hat{\mathcal{X}}$, is described by a conditional distribution $p_{\hat{X}|X}$. R-D theory defines a distortion measure that quantifies the difference between pairs of original data samples and reconstructions. Given a distortion function $d:\mathcal{X}\times\mathcal{\hat{X}}\rightarrow\mathcal{R}^{+}$, the average distortion is $D=\mathbb{E}_{p_{X,\hat{X}}}[d(x,\hat{x})]$. The theory also defines $I(X;\hat{X})$, the mutual information, that quantifies the amount of information needed to obtain reconstruction. 

For a given distortion threshold $D$, the minimum of rates $R$ is characterized by the rate distortion function \cite{shannon1959coding},
\vspace{-0.3em}
\begin{equation}\label{eq_rdfunction}
\begin{aligned}
    R(D) = \min_{p_{\hat{X}|X}:\mathbb{E}_{p_{\hat{X},X} }[d(x,\hat{x})]\leq D}I(X;\hat{X}).
\end{aligned}
\vspace{-0.6em}
\end{equation}
It can be relaxed to an unconstrained one, by introducing Lagrangian multiplier \cite{blahut1972computation},
\vspace{-0.3em}
\begin{equation}\label{eq_lagrangian}
\begin{aligned}
    L = I(X;\hat{X}) + \lambda\mathbb{E}_{p_{X,\hat{X}}}[d(x,\hat{x})],
\end{aligned}
\vspace{-0.6em}
\end{equation}
where $\lambda$ controls the tradeoff between rate and distortion.
\vspace{-0.8em}
\subsection{Learned Lossy Image Compression}
The scheme of learned image compression \cite{balle2020nonlinear,balle2016end} 
can be represented as $X\rightarrow Y \rightarrow\hat{Y}\rightarrow\hat{X}$.
The sender applies an analysis transform to an image $x$, generating latent representation $y=g_a(x|\phi)$. Rounding is then applied to obtain a discrete latent $\hat{y}=Q(y)$. The discretized latent can be losslessly coded under an entropy model $q_{\hat{Y}}(\hat{y}|\psi)$. Finally, the receiver recovers $\hat{y}$, and generates a reconstruction $\hat{x}=g_s(\hat{y}|\theta)$ through a synthesis transform. The notations $\phi$, $\theta$, and $\psi$ are trainable parameters of the analysis transform, synthesis transform, and entropy model, respectively.

The transforms in learned lossy image compression are typically deterministic but may not be invertible, resulting in some information loss when passing through the synthesis transform. However, the information rate transmitted from sender to receiver is not affected by the synthesis transform. The information rate is $I(X;\hat{Y})=I(Y;\hat{Y})$. If the quantizer is deterministic (e.g. rounding), then we have
\vspace{-0.3em}
\begin{equation}\label{eq_infor}
\begin{aligned}
   I(X;\hat{Y})=I(Y;\hat{Y})=H(\hat{Y}).
\end{aligned}
\vspace{-0.3em}
\end{equation}
In practice, we do not know the actual distribution $p_{\hat{Y}}$ but need to estimate an entropy model $q_{\hat{Y}}$ to losslessly entropy code $\hat{y}$. The average rate required is the cross-entropy between $p_{\hat{Y}}$ and $q_{\hat{Y}}$,
\vspace{-0.3em}
\begin{equation}\label{eq_bitrate}
\begin{aligned}
  R = H(p_{\hat{Y}},q_{\hat{Y}})=\mathbb{E}_{p_{\hat{Y}}}[-\log_{2}(q_{\hat{Y}}(\hat{y}))]. 
\end{aligned}
\vspace{-0.3em}
\end{equation}
$H(p_{\hat{Y}},q_{\hat{Y}})$ is always larger than $H(\hat{Y})$ unless $p_{\hat{Y}}= q_{\hat{Y}}$.

During training, the parameters $\phi$, $\theta$, and $\psi$ are jointly optimized with the loss function 
\vspace{-0.3em}
\begin{equation}\label{eq_loss}
\begin{aligned}
   L=& \mathbb{E}_{p_X}[-\log_{2}(q_{\hat{Y}}(\hat{y}|\psi))+\lambda d(x,\hat{x})],
\end{aligned}
\vspace{-0.3em}
\end{equation}
where $\hat{y}=Q(g_a(x|\phi))$, and $\hat{x}=g_s(\hat{y}|\theta)$. 

\vspace{-0.5em}
\subsection{Quantization in Training}
The loss function described in Eq. (\ref{eq_loss}) can be split into three parts, respectively for synthesis transform, entropy model, and analysis transform,
\vspace{-0.1em}
\begin{align}
   & \min_{\theta} \lambda \mathbb{E}_{p_{X}}[d(g_s(\hat{y}|\theta),x)],\label{eq_synthesis} \\ 
   & \min_{\psi} \mathbb{E}_{p_{X}}[-\log(q_{\hat{Y}}(\hat{y}|\psi))], \label{eq_entropy} \\ 
    \min_{\phi} \mathbb{E}_{p_{X}}&[-\log(q_{\hat{Y}}(\hat{y}|\psi))+\lambda d(g_s(\hat{y}|\theta),x)]. \label{eq_analysis}
    \vspace{-0.3em}
\end{align}
The loss Eq. (\ref{eq_analysis}) is not appropriate for end-to-end training with gradient descent since the derivative of rounding is almost zero everywhere. This prevents us from optimizing the analysis transform $g_a(\cdot|\phi)$ with a true quantization.

In this paper, we divide the training stage into two phases: joint-training and post-training. During \textbf{joint-training}, analysis transform, synthesis transform, and entropy model are optimized jointly, which requires a differentiable quantizer.
During \textbf{post-training}, only synthesis transform and entropy model are optimized with analysis transform and quantizer fixed.
During joint-training, surrogates are needed to replace rounding, which will be thoroughly introduced in Sec. \ref{sec:quantization}.
For post-training, we extend the soft-then-hard (STH) \cite{guo2021soft} idea to a general case, using rounded latent to fine-tune synthesis transform and entropy model after joint training. Post-training can always be applied to reduce the effect of train-test mismatch after joint-training.

\vspace{-0.3em}
\section{Analyses}
\label{sec:analysis}
In this section, we systematically analyze the train-test mismatch and gradient estimation risk of various quantization surrogates. First, we introduce and extend various quantization surrogates for joint-training in Sec. \ref{sec:quantization}. Second, we separately study the train-test mismatch in rate and distortion in Sec. \ref{sec:distortion} and \ref{sec:rate}, respectively, since the forward calculation for these two terms can be different.
Third, we analyze the gradient estimation risk in the backward calculation in Sec. \ref{sec:gradient}. Fourth, we show the tradeoff between the train-test mismatch and the gradient estimation risk, and how they hurt the compression performance in Sec. \ref{sec:tradeoff}. Finally, we analyze two subtle tricks that are closely related to train-test mismatch in Sec. \ref{sec:subtle}.
\subsection{Learned Quantization for Joint-Training}
\label{sec:quantization}
Quantization surrogate consists of two parts: forward calculation and backward calculation. We introduce them separately since the forward calculation is not limited
to a single backward calculation method. 

\subsubsection{Forward Calculations}
Forward calculations can be grouped into two categories: deterministic calculations and stochastic calculations. The deterministic calculations can be further grouped into differential and non-differential operators. To distinguish with the actual quantized latent $\hat{y}$, we use $\tilde{y}$ to represent the latent obtained by the forward calculation of surrogate during joint-training, and $y_i$ is one component in latent $y$. 

Firstly, we introduce the deterministic forward calculations:

\textbf{Soft-to-hard annealing (SHA).} \label{para:sha} 
Agustsson \textit{et al.} \cite{agustsson2017soft} proposed soft-to-hard annealing vector quantization during training. By changing the value of a temperature coefficient, the differentiable function goes towards hard quantization gradually. The vector quantization version can be reduced to a scalar version, which replaces rounding with a soft function. The shape of soft function $s_{\alpha}(\cdot)$ is controlled by its parameter $\alpha$,
\begin{equation}\label{eq_soft}
\begin{aligned}
       \tilde{y_i}=s_{\alpha}(y_i)=\lfloor y_i\rfloor+\frac{\tanh(\alpha r_i)}{2\tanh(\alpha/2)}+0.5,r_i=y_i-\lfloor y_i\rfloor-0.5.
\end{aligned}
\end{equation}
The soft function is differentiable everywhere \cite{agustsson2020universally}, which allows gradient propagation. However, SHA does not introduce any information loss because it is invertible.

\textbf{Rounding.}
Directly using rounding in the forward pass is adopted in \cite{lee2018context,minnen2020channel}. It is clear that rounding is not differentiable, which requires defining a gradient in the backward pass.

Then, we introduce the stochastic forward calculations:

\textbf{Additive uniform noise (AUN).} \label{para:AUN}
Balle \textit{et al.} \cite{balle2016end} proposed to replace rounding with additive uniform noise for joint-training,
\begin{equation}\label{eq_addnoise}
\begin{aligned}
       \tilde{y_i}=y_i+u_i,~u_i\sim \mathcal{U}(-0.5,0.5).
\end{aligned}
\end{equation}
They show the statistical characteristics of AUN and rounding are similar.
After taking expectations, we get the surrogate training loss,
\begin{equation}\label{eq_noiseeploss}
\begin{aligned}
       & \mathbb{E}_{p_{X},p_{U}}[-\log_{2}(q_{\tilde{Y}}(y+u))+\lambda d(x,g_s(y+u))],
\end{aligned}
\end{equation}
which is differentiable to all components, and can be simply estimated by Monte-Carlo sampling.

\textbf{Universal quantization (UQ).} UQ \cite{ziv1985universal} has two variants. The first, \textbf{UQ-s}, adds the same uniform noise for all components of the latent,
\vspace{-0.2em}
\begin{equation}\label{eq_addnoise2}
\begin{aligned}
       \tilde{y_i}=\lfloor y_i+u \rceil-u,~u\sim \mathcal{U}(-0.5,0.5).
\end{aligned}
\vspace{-0.2em}
\end{equation}
UQ-s was used in \cite{choi2019variable} as an alternative for AUN.
The second, \textbf{UQ-i}, adds independent uniform noise for each component,
\vspace{-0.2em}
\begin{equation}\label{eq_addnoise3}
\begin{aligned}
       \tilde{y_i}=\lfloor y_i+u_i \rceil-u_i,~u_i\sim \mathcal{U}(-0.5,0.5).
\end{aligned}
\vspace{-0.2em}
\end{equation}

\textbf{Stochastic Gumbel annealing (SGA).} \label{para:sga} 
Training with AUN leads to train-test mismatch. Stochastic Gumbel annealing is proposed in \cite{yang2020improving} to close the discretization gap through iterative inference. SGA can be written as $\tilde{y_i}=\lfloor y_i\rfloor+\delta_i$, where
\begin{equation}\label{eq_sga}
\begin{aligned}
       & p(\delta_i=0)\propto \exp(-\tanh^{-1} (y_i-\lfloor y_i\rceil)/\tau),\\
       & p(\delta_i=1)\propto \exp(-\tanh^{-1} (\lfloor y_i\rceil-y_i)/\tau),\\
\end{aligned}
\end{equation}
By decreasing the value of a temperature coefficient $\tau$, the distribution goes towards deterministic gradually. 
Yang \textit{et al.} \cite{yang2020improving} applies the Gumbel-Softmax trick \cite{jang2016categorical} to relax the problem to a continuous one. They take the expectation of $\tilde{y}$ whose probability is drawn from a Gumbel distribution and followed by softmax.

\textbf{Stochastic uniform annealing (SUA).} \label{para:sua} 
Agustsson and Theis \cite{agustsson2020universally} tried to implement adding uniform noise at the inference stage to eliminate the train-test mismatch issue. They show a way to smoothly interpolate between uniform noise and rounding while maintaining differentiability through soft simulation, which is similar to SGA but sampling from uniform distribution instead of Gumbel distribution,
\begin{equation}\label{eq_univesalsoft}
\begin{aligned}
       & \tilde{y_i}=r_{\alpha}(s_{\alpha}(y_i)+u_i),\\
       & r_{\alpha}(z)=s_{\alpha}^{-1}(z-0.5)+0.5,\\
\end{aligned}
\end{equation}
where $s_{\alpha}$ is described in Eq. (\ref{eq_soft}) and $r_{\alpha}$ \cite{agustsson2020universally} is used for denoising. 
$s_{\alpha}$ is seen as a part of analysis transform, and $r_{\alpha}$ is seen as a part of synthesis transform in \cite{agustsson2020universally}. 
The loss is
\begin{equation}\label{eq_losssoft}
\begin{aligned}
       & \mathbb{E}_{p_{X},p_{U}}[-\log_{2}(q_{\tilde{Y}}(s_{\alpha}(y)+u))+\lambda d(x,g_s(\tilde{y}))].
\end{aligned}
\end{equation}
 At the inference stage, $s_{\alpha}(y)+u$ is transmitted to the decoder side through a shared noise source \cite{agustsson2020universally}.
In this paper, we denote the forward calculation $\tilde{y}_i=r_{\alpha}(s_{\alpha}(y_i)+u_i)$ as \textbf{SUA}, and denote $\tilde{y}_i=s_{\alpha}(y_i)+u_i$ (without the denoising function) as \textbf{SUA-n}. Note that we analyze SUA and SUA-n as if they are quantization surrogates for rounding. Different from \cite{agustsson2020universally}, in this paper, we always use the deterministic quantizer, rounding, at the inference stage instead of transmitting a random sample.

\textbf{Stochastic rounding (SR).} \label{para:sr} Theis \textit{et al.} \cite{theis2017lossy} applied stochastic rounding in the forward process, which outputs discrete value but maintains stochasticity.
\begin{equation}\label{eq_univesalsoft2}
\begin{aligned}
       & P(\tilde{y_i}=\lfloor y_i \rfloor)= \lceil y_i \rceil -y_i ,\\
       & P(\tilde{y_i}=\lceil y_i \rceil)= y_i-\lfloor y_i \rfloor.\\
\end{aligned}
\end{equation}
The loss is $\mathbb{E}_{p_{X},p_{\tilde{Y}}}[-\log_{2}(q_{\tilde{Y}}(\tilde{y}))+\lambda d(x,g_s(\tilde{y}))]$.
\textbf{Stochastic rounding annealing (SRA)} \label{para:sra} can be achieved with a soft function. $P(\tilde{y_i}=\lfloor y_i \rfloor)= \lceil s_{\alpha}(y_i) \rceil -s_{\alpha}(y_i)$. As $\alpha$ increases, SRA goes towards rounding.

\subsubsection{Backward Calculations}
In this section, we study four types of gradient estimators for backward calculation: standard gradient, unbiased pathwise gradient estimator, straight-through estimator, and direct expected gradient.

\textbf{Standard gradient.} For a deterministic and differentiable forward function, such as SHA, the gradient can be computed via standard back-propagation. 

\textbf{Unbiased pathwise gradient estimator (PGE).} \label{para:pge}
For forward calculations based on reparameterization, such as AUN, SUA, and SGA, the gradient of the expected loss can be estimated through an unbiased pathwise gradient estimator \cite{kingma2013auto}. Taking AUN as an example, the gradient can be estimated by
\vspace{-0.2em}
\begin{equation}\label{eq_univesalsoft3}
\begin{aligned}
\frac{\partial \mathbb{E}_{p_U}[L(\tilde{y})]}{\partial y}=\mathbb{E}_{p_U}[\frac{\partial L(\tilde{y})}{\partial \tilde{y}}\frac{\partial \tilde{y}}{\partial y}].
\end{aligned}
\vspace{-0.2em}
\end{equation}

\textbf{Straight-through estimator (STE).} \label{para:ste} The key idea of straight-through estimator \cite{bengio2013estimating} is to back-propagate through the hard function as if it had been the identity function. Taking rounding as an example, the gradient propagated to $y$ is 
\begin{equation}\label{eq_ste}
\begin{aligned}
       & \frac{\partial L(\lfloor y\rceil)}{\partial y}=\frac{\partial L(\lfloor y\rceil)}{\partial \lfloor y\rceil}\frac{\partial \lfloor y\rceil}{y}\approx\frac{\partial L(\lfloor y\rceil)}{\partial \lfloor y\rceil}.
\end{aligned}
\end{equation}
Straight-through estimator is clearly a biased pathwise gradient estimator \cite{bengio2013estimating}. We generalize the idea of STE to a broader context, to back-propagate through any function as if it had been the identity function. The generalized STE can be applied even if $\partial\tilde{y}/\partial y$ exists.

\begin{figure*}[t]
  \centering
\includegraphics[width=0.99\linewidth]{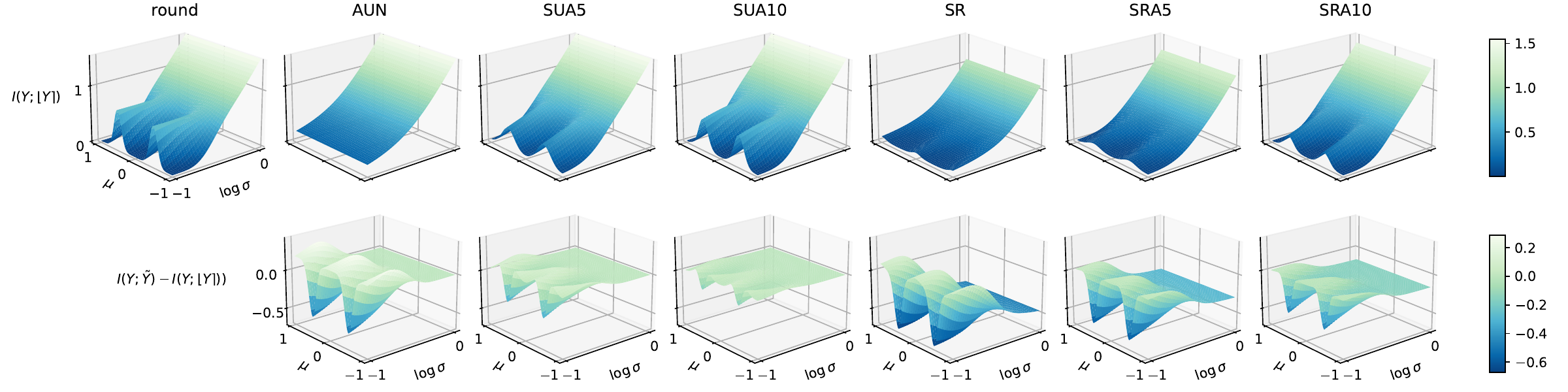}
\vspace{-0.3em}
\caption{
\label{fig:mutual_info}
Visualization of $I(Y;\tilde{Y})$ and $I(Y;\tilde{Y})-I(Y;\lfloor Y\rceil)$ with respect to different forward calculations. $Y\sim \mathcal{N}(\mu,\sigma^2)$. SUA5 stands for SUA with a temperature coefficient of 5, similar to SUA10, SRA5, etc.}
\vspace{-0.3em}
\end{figure*}

\textbf{Direct expected gradient (EP).} We can directly calculate the gradient of the expected loss function if it exists. For these stochastic discrete forward calculations, the REINFORCE algorithm \cite{williams1992simple} provides a powerful gradient estimator. The core idea is that even though $L(\tilde{y})$ is a step function with useless derivatives, the expected cost $\mathbb{E}_{p_{\tilde{Y}}}[L(\tilde{y})]$ is usually a smooth function amenable to gradient descent. Taking stochastic rounding as an example, the expected gradient is 
\vspace{-0.2em}
\begin{equation}\label{eq_ste2}
\begin{aligned}
       &\frac{\partial \mathbb{E}_{p_{\tilde{Y}}} L(\tilde{y})}{\partial y}= \frac{\partial \sum_{\tilde{y}} L(\tilde{y})p_{\tilde{Y}}(\tilde{y})}{\partial y}=\sum_{\tilde{y}} L(\tilde{y}) \frac{\partial p_{\tilde{Y}}(\tilde{y})}{\partial y}.
\end{aligned}
\vspace{-0.2em}
\end{equation}
It is easy to calculate if $y$ is a scalar. Agustsson and Theis \cite{agustsson2020universally} presented a continuous scalar version expected derivative for AUN,
\begin{equation}\label{eq_ste3}
\begin{aligned}
       &\frac{\partial \mathbb{E}_{p_U} L(\tilde{y})}{\partial y}= \frac{\partial \int_{u} L(y+u)du}{\partial y}=L(y+0.5)-L(y-0.5),
\end{aligned}
\end{equation}
where $y$ is a scalar. The expected gradient of a scalar function \cite{agustsson2020universally} is easy to apply, while the calculation for a vector function suffers from high computation complexity. That is to say, EP is seldom useful for practical learned compression.

\vspace{-0.5em}
\subsection{ Distortion Estimation}
\label{sec:distortion}
The estimated distortion of forward calculation is $D(\tilde{Y})=\mathbb{E}_{p_{X}}[d(x,\tilde{x})]$, where $\tilde{x}=g_s(\tilde{y})$. As described in Eq. (\ref{eq_rdfunction}) and (\ref{eq_infor}), the information $I(Y;\tilde{Y})$ determines the lower limit of distortion. Therefore, we firstly theoretically compare $I(Y;\tilde{Y})$ of different forward calculations in Sec. \ref{sec_distor:mutual}. Furthermore, in Sec. \ref{sec_distor:gaussian}, we simulate with toy sources to estimate distortion through training a neural network. Since the latent is usually assumed to be a Gaussian variable in learned image compression, we use Gaussian variables in our simulations.
\subsubsection{Analyses of Mutual Information}
\label{sec_distor:mutual}
We set that $Y\sim \mathcal{N}(\mu,\sigma^2)$. For SHA, it does not cause any information loss as the soft function is invertible.
In Fig. \ref{fig:mutual_info}, we visualize $I(Y;\tilde{Y})$ and $I(Y;\tilde{Y})-I(Y;\lfloor Y\rceil)$ of AUN, SR, SUA, and SRA, because for these cases the mutual information is easier to calculate. 
 For AUN, $I(Y;\tilde{Y})-I(Y;\lfloor Y\rceil)$ is negligible when $\sigma$ is relatively large, while it is substantial and highly dependent on $\mu$ when $\sigma$ is small.
In the case of SR, the difference compared to rounding is significant even when $\sigma$ is large. By incorporating annealing techniques for SUA and SRA, $I(Y;\tilde{Y})-I(Y;\lfloor Y\rceil)$ gradually diminishes as the parameter $\alpha$ increases.

\begin{figure*}[ht]
  \centering
    \subfloat[Distortion estimation error (relative), i.e., $(D(\tilde{Y})-D(\lfloor Y \rceil))/D(\lfloor Y \rceil)$, of various forward calculations with respect to different latent distributions. $X\sim \mathcal{N}(0,1),~Y=\sigma X+\mu,~Y\sim \mathcal{N}(\mu,\sigma^2),~D(\tilde{Y})=\mathbb{E}_{p_X}d(x,g_s(\tilde{y}))$.]
    {  
        \includegraphics[width=\linewidth]{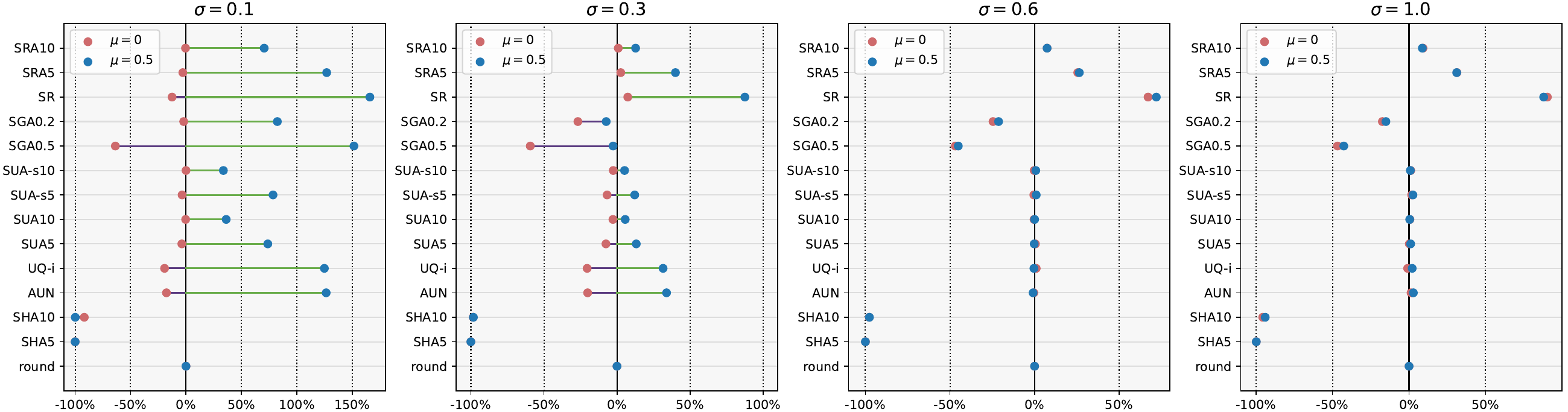}
        \label{fig:mse_single}
    } 
    \vspace{-0.4em}
    \quad
    \subfloat[Visualization of the learned synthesis transforms of different forward calculations in (a) under the setting $Y\sim\mathcal{N}(0,0.3^2)$.]
    { 
        \includegraphics[width=\linewidth]{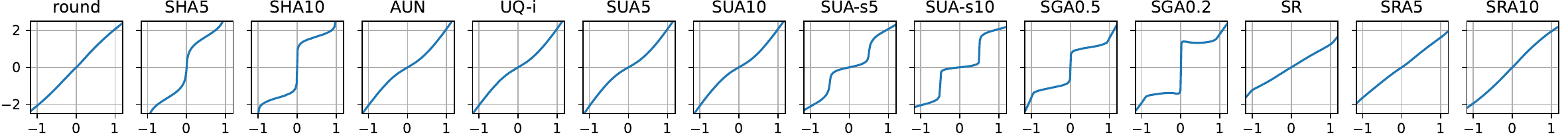}
        \label{fig:func_visual}
    }
    \vspace{-0.4em}
    \quad
    \subfloat[Distortion estimation error (relative), i.e., $(D(\tilde{Y})-D(\lfloor Y \rceil))/D(\lfloor Y \rceil)$, of various forward calculations with respect to different latent distributions. $X\sim \mathcal{N}(0,\Sigma),~Y=\sigma X,~Y\sim \mathcal{N}(0,\sigma^2\Sigma)$, $\Sigma$ is described in Eq. (\ref{eq:multi}).]{  
        \includegraphics[width=\linewidth]{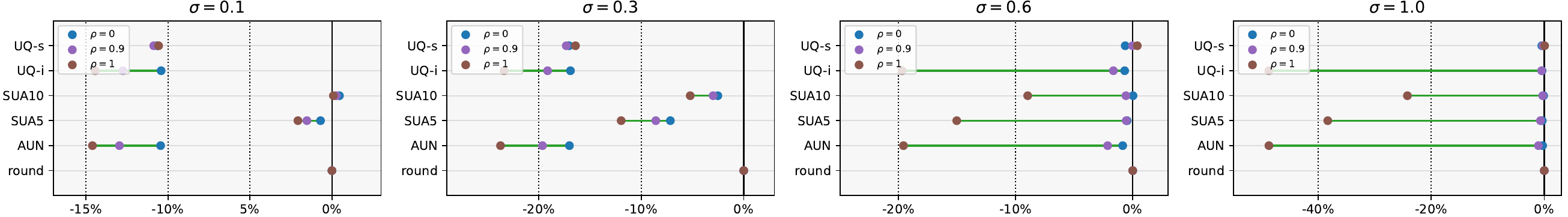}
        \label{fig:mse_multi}
    }
    \vspace{-0.4em}
    \caption{
    Visualization of the distortion estimation error with respect to various forward calculations and latent distributions. 
    }
    \label{fig:mse}
\vspace{-0.5em}
\end{figure*}

\begin{figure}[ht]
    \vspace{-0.4cm}
    \centering
    \subfloat{ 
    \includegraphics[width=0.38\linewidth]{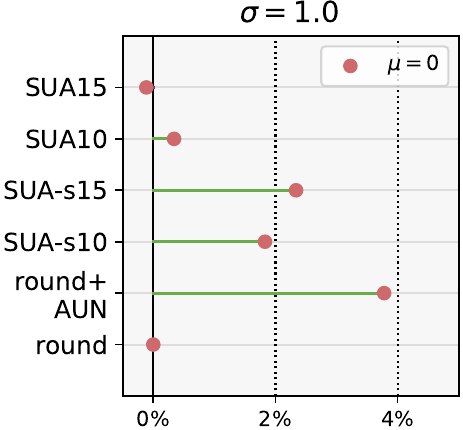}
    \label{fig:mse_extra}
    }
    \subfloat{ 
    \includegraphics[width=0.51\linewidth]{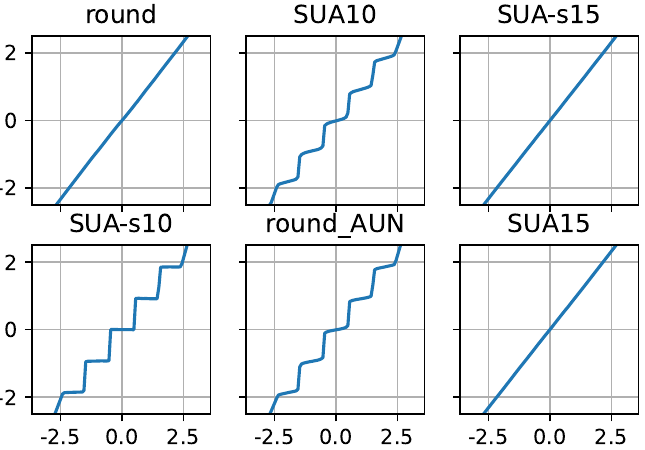}
    \label{fig:func_extra}
    }
    \vspace{-0.6em}
    \caption{
    Comparison between SUA and SUA-n. Left: distortion estimation error (relative), i.e., $(D(\tilde{Y})-D(\lfloor Y \rceil))/D(\lfloor Y \rceil)$. Right: visualization of the learned synthesis transforms.
    }
    \label{fig:extra}
\vspace{-0.5em}
\end{figure}

\subsubsection{Simulations on Gaussian Source}
\label{sec_distor:gaussian}
To better reflect the distortion estimation error of different forward calculations, we simulate with Gaussian sources with training a synthesis transform.
We set a scalar Gaussian source $X\sim\mathcal{N}(0,1)$, and restrict the analysis transform to generate $Y=\sigma X+\mu,~Y\sim\mathcal{N}(\mu,\sigma^2)$. We then train a synthesis transform with respect to each forward calculation to reconstruct $\tilde{X}$ from $\tilde{Y}$. We use mean-square error as the distortion measure, $d=\|x-\tilde{x}\|^2_2$. We define the distortion estimation error $\Delta D=D(\tilde{Y})-D(\lfloor Y \rceil)$, which reflects the degree of train-test mismatch in the distortion term.
Fig. \ref{fig:mse_single} shows the relative distortion estimation error $\Delta D/D(\lfloor Y \rceil)$ of different forward calculations, and the variation with respect to the distribution of $Y$.

As shown in Fig. \ref{fig:mse_single}, since SHA is an invertible function, $D(\tilde{Y})$ of SHA is almost zero, which is significantly different from rounding. 
For AUN, $|\Delta D|$ is large when $\sigma$ of the latent distribution is small, and it is highly related to $\mu$. As $\sigma$ increases, $|\Delta D|$ of AUN gradually diminishes. For UQ-i, $\Delta D$ is similar to AUN since they have the same marginal distribution \cite{agustsson2020universally}. By incorporating the annealing technique with AUN, such as SUA, SUA-n, $|\Delta D|$ decreases as the parameter $\alpha$ increases. 
As shown in Fig. \ref{fig:func_visual}, compared to SUA-n, SUA has a denoising function, $r_{\alpha}$, that can relieve the synthesis transform from implementing such a function.
Unlike AUN and SUA, whose $|\Delta D|$ are negligible for a large $\sigma$, the estimated distortion of SGA is always significantly different from rounding. Additionally, the synthesis transform learned with SGA is very sharp. For SR, $|\Delta D|$ is large in most cases. When $\sigma$ is large, the information transmitted by SR is much smaller than rounding, resulting in a larger estimated distortion than rounding. For SRA, as the parameter $\alpha$ increases, the stochasticity of SRA decreases, resulting in $|\Delta D|$ diminishes.

We further investigate the difference between SUA and SUA-n in detail in Fig. \ref{fig:extra}. Theoretically, the information transmitted by SUA and SUA-n with the same $\alpha$ is the same. However, we observe $D(\tilde{Y})$ of SUA-n is larger than SUA when $\sigma$=1, indicating that the neural network may not be that powerful to fit an ideal denoising function. We provide a more extreme example to illustrate how the capacity of the synthesis transform can influence the distortion estimation error. Adding uniform noise to the rounded latent preserves the same information as directly rounding the latent, but the estimated distortion can be significantly different. The ideal synthesis transform for the rounded latent after adding uniform noise is a step function, which is challenging to learn through gradient descent.

We then study the difference between UQ-i and UQ-s. If $Y$ is a scalar, the properties of UQ-s and UQ-i are the same. To investigate the properties of UQ-s, we further consider a two-dimensional Gaussian source $X$ in our experiments, $X\sim \mathcal{N}(0,\Sigma),~ Y=\sigma X,~Y\sim\mathcal{N}(0,\sigma^2 \Sigma),~Y=[Y_1,Y_2]$,
 \begin{equation}
 \Sigma=
 \left[
 \begin{array}{cc}
     1 & \rho \\
     \rho & 1 
 \end{array}
 \right],    
  \sigma^2 \Sigma=
 \left[
 \begin{array}{cc}
     \sigma^2 & \sigma^2\rho \\
     \sigma^2\rho & \sigma^2
 \end{array}
 \right].
 \label{eq:multi}
 \end{equation}
Fig. \ref{fig:mse_multi} presents the key difference between UQ-i and UQ-s, that the estimated distortion $D(\tilde{Y})$ of UQ-i is heavily influenced by the covariance of $Y$, while UQ-s is not. When the correlation coefficient $\rho=0$, indicating that $Y_1$ is independent of $Y_2$, $\Delta D$ of UQ-i and UQ-s are similar. However, when $\rho=1$, indicating that $Y_1=Y_2$, the estimated distortion $D(\tilde{Y})$ of UQ-i is significantly different from that of UQ-s and much smaller than $D(\lfloor Y \rceil)$. The estimated distortion of AUN and UQ-i is the same. For a large $\sigma$, when $\rho=1$, $|\Delta D|$ is large even if applying SUA. From a theoretical perspective, when $\rho=1$, rounding has $I(Y;\lfloor Y\rfloor)=I(Y_1;\lfloor Y_1 \rfloor)$; UQ-s has $I(Y;\tilde{Y})=I(Y_1; \tilde{Y_1})$; and UQ-i has $I(Y;\tilde{Y})=2I(Y_1; \tilde{Y_1})-I(\tilde{Y_1};\tilde{Y_2})\geq I(Y_1;\tilde{Y_1})$.

In summary, regarding distortion estimation, AUN outperforms SHA and SR. UQ-s enhance AUN when there is a high correlation among latent. SUA and SUA-n enhance AUN when $\sigma$ of the latent distribution is small. 
The improvement provided by SUA increases as $\alpha$ increases, while this effect does not hold true for SUA-n. 
In the case of SGA and SRA, using the commonly employed annealing coefficient does not yield better performance compared to AUN in most scenarios.

\begin{figure*}[ht]
    \centering
    \includegraphics[width=\linewidth]{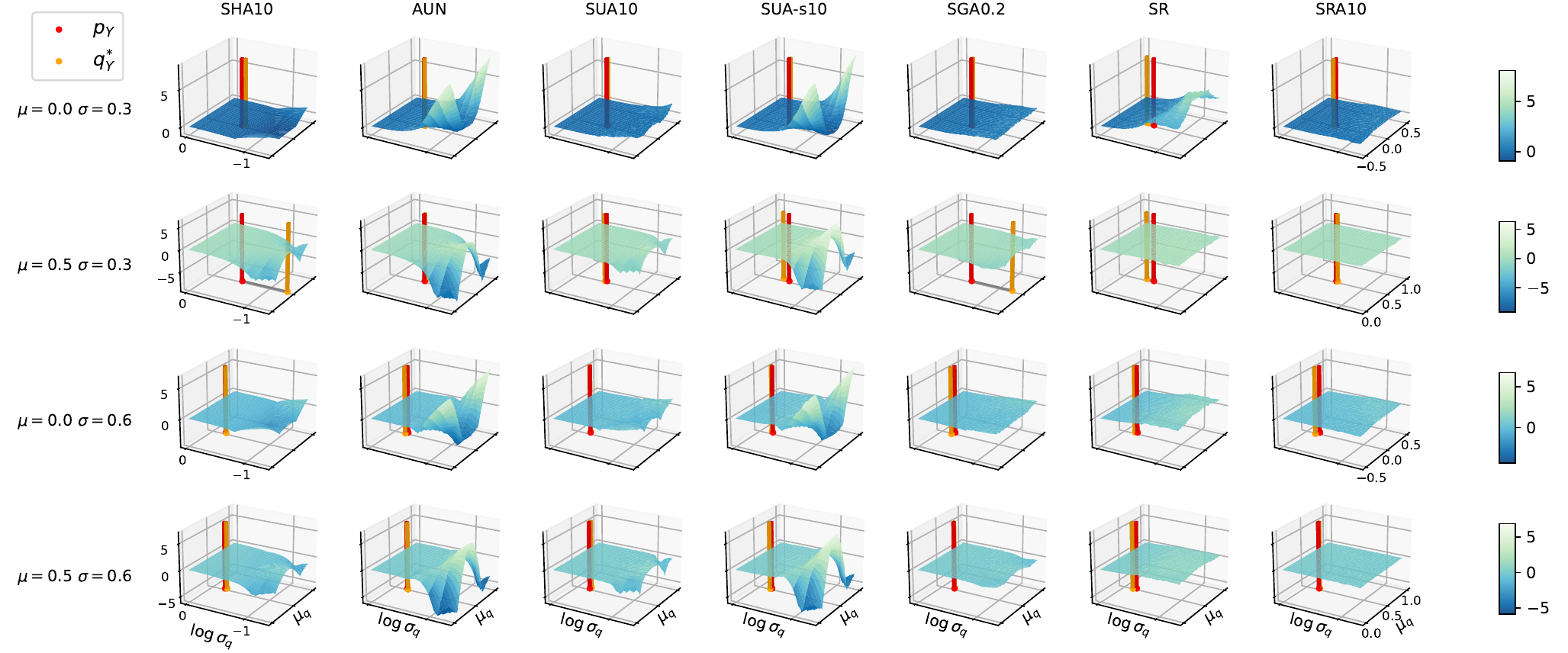}
    \vspace{-1.5em}
    \caption{
    Rate estimation error of various forward calculations with respect to different latent distributions. Each subplot shows the rate estimation error with respect to different estimated latent distributions and the distance between the optimal estimated distribution and the actual distribution.
    }
    \label{fig:rate}
\vspace{-0.3em}
\end{figure*}

\begin{figure}[ht]
    \centering
    \includegraphics[width=0.94\linewidth]{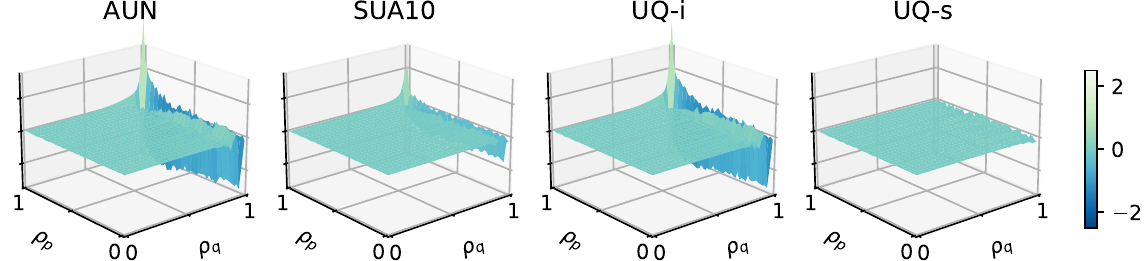}
    \vspace{-0.5em}
    \caption{
    Rate estimation error. Both the latent distribution and estimated distribution are two-dimensional Gaussian as described in Eq. \ref{eq:multi_rate}.
    }
    \label{fig:rate_multi}
\vspace{-0.5em}
\end{figure}

\subsection{Rate Estimation }
\label{sec:rate}
Most studies based on uniform scalar quantization employ a continuous cumulative distribution function (CDF) \label{para:cdf} to estimate the latent distribution.  
Let $c(\cdot)$ represent the estimated CDF. The probability of $\hat{y}$ is calculated through 
\begin{equation}
    q_{\hat{Y}}({\hat{y}})=\int_{\hat{y}-0.5}^{\hat{y}+0.5}q_Y(y)d y=c(\hat{y}+0.5)-c(\hat{y}-0.5).
\end{equation}
During training, the estimated rate is computed as
\begin{equation}\label{eq:rate}
    R(\tilde{Y}) = \mathbb{E}_{p_Y} [-\log q_{\tilde{Y}}(\tilde{y})],
\end{equation}
where $q_{\tilde{Y}}({\tilde{y}})=c(\tilde{y}+0.5)-c(\tilde{y}-0.5)$.
Gaussian distribution is widely used as a parametric probability model \cite{balle2018variational}. In our analyses, we set that both the latent distribution $p_{Y}$ and the estimated distribution $q_{Y}$ are Gaussian. The parameters of $q_{Y}$ are $\mu_q$ and $\sigma_q$. In our simulations, we set $Y\sim\mathcal{N}(\mu,\sigma^2),~\mu \in \{0,0.5\},~\sigma\in \{0.1,0.3,0.6\}$, and the estimated CDF is $c(x;\mu_q,\sigma_q)=\Phi(\frac{x-\mu_q}{\sigma_q})$, where $\mu_q \in [\mu-0.5,\mu+0.5],\sigma_q\in [0.05,1]$, and $\Phi(\cdot)$ is the CDF of standard Gaussian distribution. The rate of various forward calculations is then calculated using Eq. (\ref{eq:rate}). We define the rate estimation error $\Delta R=R(\tilde{Y})-R(\lfloor Y \rceil)$ to measure the train-test mismatch in the rate term. In addition, the distance between the optimal estimated distribution $q_Y^*=\mathop{\arg\min}_{q_Y} R(\tilde{Y})$
and $p_Y$ can also reflect the train-test mismatch in the rate term. 

The results are shown in Fig. \ref{fig:rate}. For SHA, when $\mu=0.5$, $|\Delta R|$ is large if $\sigma_q$ is small, and $q_Y^*$ is far away from $p_Y$. For AUN, $|\Delta R|$ is always large when $\sigma_q$ is small, but $q_Y^*$ estimated by AUN is close to $p_Y$. Compared to AUN, SUA makes $|\Delta R|$ much smaller and maintains $q_Y^*$ close to $p_Y$. For SUA-n, $|\Delta R|$ is large when $\sigma_q$ is small, and the distance between $q_Y^*$ and $p_Y$ increases compared to SUA. $|\Delta R|$ of SGA is relatively closer to $R(\lfloor Y \rceil)$ compared with AUN, while $q_Y^*$ differs a lot from $p_Y$ when $\mu=0.5$. For SR, $R(\tilde{Y})$ is larger than $R(\lfloor Y \rceil)$ in most cases, and $q_Y^*$ is slightly different from $p_Y$. After incorporating the annealing technique, $R(\tilde{Y})$ of SRA is much closer to $R(\lfloor Y \rceil)$ compared with SR. 

To study the properties of UQ-s, we further set $Y$ as a two-dimensional Gaussian variable, $Y\sim\mathcal{N}(0,\Sigma_p)$. The estimated distribution is also a two-dimensional Gaussian $\mathcal{N}(0,\Sigma_q)$,
 \begin{equation}
 \Sigma_p=
 \left[
 \begin{array}{cc}
     1 & \rho_p \\
     \rho_p & 1 
 \end{array}
 \right],    
\Sigma_q=
 \left[
 \begin{array}{cc}
     1 & \rho_q \\
     \rho_q & 1 
 \end{array}
 \right].
 \label{eq:multi_rate}
 \end{equation}
 As shown in Fig. \ref{fig:rate_multi}, $|\Delta R|$ of AUN and UQ-i are large when the correlation among latent is high. SUA can improve the performance of AUN, but cannot beat UQ-s. 

 In summary, for almost all forward calculations, $|\Delta R|$ is larger when $\sigma_q$ is small compared to when $\sigma_q$ is larger. 
 On average, SUA and SRA are superior to other forward calculations. UQ-s performs well when there is a high correlation among latent.
\vspace{-0.5em}
\subsection{Gradient Estimation}
\label{sec:gradient}
Gradient estimation risk consists of the bias and variance of gradient estimation, which are crucial factors in assessing the performance of gradient estimator \cite{mohamed2020monte}. Unbiasedness is usually preferred as it ensures the convergence of stochastic optimization procedures. Low-variance gradient facilitates more efficient learning by allowing for larger step sizes. Ghosh \textit{et al.} \cite{ghosh2023gradient} show that the effect of bias and variance of gradient estimation is closely related to network architecture. In our study, we focus on analyzing the gradient estimation risk of STE and PGE when applied to different forward calculations. Notably, we do not study the gradient bias and variance caused by stochastic gradient descent.

When rounding is employed as the forward calculation, STE is necessary during training. For differential deterministic functions, such as SHA, standard gradient descent is feasible. Regarding stochastic forward calculations, the expected loss function is usually differentiable with respect to $y$, indicating the existence of the expected gradient. However, directly calculating the expected gradient, which will be discussed in Sec. \ref{sec_gradient:ep}, is computationally prohibitive for high-dimensional latent spaces. In the case of reparameterization-based calculations like AUN, SUA, and SGA, the gradient of the expected loss function can be estimated using PGE. Such an unbiased estimator may introduce high variance when applied to annealing-based methods. To mitigate this, the generalized STE can be applied to SUA and SGA to reduce gradient estimation variance. For stochastic calculations involving a hard function, such as UQ-s, SR, and SRA, we only consider applying STE. 

\subsubsection{Calculation of Expected Gradient}
\label{sec_gradient:ep}
For stochastic forward calculations like AUN, SUA, SR, and SRA, the loss function is typically a continuous function with respect to $y$. We take SUA and SRA as examples to show the calculation of the expected gradient. We assume $y\in\mathbb{R}^{n},~u\sim\mathcal{U}(-0.5,0.5)^n$ . The expected loss and expected gradient of using SUA are
\begin{equation}\label{eq_univesalsoft4}
\begin{aligned}
\mathbb{E}_{p_{U}}L(\tilde{y})&=\int_{u} L(\tilde{y}) du=\int_{u_i}f(\tilde{y_i})du_i
\end{aligned}
\end{equation}
\begin{equation}\label{eq_univesalsoft5}
\begin{aligned}
\frac{\partial \mathbb{E}_{p_{U}}L(\tilde{y})}{\partial y_i}&=\frac{\partial \mathbb{E}_{p_{U}}L(\tilde{y})}{\partial s_{\alpha}(y_i)}\frac{\partial s_{\alpha}(y_i)}{\partial y_i}\\
&=\left[f(\tilde{y_i})|_{u_i=0.5}-f(\tilde{y_i})|_{u_i=-0.5}\right]\frac{\partial s_{\alpha}(y_i)}{\partial y_i},\\
\text{where }& f(\tilde{y_i})=\int_{u_{j\neq i}}L(\tilde{y_i},\tilde{y}_{j\neq i})du_{j\neq i}.
\end{aligned}
\end{equation}
The expected gradient of $y_i$ remains an integral, which is challenging to directly compute. However, this gradient can be easily estimated using PGE,
\begin{equation}\label{eq_suaunbiased}
\begin{aligned}
\frac{\partial \mathbb{E}_{p_{U}}L(\tilde{y})}{\partial y_i}
&=\left[\mathbb{E}_{p_{U}} \frac{\partial L(\tilde{y})}{\partial \tilde{y}_i}\frac{\partial \tilde{y}_i}{\partial s_{\alpha}(y_i)} \right]\frac{\partial s_{\alpha}(y_i)}{\partial y_i}, \\
\text{where }& \tilde{y}_i=r_{\alpha}(s_{\alpha}(y_i)+u_i).
\end{aligned}
\end{equation}
For rate term, $R(\tilde{y})=\sum^{n}_{i=1}R(\tilde{y}_i)$, we can directly calculate the expected gradient,
\begin{equation}\label{eq_suarep}
\begin{aligned}
\frac{\partial \mathbb{E}_{p_{U}}R(\tilde{y})}{\partial y_i}&=\frac{\partial \mathbb{E}_{p_{U_i}}R(\tilde{y}_i)}{\partial y_i}\\
&=\frac{\partial s_{\alpha}(y_i)}{\partial y_i}[R(\tilde{y}_i)|_{u_i=0.5}-R(\tilde{y}_i)|_{u_i=-0.5}].
\end{aligned}
\end{equation}
If $y$ is a scalar, we can also calculate the expected gradient of the distortion term. In the case of using SRA, the expected loss function and expected gradient are 
\vspace{-0.3em}
\begin{equation}\label{eq_univesalsoft6}
\begin{aligned}
\mathbb{E}_{p_{\tilde{Y}}}L(\tilde{y})&=\sum_{\tilde{y}} L(\tilde{y})p_{\tilde{Y}}(\tilde{y})=\sum_{\tilde{y}_i}p_{\tilde{Y}_{i}}(\tilde{y}_i)f(\tilde{y}_i)
\end{aligned}
\vspace{-0.3em}
\end{equation}
\begin{equation}\label{eq_univesalsoft7}
\begin{aligned}
\frac{\partial \mathbb{E}_{p_{\tilde{Y}}}L(\tilde{y})}{\partial y_i}&=\sum_{\tilde{y}_i} \frac{\partial p_{\tilde{Y}_{i}}(\tilde{y}_i)}{\partial y_i}f(\tilde{y}_i)\\
&=[f(\lceil y_i \rceil)-f(\lfloor y_i \rfloor)]\frac{\partial s_{\alpha}(y_i)}{\partial y_i}\\
\text{where }&f(\tilde{y}_i)=\sum_{\tilde{y}_{j\neq i}}L(\tilde{y_i},\tilde{y}_{j\neq i})p_{\tilde{Y}_{j\neq i}}(\tilde{y}_{j\neq i}).\\
\end{aligned}
\vspace{-0.3em}
\end{equation}
The expected gradient of SRA can always be calculated, but the computational complexity increases rapidly as the dimension of $y$ grows. Similar to SUA, for the rate term, the expected gradient can be easily computed, 
$\frac{\partial \mathbb{E}_{p_{\tilde{Y}}}R(\tilde{y})}{\partial y_i}=\frac{\partial s_{\alpha}(y_i)}{\partial y_i}[R(\lceil y_i \rceil)-R(\lfloor y_i \rfloor)]$.
Once we obtain the expected gradient of $y$, we can compute the gradient of $\phi$ using the chain rule, $\frac{\partial \mathbb{E}_{p_{\tilde{Y}}}L(\tilde{y})}{\partial \phi}=\frac{\partial \mathbb{E}_{p_{\tilde{Y}}}L(\tilde{y})}{\partial y}\frac{\partial y}{\partial \phi}$.

\subsubsection{Bias of Gradient Estimator}
In this section, we show the gradient estimation bias when STE is applied. We denote $g$ as the estimated gradient and $g_{ep}$ as the gradient of expected loss. Then the gradient bias can be defined as $bias=\mathbb{E}_{p_g}|g-g_{ep}|$. We take SUA and SRA as examples to show the calculation of gradient bias. When applying generalized STE to SUA, the gradient propagated to $y_i$ is estimated by
\begin{equation}\label{eq_suaste}
\begin{aligned}
\frac{\partial \mathbb{E}_{p_{U}}L(\tilde{y})}{\partial y_i}
&\approx\left[\mathbb{E}_{p_{U}} \frac{\partial L(\tilde{y})}{\partial \tilde{y}_i} \right]\frac{\partial s_{\alpha}(y_i)}{\partial y_i}. \\
\end{aligned}
\end{equation}
When applying STE to SRA, the gradient propagated to $y_i$ is estimated by
\vspace{-0.3em}
\begin{equation}\label{eq_srabiased}
\begin{aligned}
\frac{\partial \mathbb{E}_{p_{\tilde{Y}}}L(\tilde{y})}{\partial y_i}
&\approx\left[\sum_{\tilde{y}} \frac{\partial L(\tilde{y})}{\partial \tilde{y}_i} p_{\tilde{Y}}(\tilde{y}) \right]\frac{\partial s_{\alpha}(y_i)}{\partial y_i}. \\
\end{aligned}
\vspace{-0.5em}
\end{equation}
\begin{table}[t]
\centering
\caption{Bias and variance of gradient estimation of the rate term.}
\vspace{-0.9em}
\begin{threeparttable}
    \begin{tabular}{ c|c|cc|cc } 
    \hline
    \multicolumn{2}{c|}{methods}  & \multicolumn{2}{c|}{$\sigma_q=0.3$} &\multicolumn{2}{c}{$\sigma_q=1.0$} \\
    \hline
    backward&forward & bias & var & bias& var \\
    \hline
    \multirow{3}{*}{\Centerstack[c]{PGE}}&AUN&0.00&11.93&0.00&0.15\\
    &SUA5\tnote{1}&0.00&433.86&0.00&6.33\\
    &SUA10&0.00&29,351.05&0.00&432.30\\
    \hline
    \multirow{5}{*}{STE}&SUA5&1.36&29.28&0.16&0.36\\
    &SUA10&2.10&74.88&0.24&0.92\\
    &SR&2.95&22.30&0.33&0.30\\
    &SRA5&2.96&46.19&0.33&0.62\\
    &SRA10&2.96&92.58&0.33&1.23\\
    \hline
    \end{tabular}
    \label{tab:bias_var_rate}
 \begin{tablenotes}
    \footnotesize
    \item[1] SUA5 stands for SUA with a temperature coefficient of 5, similar to SUA10, SRA5, etc.
\end{tablenotes}
\end{threeparttable}
\vspace{-1.2em}
\end{table}

\begin{figure}[t]
    \centering
    \includegraphics[width=0.9\linewidth]{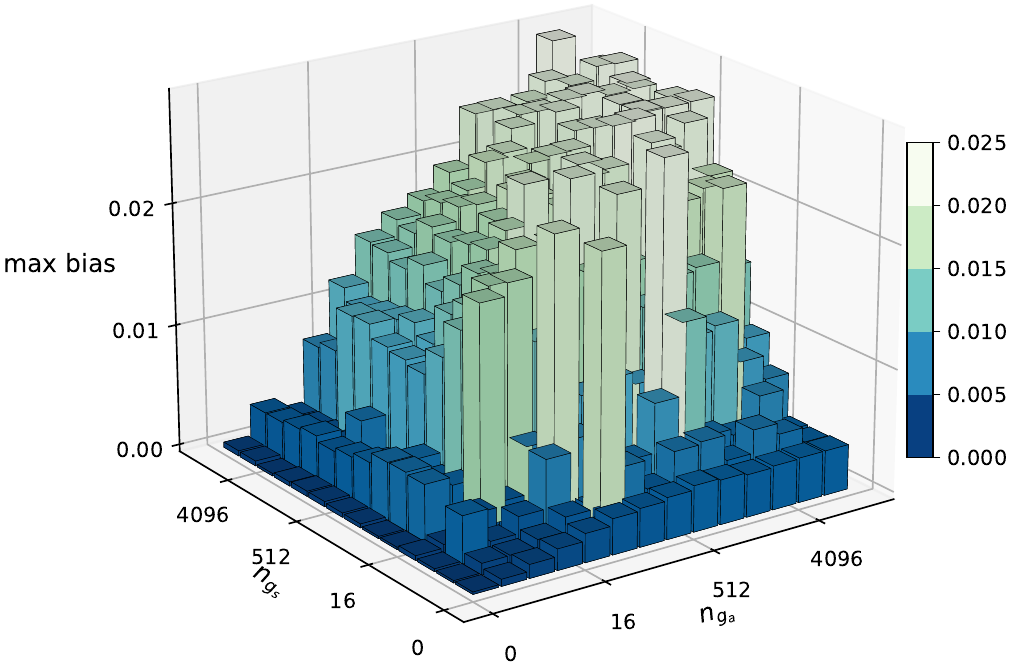}
    \vspace{-1em}
    \caption{Maximal value of gradient estimation bias with respect to the number of hidden units in the analysis transform $n_{g_a}$ and in the synthesis transform $n_{g_s}$.}
    \label{fig:bias_net}
    \vspace{-0.5em}
\end{figure}

Table \ref{tab:bias_var_rate} presents the average bias of gradient estimation in rate term when STE is applied. The bias of SUA increases as $\alpha$ increases. The change in bias of SRA is not significant when varying $\alpha$. Moreover, it appears that a larger $\sigma_q$ tends to result in a smaller gradient estimation bias.

We further analyze how network structures affect the gradient estimation bias. Note that we did not consider network structure when studying the gradient estimation bias in the rate term. 
We utilized the MNIST dataset \cite{deng2012mnist} to train compression networks with various network sizes. The dimension of latent  $y$ is 2.  
Fig. \ref{fig:bias_net} shows the maximum gradient estimation bias of $\frac{\partial D}{\partial \phi}$ with different numbers of hidden units in analysis transform and synthesis transform when the forward calculation is SR and backward calculation is STE. It can be observed that the maximum bias of $\frac{\partial D}{\partial \phi}$ roughly increases with the size of the analysis and synthesis transforms growing.
This implies that larger and more powerful networks suffer more from the gradient bias, and then may not be well-trained when STE is applied.

\begin{figure}[t]
    \centering
    \includegraphics[width=0.9\linewidth]{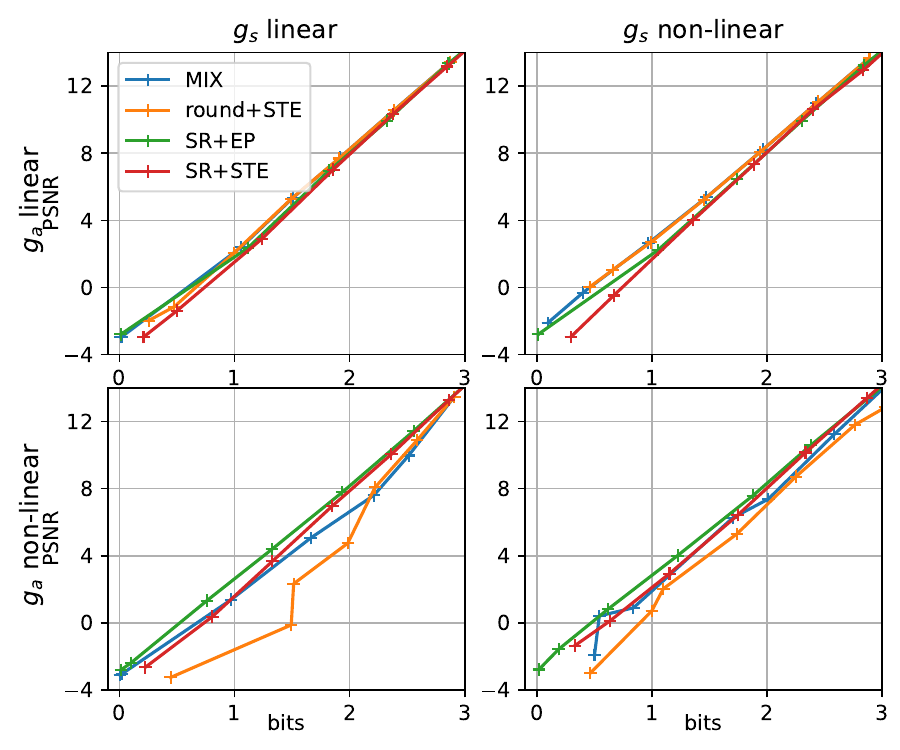}
    \vspace{-1em}
    \caption{Rate-distortion performance on the standard Laplacian source with respect to different quantization surrogates and network structures.}
    \label{fig:laplace}
    \vspace{-1.5em}
\end{figure}
We conducted experiments using a standard Laplacian source to illustrate the effect of bias on performance. Both the source $X$ and the latent $Y$ are scalar, which makes it easier to directly calculate the expected gradient. We explore various settings, including linear and non-linear transforms. The linear transform utilizes a basic affine transform, while the non-linear transform employs stacked residual blocks. The experimental results are presented in Fig. \ref{fig:laplace}.
When the forward calculation is SR, optimizing networks using the expected gradient consistently outperforms using STE. When the analysis transform is linear, the difference between using STE and the expected gradient is small. However, when the analysis transform is non-linear, using the expected gradient yields significantly better results than using STE. Furthermore, the performance gap between STE and the expected gradient is more pronounced at lower bitrate. This phenomenon aligns with the analyses of gradient estimation bias in the rate term that a larger estimated $\sigma_q$ tends to result in a smaller bias.
In the case where the forward calculation is rounding, the performance with a linear analysis transform is better than using SR. This is because there is no train-test mismatch when the forward calculation is rounding, and the effect of bias is relatively small.
However, when the analysis transform is non-linear, the performance becomes extremely poor. Training with the MIX method slightly improves the performance, but it still fails when the analysis transform is non-linear. 
\vspace{-0.5em}
\subsubsection{Variance of Gradient Estimator}
The variance of gradient estimation arises from the stochastic nature of stochastic forward calculations and PGE. It is defined as $var=\mathbb{E}_{p_g}\|g-\mathbb{E}_{p_g}[g]\|_2^2$. As described in Eq. (\ref{eq_suaunbiased}), (\ref{eq_suaste}), and (\ref{eq_srabiased}), the gradient is estimated through sampling, which introduces variance. We present the gradient estimation variance of the rate term in Table \ref{tab:bias_var_rate}. 
When the forward calculation is SUA and the backward calculation is PGE, the variance increases rapidly as $\alpha$ increases.  
Applying generalized STE to SUA helps reduce the gradient variance but introduces bias. Moreover, the variance diminishes as $\sigma_q$ increases. For SRA with STE, the variance is always larger than SUA with the same temperature coefficient.

\begin{table}[t]
\centering
\caption{BD-rate (\%) tested on Kodak of training the mean-scale hyper model with zero-center quantization (MS-hyper-zero). The anchor is BPG. The best performance of each quantization surrogate is underlined.}
\vspace{-0.8em}
\begin{threeparttable}
    \tabcolsep=9.5pt
    \begin{tabular}{ cc|c|cc } 
    \hline
    rate & distortion & $\alpha_{\text{m}}$\tnote{1} & joint& post \tnote{2}\\
    \hline
    \multirow{3}{*}{SUA+PGE}&\multirow{3}{*}{SUA+STE}&4&-5.64&-9.02\\
    &&8&\underline{-8.32}&\underline{-9.36}\\
    &&12&-6.30&-6.89\\
    \hline
    \multirow{3}{*}{SUA+EP\tnote{3}}&\multirow{3}{*}{SUA+STE}&4&-6.19&-9.32\\
    &&8&-8.79&-9.78\\
    &&12&\underline{-9.14}&\underline{-9.89}\\
    \hline
    \multirow{3}{*}{SUA+EP}&\multirow{3}{*}{SUA+PGE}&4&-6.23&-9.63\\
    &&8&\underline{-8.50}&\underline{-9.69}\\
    &&12&-3.86&-4.53\\
    \hline
    \end{tabular}
 \begin{tablenotes}
    \footnotesize
    \item[1] $\alpha_m$ is the maximal temperature coefficient for annealing.
    \item[2] 'joint' stands for only applying joint-training, and 'post' stands for applying post-training after joint-training.
    \item[3] 'EP' stands for calculating the expected gradient during backward calculation.
\end{tablenotes}        
\end{threeparttable}
\label{tab:sua_bias_var}
\vspace{-1em}
\end{table}

\begin{table*}[t]
    \centering
    \caption{
    Loss reduction, denoted by $\Delta L$ (\%), tested on Kodak of MS-hyper-zero. $L_{\text{test}}$ is the true rate-distortion cost ($D+\lambda R$) calculated with rounding. $L_{\text{train}}$ is the approximate rate-distortion cost calculated with the quantization surrogate used during joint-training instead of rounding.}
    \vspace{-1em}
    \begin{threeparttable}
    \begin{tabular}{ c|c||cccc|c|c||cccc|c|c } 
    \hline
    \multicolumn{2}{c||}{} & \multicolumn{6}{c||}{$\Delta L (\%),~\lambda=0.0018$} &\multicolumn{6}{c}{$\Delta L(\%),~\lambda=0.0130$} \\
    \hline
      \multicolumn{2}{c||}{transforms} & \multicolumn{4}{c|}{$ \Delta L_{\text{test}}$} & \multicolumn{2}{c||}{$\Delta L_{\text{train}}$} &\multicolumn{4}{c|}{$\Delta L_{\text{test}}$} & \multicolumn{2}{c}{$\Delta L_{\text{train}}$}\\
    \hline
    $g_a$&$g_s$&\Centerstack[c]{AUN\\joint}&\Centerstack[c]{MIX\\joint}&\Centerstack[c]{AUN\\post}&\Centerstack[c]{MIX\\post}&\Centerstack[c]{AUN\\joint}&\Centerstack[c]{MIX\\joint}&\Centerstack[c]{AUN\\joint}&\Centerstack[c]{MIX\\joint}&\Centerstack[c]{AUN\\post}&\Centerstack[c]{MIX\\post}&\Centerstack[c]{AUN\\joint}&\Centerstack[c]{MIX\\joint}\\ 
    \hline
    res&res&0.00\tnote{1}&-2.43&-3.01&-3.30&0.00&0.00&0.00&-3.07&-3.47&-3.25&0.00&0.00\\
    res&res+attn \tnote{2}&3.01&-5.03&-2.56&-5.34&-1.89&-2.89&-1.92&-4.23&-4.35&-4.48&-1.24&-1.17\\
    res+attn&res&-0.75&-2.30&-3.47&-2.59&-0.38&1.77&-1.44&-1.64&-4.22&-1.50&-0.92&1.31\\
    res+attn&res+attn&0.71&-3.56&-3.37&-3.97&-2.83&1.02&-2.62&-3.03&-5.28&-2.90&-2.28&0.15\\
    \hline
    \end{tabular}
 \begin{tablenotes}
    \footnotesize
    \item[1] '0.00' stands for the anchor of the corresponding column/rows.
    \item[2] 'res' stands for using residual block, and 'attn' stands for using attention module.
\end{tablenotes}   
\end{threeparttable}
\label{tab:tradeoff}
\vspace{-1em}
\end{table*}
The bias and variance of gradient estimation are inevitable when directly computing the expected gradient is not feasible. We conduct experiments on the 
mean-scale hyper (MS-hyper) model  (details in Sec. \ref{para:mshyper})
to illustrate the effect of gradient bias and variance. During training, we linearly anneal $\alpha$ to $\alpha_{\text{m}}$. As shown in Table \ref{tab:sua_bias_var}, optimizing the rate term with the expected gradient consistently outperforms using PGE, which suffers from higher variance. Regarding the distortion term, the PGE performs slightly better than STE with a small $\alpha_{\text{m}}$, since PGE is unbiased and the model may be able to tolerate the gradient variance of PGE when $\alpha=4$. However, as $\alpha$ increases, the performance of PGE, which is prone to high variance, becomes inferior to using STE.
\vspace{-0.5em}
\subsubsection{Summary}
For the rate term, directly calculating the expected gradient can be easily applied for most surrogates, which introduce no gradient estimation risk. For the distortion term, PGE or STE is required. The effect of gradient bias caused by STE is highly related to network structures and may become severe as the network complexity increases. For annealing-based methods, such as SUA and SRA, the gradient estimation risk increases as the temperature coefficient increases. In particular, both PGE and STE can be applied to SUA, but the performance of these two estimators may differ across different models since the effect of bias and variance is related to network structure.

\vspace{-0.5em}
\subsection{Tradeoff between Train-Test Mismatch and Gradient Estimation Risk}
\label{sec:tradeoff}
When the objective is to avoid train-test mismatch, rounding must be used as the forward calculation. However, it is intractable to obtain a low gradient estimation risk. If the objective is to obtain a low-risk gradient estimator, using either stochastic or deterministic differential forward calculations is necessary. However, these forward calculations will inevitably introduce the train-test mismatch.
Training with rounding as the forward calculation and using STE as the backward calculation has been reported to result in poor performance \cite{minnen2020channel,guo2021soft,tsubota2023comprehensive}. On the other hand, if the forward calculation is AUN, the backward calculation could have a low risk. However, this approach introduces a significant train-test mismatch. The widely used MIX method combines these two methods, employing AUN for rate estimation and using rounding and STE for reconstruction.
The MIX method preserves biased estimator and train-test mismatch while resulting in better performance in \cite{minnen2020channel}. This implies that there may be a tradeoff between train-test mismatch and the gradient estimation risk.

We take two widely used quantization surrogates: AUN and MIX quantization methods as examples to show the negative effect of train-test mismatch and gradient estimation risk on practical learned image compression models
and how this negative effect varies across different models. 
Regarding the train-test mismatch, the AUN method is worse than MIX, while regarding the gradient estimation risk, the AUN method is better than MIX.
Evaluating the exact influence of train-test mismatch and gradient estimation risk is challenging since it is impossible to assess the performance without such effects. However, we can gain some insights into their effect by comparing the performance of AUN and MIX methods applied to different network structures, as shown in Table \ref{tab:tradeoff}. 

First, we examine how the gradient estimation risk affects performance. Since train-test mismatch is not a concern when assessing the influence of gradient estimation risk, we compare $L_{\text{train}}$, the approximate rate-distortion cost calculated with the forward calculation employed during joint-training instead of rounding. When training with MIX, adding the simplified attention block \cite{cheng2020learned} in analysis transform leads to an increase in $L_{\text{train}}$. However, when training with AUN, adding 'attn' in the analysis transform leads to a decrease in $L_{\text{train}}$. This phenomenon indicates that for the model we tested, the negative effect of gradient estimation risk may be more pronounced in models with more powerful analysis transforms. 

Next, we investigate how train-test mismatch affects performance. To assess the effect, we need to further consider the difference between the true rate-distortion cost $L_{\text{test}}$, calculated with rounding, and $L_{\text{train}}$. When training with AUN at low bitrate ($\lambda=0.0018$), further adding 'attn' in the synthesis transform causes an increase in $L_{\text{test}}$. However, $L_{\text{train}}$ continues to decrease. In contrast, when training with MIX, both $L_{\text{test}}$ and $L_{\text{train}}$ decrease simultaneously after adding 'attn' in the synthesis transform. This demonstrates that the effect of train-test mismatch is more severe when applying such a more powerful synthesis transform in our tested model. This observation is only prevalent at low bitrate and not at higher bitrate ($\lambda=0.0130$), which aligns with our analyses in Sec. \ref{sec:rate} that the mismatch between AUN and rounding is large if $\sigma$ of latent distribution is small. Notably, even though post-training is employed, the performance drop caused by train-test mismatch is not completely eliminated.

Both our analyses and simulations imply that there exists a tradeoff between the train-test mismatch and the gradient estimation risk, and this tradeoff varies across different network structures.

\vspace{-0.65em}
\subsection{Two Tricks}
\label{sec:subtle}
\vspace{-0.1em}
\subsubsection{Lower Bound for $\sigma_q$}
\label{para:sigma}
\begin{figure*}
    \centering
    \subfloat[]{ 
    \includegraphics[width=0.8\linewidth]{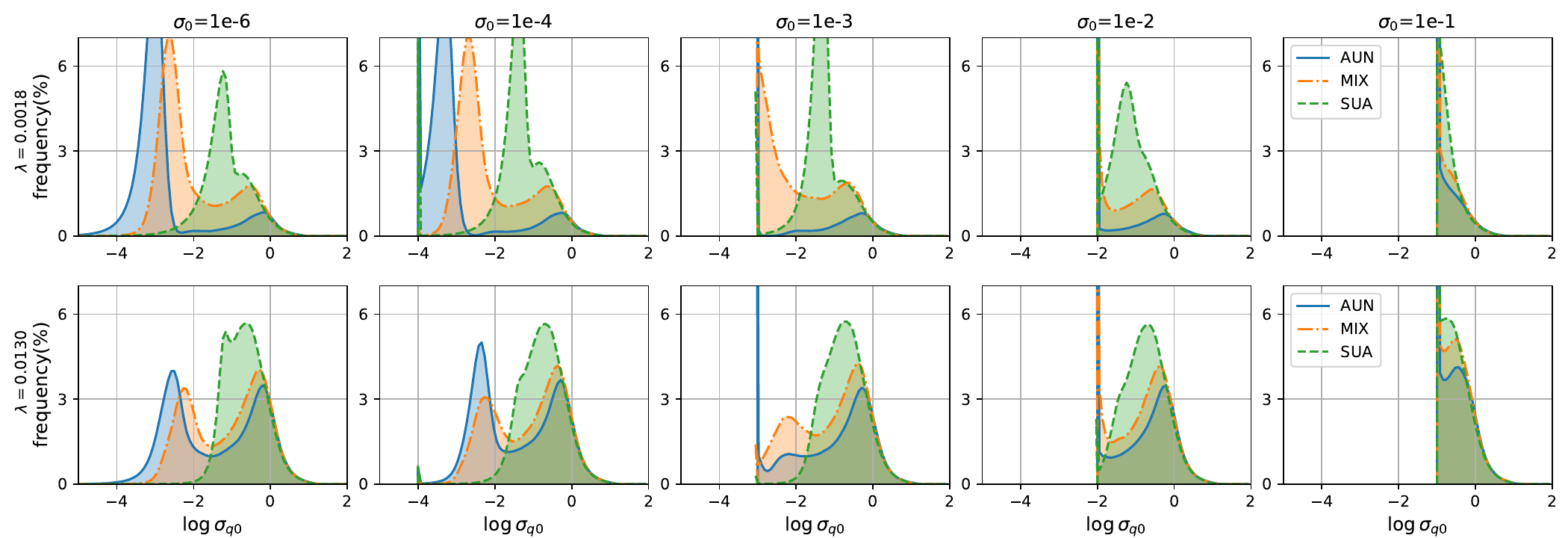}
    \label{fig:mse_extra2}
    }
    \subfloat[]{
    \includegraphics[width=0.17\linewidth]{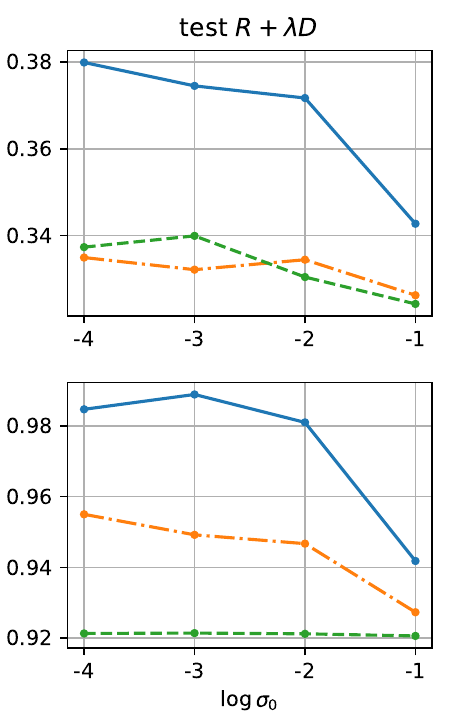}
    \label{fig:func_extra2}
    }
    \vspace{-0.7em}
    \caption{
    Visualization of the effect of $\sigma_0$ on the mean-scale hyper model with zero-center quantization (MS-hyper-zero). (a) Distribution of estimated $\sigma_{q0}$ with respect to different lower bound parameters and different quantization surrogates. (b) Rate-distortion cost (lower is better) on Kodak with respect to different lower bound parameters without post-training.
    }
\label{fig:lower_bound}
\vspace{-1em}
\end{figure*}

During training, to prevent the estimated variance parameter $\sigma_q$ from approaching zero (causing numerical calculation problems), a value $\sigma_0$, usually a tiny value, is imposed as a lower bound.
In Sec. \ref{sec:rate}, we have discussed that the train-test mismatch in most forward calculations is closely related to the value of $\sigma_q$. Therefore, in addition to being a numerical calculation trick, the bound $\sigma_0$ also affects the train-test mismatch by adjusting the distribution of $\sigma_q$. 

We denote the bounded variance parameter as 
\vspace{-0.3em}
\begin{equation}\label{eq:bound}
\begin{aligned}
\sigma_{q0}=
\begin{cases}
\sigma_q, \sigma_q>\sigma_0;\\
\sigma_0, \sigma_q\leq \sigma_0,
\end{cases}
\sigma_{q0}^*=\mathop{\arg\min}\limits_{\sigma_{q0}} R(\tilde{Y}),
\end{aligned}
\vspace{-0.2em}
\end{equation}
and $\sigma_{q0}^*$ is the optimal $\sigma_{q0}$. The optimal $\sigma_q$ before being bounded is $\sigma_q^*$. For a latent $Y\sim \mathcal{N}(0,\sigma^2)$, the information of the rounded latent $\lfloor Y \rceil$ tends to approach zero when $\sigma$ is extremely small, as shown in Fig. \ref{fig:mutual_info}. Therefore, the latent with an extremely small $\sigma$ can be ignored for both training and testing. 
Besides, the CDF utilized for lossless entropy coding is a list of quantized values, typically represented by 16-bit or 32-bit integers. Thus, if $\sigma_q$ is small enough, the quantized CDF remains unaltered. Therefore, it is possible to set an appropriate larger $\sigma_0$ during joint-training.
A proper lower bound helps 
to alleviate the train-test mismatch in the rate term while maintaining $\sigma_{q0}^*=\sigma_q^*$. Additionally, in the ideal case that the $\sigma$ of latent distribution is the same as the estimated $\sigma_{q0}^*$, the mismatch in distortion term can also be reduced by setting an appropriate $\sigma_0$. 

We visualize the distribution of $\sigma_{q0}$ with a tiny $\sigma_0=1e-6$ in Fig. \ref{fig:lower_bound}a. The distribution is collected from the Kodak dataset on the MS-hyper-zero model (model details in Sec. \ref{para:mshyper}). It can be observed that, for AUN and MIX, $\sigma_{q0}$ tends to be highly concentrated towards the lower end, especially at low bitrate ($\lambda=0.0018$), which introduces substantial train-test mismatch. 
With the assumption that the probability model is a zero-mean Gaussian, the quantized CDF with $\sigma_q<0.1$ remains unaltered. Then we present the results with varying $\sigma_0$ from 1e-4 to 1e-1, as shown in Fig. \ref{fig:lower_bound}a. Across three quantization methods, an increase in the lower bound from 1e-4 to 1e-1 leads to a gradual concentration of $\sigma_{q0}$ towards larger values, which reduces train-test mismatch. This reduction of train-test mismatch results in improved performance, especially at a lower bitrate, as shown in Fig. \ref{fig:lower_bound}b. For the SUA-based method, even with a small lower bound, it exhibits a concentration of $\sigma_{q0}$ towards larger values due to the reduced train-test mismatch brought by SUA. On average, the performance of SUA is better than MIX and AUN, as shown in Fig. \ref{fig:lower_bound}b.
Compared with AUN, using the MIX method results in a stronger concentration of $\sigma_{q0}$ towards larger values as there is no train-test mismatch in the distortion term.
\vspace{-0.3em}
\subsubsection{Zero-Center Quantization}
In learned image compression, several studies \cite{balle2020nonlinear,minnen2020channel,zou2022devil,liu2023learned,agustsson2020universally,xie2021enhanced} have employed zero-center quantization, denoted as $\hat{y}=\lfloor y-\mu_q\rceil+\mu_q$, where the quantization offset is aligned with the Gaussian conditional mean, as an alternative to nonzero-center quantization, denoted as $\hat{y}=\lfloor y\rceil$. 
Balle \textit{et al.} \cite{balle2020nonlinear} treat AUN as quantization with a random offset, and the optimal offset can be determined after training. They also suggest that when utilizing Gaussian as the probability model, setting the mode of the Gaussian distribution $\mu_q$, as the quantization offset yields the best performance. This aligns with the results presented in Table \ref{tab:zerocenter}, which shows that without joint-training, zero-center quantization outperforms nonzero-center quantization when training with AUN.
\begin{table}[t]
    \centering
    \caption{BD-rate (\%) tested on Kodak using nonzero-center quantization and zero-center quantization. The anchor is BPG.}
    \vspace{-0.9em}
    \begin{threeparttable}
    \begin{tabular}{ c|c|cccc} 
    \hline
    &\multirow{2}{*}{$\sigma_0$}&\multicolumn{2}{c}{AUN}&\multicolumn{2}{c}{MIX}\\
    &&joint&post&joint&post\\
    \hline
    MS-hyper\tnote{1} &1e-6 &11.96&8.05&7.20&5.89\\
    MS-hyper-zero &1e-6 &11.27&7.42&-3.18&-4.19\\
    \hline
    MS-hyper &0.11 &1.15&-6.13&-8.03&-9.20\\
    MS-hyper-zero &0.11 &-0.53&-5.78&-8.39&-8.93\\
    \hline
    Charm\tnote{2} &0.11 &1.79&-14.68&-13.01&-15.30\\
    Charm-zero &0.11 &-4.18&-18.14&-19.23&-19.96\\
    \hline
    \end{tabular}
 \begin{tablenotes}
    \footnotesize
    \item[1] MS-hyper/MS-hyper-zero stands for the mean-scale hyper model without/with zero-center quantization.
    \item[2] Charm/Charm-zero stands for the channel-wise autoregressive model without/with zero-center quantization.
\end{tablenotes}   
    \label{tab:zerocenter}
    \end{threeparttable}
\vspace{-0.9em}
\end{table}

For training with MIX, the quantization offset during training should be aligned with testing. Since there is no train-test mismatch in the distortion term during joint-training, the primary focus is on rate estimation. We assume the estimated $\mu_q$
is same with $\mu$. In the case of zero-center quantization, where $Y-\mu$ follows a zero-mean Gaussian distribution, the estimated rate consistently exceeds the transmitted information rate, $R(\tilde{Y}-\mu)=R(\tilde{Y})\geq H(\tilde{Y})\geq H(\lfloor Y-\mu\rceil)=I(Y-\mu;\lfloor Y-\mu\rceil)$.
Therefore, optimizing this upper bound of the true loss consistently leads to improvement in performance. In the case of nonzero-center quantization, the rate estimated by AUN may be significantly smaller than the actual rate, as illustrated in Fig. \ref{fig:rate}. Consequently, a decrease in the estimated loss function may not guarantee a decrease in the true loss. Table \ref{tab:zerocenter} demonstrates that when training with MIX, the performance is better by utilizing zero-center quantization, particularly with a tiny $\sigma_0$. 

\vspace{-0.7em}
\section{Method}
\label{sec:method}
\vspace{-0.1em}
\subsection{SUA-Based Methods}
As discussed in Sec. \ref{sec:tradeoff}, there is a tradeoff between the train-test mismatch and the gradient estimation risk, and this tradeoff varies across different models. 
Consequently, we tried to control this tradeoff flexibly, to boost the performance of different models. As discussed in Sec. \ref{sec:analysis}, 
annealing-based methods provide a way to control the tradeoff between the train-test mismatch and gradient estimation risk by adjusting the temperature coefficient. Take SUA as an example, by increasing $\alpha$, the difference between the estimated rate and distortion compared to rounding gradually diminishes, while the gradient estimation risk increases if directly calculating the expected gradient is not feasible.
\begin{table}[t]
    \centering
    \caption{BD-rate (\%) tested on Kodak with or without partial stop-gradient on MS-hyper-zero. The anchor is BPG.}
    \vspace{-0.9em}
    \begin{tabular}{ cc|c|cc }
    \hline
    rate & distortion&sg($\mu_q$)& joint& post \\
    \hline
    \multirow{2}{*}{SUA+EP}&\multirow{2}{*}{SUA-n+PGE}&\usym{2717}&4.96&-4.54\\
    &&\usym{2713}&2.39&-6.63\\
    \hline
    \multirow{2}{*}{SUA+EP}&\multirow{2}{*}{SUA+PGE}&\usym{2717}&-4.33&-6.92\\
    &&\usym{2713}&-8.50&-9.69\\
    \hline
    \multirow{2}{*}{SUA+EP}&\multirow{2}{*}{SUA+STE}&\usym{2717}&-6.91&-7.54\\
    &&\usym{2713}&-9.14&-9.89\\
    \hline
    \multirow{2}{*}{SRA+EP}&\multirow{2}{*}{SUA+STE}&\usym{2717}&2.37&-1.52\\
    &&\usym{2713}&-3.60&-4.08\\
    \hline
    \end{tabular}
    \label{tab:suas}
\vspace{-0.9em}
\end{table}
\begin{table*}
    \centering
    \caption{BD-rate (\%) tested on Kodak of different probability models and different values of the lower bound $\sigma_0$. Gaussian and Laplacian are applied on MS-hyper-zero, and Gaussian Mixture is applied on MS-hyper. The anchor is BPG. The best performance of each column is underlined.}
    \tabcolsep=5pt
    \vspace{-0.8em}
    \begin{threeparttable}
    \resizebox{\linewidth}{!}{ 
    \begin{tabular}{ c||cc|cc|cc||cc|cc|cc||c||cc|cc } 
    \hline
    \multicolumn{1}{c||}{} & \multicolumn{6}{c||}{BD-rate (\%),~Gaussian} &\multicolumn{6}{c||}{BD-rate (\%),~Gaussian Mixture }&\multicolumn{5}{c}{BD-rate (\%),~Laplacian} \\
    \hline
    \multirow{2}{*}{$\sigma_0$}&\multicolumn{2}{c|}{AUN}&\multicolumn{2}{c|}{MIX}&\multicolumn{2}{c||}{SUA}&\multicolumn{2}{c|}{AUN}&\multicolumn{2}{c|}{MIX}&\multicolumn{2}{c||}{SUA}&\multirow{2}{*}{$b_0$\tnote{1}}&\multicolumn{2}{c|}{AUN}&\multicolumn{2}{c}{MIX}\\
    &joint&post&joint&post&joint&post&joint&post&joint&post&joint&post&&joint&post&joint&post\\
    \hline
    1e-6&11.27&7.42&-3.18&-4.19&-7.16&-4.62&9.11&6.36&2.11&1.51&-7.47&-7.51&1e-6&12.07&8.12&-2.14&-2.99\\
    0.05&0.30&-5.20&-6.28&-6.85&-8.84&-9.44&-0.11&-6.16&-8.20&-8.69&-11.44&-12.15&0.01&8.78&5.91&-4.49&-4.34\\
    0.07&-0.76&-6.90&-7.45&-8.10&-8.64&-9.30&0.13&-7.54&-9.35&-9.85&-11.39&-12.06&0.02&5.56&2.05&-5.33&-5.58\\
    0.09&\underline{-1.36}&-7.81&-8.39&-8.99&-8.79&-9.49&-0.76&-8.28&-10.17&-11.06&-11.26&-11.96&0.04&1.01&-4.44&-7.44&-7.49\\
    0.11&-0.53&-5.78&-8.39&-8.93&\underline{-9.22}&\underline{-9.89}&\underline{-1.40}&\underline{-9.35}&\underline{-10.62}&-11.26&\underline{-11.54}&\underline{-12.21}&0.06&-0.05&-7.19&\underline{-8.85}&-7.92\\
    0.13&-0.68&-8.05&\underline{-9.24}&\underline{-9.78}&-9.18&-9.87&0.08&-9.08&-10.38&-11.09&-11.20&-11.93&0.08&\underline{-0.12}&-8.05&-8.84&-8.70\\
    0.16&0.07&\underline{-8.22}&-8.92&-9.62&-8.61&-9.69&0.01&-8.94&-10.29&\underline{-11.48}&-10.80&-11.99&0.11&3.02&-7.51&-6.87&\underline{-9.79}\\
    0.20&5.94&-7.67&-6.07&-9.44&-5.27&-9.70&6.39&-8.46&-6.69&-10.60&-7.14&-11.64&0.15&13.86&\underline{-8.49}&0.25&-9.67\\
    0.25&21.16&-6.81&3.16&-9.22&5.83&-9.02&20.61&-7.85&2.36&-10.29&3.53&-11.44&0.20&31.88&-8.33&13.72&-9.12\\
    \hline
    \end{tabular}
    }
 \begin{tablenotes}
    \footnotesize
    \item[1] $b_0$ stands for the lower bound for the scale parameter $b$ of the Laplacian model.
\end{tablenotes} 
    \end{threeparttable}
    \label{tab:lower_bound}
    \vspace{-1em}
\end{table*}

For distortion term, as discussed in Sec. \ref{sec:distortion}, SUA and SUA-n are superior to other annealing-based forward calculations. In the backward pass, PGE is required for SUA-n. For SUA, we have two options to estimate the gradient: employing PGE, which is unbiased and has a higher variance, or using the biased STE, which has a lower variance, as described in Eq. (\ref{eq_suaunbiased}) or (\ref{eq_suaste}). During joint training, we select either PGE or STE taking the tolerance to gradient bias and variance into account.
For rate term, as discussed in Sec. \ref{sec:rate}, two annealing-based forward calculations, SUA and SRA, perform well. In the backward pass, we can always directly calculate the expected gradient when using SUA and SRA.

For distortion term, using SUA shows superior performance compared with SUA-n, as evidenced in Table \ref{tab:suas}. The reason has been analyzed in Sec. \ref{sec:distortion}, where it is observed that although the information transmitted by SUA-n and SUA is identical, SUA alleviates the need for the network to implement a denoising function. For SUA, the best performance of using PGE and STE is comparable on MS-hyper-zero. For the rate term, using SUA is better than SRA. The reason may be that the estimated rate by SRA may differ from the actual information rate transmitted by SUA.

Consequently, we propose to adopt SUA-based methods to control the tradeoff between train-test mismatch and gradient estimation risk. In the backward pass, directly calculating the expected gradient is used for the rate term, and PGE and STE can be chosen for the distortion term by considering the tolerance to gradient bias and variance. Post-training can always be further used to improve performance after joint-training.

\vspace{-0.8em}
\subsection{Lower Bound with Two-Stage Training}
For nonzero-mean Gaussian or Gaussian Mixture, there is a risk that $\sigma_{q0}^*>\sigma_q^*$, even if $\sigma_0$ is extremely small. 
To address this issue, we propose to set different $\sigma_0$ during joint-training and post-training.
During joint-training, we advocate setting a relatively large lower bound, such as 0.11, for $\sigma_q$, which effectively reduces the train-test mismatch.  
During post-training, since there is no train-test mismatch, we set $\sigma_0$ as a tiny value, 1e-6, to make the estimated distribution more accurate. 

To determine a more appropriate lower bound during joint-training for the Gaussian model, we analyze the performance of different values of $\sigma_0$ around 1e-1, as presented in Table \ref{tab:lower_bound}. The optimal $\sigma_0$ always falls around 0.1, even for the Gaussian Mixture Model. A too-large $\sigma_0$, such as 0.2, during joint-training poses a risk of poor performance. However, this can be mitigated by employing a tiny $\sigma_0$ during post-training, resulting in performance improvements. Empirically, we find that when utilizing the Gaussian model, setting $\sigma_0$ within the range of 0.11 to 0.16 during joint-training yields superior performance after post-training. We also explore an appropriate lower bound for the scale parameter of the Laplacian model, as shown in Table \ref{tab:lower_bound}. Setting the value of the lower bound within the range of 0.08 to 0.15 yields superior performance for the Laplacian model.

\vspace{-0.5em}
\subsection{Zero-Center\hspace{-0.3mm} Quantization\hspace{-0.3mm} with\hspace{-0.3mm} Partial\hspace{-0.3mm} Stop-Gradient}
When incorporating SUA into zero-center quantization, both the gradients of the rate term and the distortion term are propagated to $\mu_q$, resulting in high variance. To mitigate this variance, we partially stop the gradient to the quantization offset. In the forward process, $\tilde{y}=r_{\alpha}\left(s_{\alpha}(y-\text{sg}(\mu_q))+u\right)+\text{sg}(\mu_q)$, where $\text{sg}(\cdot)$ represents the stop gradient operation. Then, the gradient of $\mu_q$ is only determined by the rate term. Although this reduction in variance introduces additional gradient bias, stopping the gradient from the distortion term contributes to improved performance, as illustrated in Table \ref{tab:suas}.

During post-training, directly using zero-center quantized latent to finetune entropy model and synthesis transform is not feasible because $\mu_q$ obtained from the entropy model also contributes to the reconstruction. To extend post-training for zero-center quantization, we also partially stop the gradient to $\mu_q$, allowing only the gradient from the rate term to pass through. 

\vspace{-0.3em}
\section{Experiments}
\vspace{-0.1em}
\subsection{Implementation}
\vspace{-0.1em}
\subsubsection{Image Compression Models}
We conducted experiments on a variety of representative image compression networks, including the mean-scale hyperprior model \cite{minnen2018joint} with different settings, the joint auto-regressive and hyperprior model \cite{minnen2018joint}, the model proposed in \cite{cheng2020learned}, the channel-wise autoregressive model \cite{minnen2020channel} with different settings, and ELIC model \cite{he2022elic}.

\label{para:mshyper}
\textbf{MS-hyper} represents the mean-scale hyperprior model (non-autoregressive) with nonzero-center quantization. The number of channels in latent space, $M$, and the transform, $N$, is set as $M=192$ and $N=128$.
\textbf{MS-hyper-zero} is based on MS-hyper, but uses zero-center quantization.

\textbf{MS-context} denotes the joint auto-regressive and hyperprior model \cite{minnen2018joint}. Due to the time-consuming nature of serial context modeling required for zero-center quantization in some quantization methods, such as MIX, we only consider nonzero-center quantization for MS-context. The setting is $M=192,~N=128$. 

\textbf{Cheng-anchor} represents the model proposed in \cite{cheng2020learned} using Gaussian Mixture, but without attention blocks. \textbf{Cheng-attn} incorporates both Gaussian Mixture and attention blocks. Similar to MS-context model, we only consider nonzero-center quantization. The setting is $M=N=192$ for the low-bitrate models and $M=N=192$ for the high-bitrate models.

\textbf{Charm} denotes the channel-wise autoregressive model \cite{minnen2020channel} with nonzero-center quantization. For simplicity, we remove the latent residual prediction module. The setting is $M=320,~N=192$. \textbf{Charm-zero} is based on Charm, but uses zero-center quantization.

\textbf{ELIC-zero} denotes the model proposed in \cite{he2022elic} with zero-center quantization. \textbf{ELIC-sm-zero} is a simplified version of ELIC-zero, which removes attention modules and uses fewer res-blocks. The setting is $M=320,~N=192$.
\subsubsection{Training}
All models were optimized for mean squared error (MSE). We trained multiple models with different values of $\lambda \in \{0.0018, 0.0035, 0.0067, 0.0130, 0.0250, 0.0483\}$. 
The training dataset is Flicker2W \cite{liu2020unified} dataset, which contains approximately 20,000 images. In each training iteration, images were randomly cropped into 256x256 patches. 

During joint-training, we applied the Adam optimizer \cite{kingma2014adam} for 450 epochs, with a batch size of 8 and an initial learning rate of 1e-4. After 400 epochs, the learning rate was reduced to 1e-5, and after 25 epochs, it was further reduced to 1e-6. During post-training, the Adam optimizer was applied for 100 epochs, with a batch size of 8 and a learning rate of 1e-4. The learning rate was reduced to 1e-5 after 50 epochs and to 1e-6 after 25 epochs. For all of our experiments, we utilize the identical random seed.

For SUA, we linearly annealed the parameter $\alpha$ of $s_{\alpha}$ and $r_{\alpha}$ from 1 to the maximum value $\alpha_m$ over 400 epochs during joint-training. The default maximum value was set as 12 in Sec. \ref{sec:analysis}.
 In Sec. \ref{sec:other}, the maximum value was selected from the set $\{4, 8, 12\}$ based on the performance after post-training.

For the setting of the lower bound, if not explicitly mentioned, the default value for $\sigma_0$ is set to 0.11 for Gaussian and Gaussian Mixture during joint-training. During post-training, $\sigma_0$ is always set as 1e-6. In the case of zero-center quantization, if not explicitly stated, partial stop-gradient for $\mu_q$ is used by default.

\vspace{-0.5em}
\subsubsection{Evaluation}
We evaluated these models using the Kodak dataset \cite{kodak} and the Tecnick dataset \cite{asuni2014testimages}. All BD-rate, which reflects the ratio of rate saving, \cite{bjontegaard2001calculation} were compared with BPG codec \cite{bpg}. The lower the BD-rate, the better the performance. The results in Sec. \ref{sec:analysis} 
were evaluated across four low-bitrate models, and the results in Sec. \ref{sec:other} were evaluated across six models.

\vspace{-0.7em}
\subsection{Experimental Results}
\label{sec:other}
We evaluate several commonly used quantization methods, such as AUN and MIX; several superior quantization methods based on UQ-s suggested in \cite{tsubota2023comprehensive} and our analyses; and our proposed SUA-based methods to assess their performance.

For the models we tested, setting an appropriate lower bound during joint-training always leads to a significant improvement in performance. For MS-context, as presented in Table \ref{tab:joint}, SUA-based methods outperform other methods by a significant margin. There is no obvious difference between the performance of PGE and STE, which implies the tolerance to bias and variance of gradient estimation of MS-context are similar. As shown in Table \ref{tab:cheng}, when training the Cheng-anchor model with AUN, we observe a severe effect of train-test mismatch at low bitrate, resulting in a performance drop. 
Applying post-training greatly enhances performance. It should be noted that training the Cheng-anchor model with the biased SUA-based method failed to converge for certain values of lambda. On the other hand, the unbiased SUA-based method exhibits improved performance compared to AUN and outperforms other methods. Similarly to Cheng-anchor, for Cheng-attn, training with the biased SUA-based method failed to converge, whereas the unbiased SUA-based method outperforms other methods. This indicates that both Cheng-anchor and Cheng-attn have a much lower tolerance for gradient bias.
Regarding Charm and Charm-zero, as shown in Table \ref{tab:charm}, SUA-based methods display significant improvements over other methods after joint training. The performance gain brought by SUA-based methods is much more significant on Charm compared to Charm-zero, since the train-test mismatch with zero-center quantization is smaller. 
For ELIC-sm-zero and ELIC-zero, as shown in Table \ref{tab:elic}, SUA with PGE is slightly better than MIX after post-training, while SUA with STE is slightly worse. For ELIC-zero, the $\alpha_m$ is set to 2 for a few high-bitrate models as they are unable to converge even with $\alpha_m=4$. The small $\alpha_m$ implies that ELIC has a much lower tolerance to gradient estimation risk than the aforementioned models.

\begin{table}[b]
    \centering
    \vspace{-0.3em}
    \caption{BD-rate (\%) of different quantization surrogates on MS-context. The anchor is BPG. The best performance of each column is underlined. SUA uses $\alpha_m = 8$.}
    \vspace{-0.7em}
    \begin{threeparttable}
    \tabcolsep=2.8pt
    \resizebox{1.03\linewidth}{!}{ 
    \begin{tabular}{ cc|c|cc|cc|cc } 
    \hline
    \multirow{2}{*}{rate} &  \multirow{2}{*}{distortion}&\multirow{2}{*}{$\sigma_0$\tnote{1}}& \multicolumn{2}{c|}{Kodak} &\multicolumn{2}{c|}{Tecnick}& \multicolumn{2}{c}{Avg}  \\
    &&&joint&post&joint&post&joint&post\\
    \hline
    AUN+PGE&AUN+PGE&1e-6&-2.26&-5.61&-7.11&-11.41&-4.69&-8.51\\
    AUN+PGE&AUN+PGE&0.11&-6.54&-12.73&-6.66&-13.22&-6.60&-12.98\\
    AUN+PGE&round+STE&0.11&-12.67&-13.49&-12.65&-14.29&-12.66&-13.89\\
    AUN+PGE&UQ-s+STE&0.11&-11.84&-13.69&-11.77&-14.74&-11.80&-14.21\\
    UQ-s+STE&UQ-s+STE&0.11&-11.67&-13.53&-11.72&-14.47&-11.70&-14.00\\
    SUA+EP&SUA+STE&0.11&\underline{-14.71}&\underline{-15.58}&-17.79&-18.70&-16.25&-17.14\\
    SUA+EP&SUA+PGE&0.11&-14.60&-15.47&\underline{-17.92}&\underline{-18.84}&\underline{-16.26}&\underline{-17.16}\\
    \hline
    \end{tabular}
    }
 \begin{tablenotes}
    \footnotesize
    \item[1] $\sigma_0$ here stands for the lower bound during joint-training. \\The value of the lower bound during post-training is always 1e-6.
\end{tablenotes} 
\end{threeparttable}
\label{tab:joint}
\end{table}

\begin{table*}[ht]
    \centering
        \caption{BD-rate (\%) of different quantization surrogates on Cheng-anchor and Cheng-attn. The anchor is BPG.  The mark '---' denotes that the training for some models can not converge with any $\alpha_{\text{m}}$ we tried. The best performance of each column is underlined. SUA uses $\alpha_m = 4$.}
    \vspace{-0.8em}
    \tabcolsep=5.5pt
    \resizebox{\linewidth}{!}{ 
    \begin{tabular}{ cc|c||cc|cc|cc||cc|cc|cc}
    \hline
    \multicolumn{3}{c||}{}&\multicolumn{6}{c||}{Cheng-anchor}&\multicolumn{6}{c}{Cheng-attn}\\
    \hline
    \multirow{2}{*}{rate} &  \multirow{2}{*}{distortion}&\multirow{2}{*}{$\sigma_0$}& \multicolumn{2}{c|}{Kodak} &\multicolumn{2}{c|}{Tecnick}& \multicolumn{2}{c||}{Avg} & \multicolumn{2}{c|}{Kodak} &\multicolumn{2}{c|}{Tecnick}& \multicolumn{2}{c}{Avg} \\

    &&&joint&post&joint&post&joint&post&joint&post&joint&post&joint&post\\
    \hline
    AUN+PGE&AUN+PGE&1e-6&-12.39&-22.58&-16.87&-28.99&-14.63&-25.79&-21.91&-24.35&-27.27&-31.45&-24.59&-27.90\\
    AUN+PGE&AUN+PGE&0.11&-15.18&-27.91&-20.00&-32.55&-17.59&-30.23&-26.83&-29.49&-30.92&-34.84&-28.87&-32.17\\
    AUN+PGE&round+STE&0.11&-25.34&-26.06&-28.90&-30.97&-27.12&-28.52&-19.58&-23.70&-19.60&-29.02&-19.59&-26.36\\
    AUN+PGE&UQ-s+STE&0.11&-25.67&-26.85&-28.82&-31.15&-27.24&-29.00&-27.01&-28.21&-30.60&-33.14&-28.80&-30.67\\
    UQ-s+STE&UQ-s+STE&0.11&-25.74&-26.88&-28.89&-31.45&-27.32&-29.17&-27.44&-28.49&-31.10&-33.70&-29.27&-31.09\\
    SUA+EP&SUA+STE&0.11&---&---&---&---&---&---&---&---&---&---&---&---\\
    SUA+EP&SUA+PGE&0.11&\underline{-27.72}&\underline{-29.03}&\underline{-33.18}&\underline{-34.74}&\underline{-30.45}&\underline{-31.89}&\underline{-29.22}&\underline{-30.19}&\underline{-34.55}&\underline{-35.83}&\underline{-31.89}&\underline{-33.01}\\
    \hline
    \end{tabular}
    }
    \label{tab:cheng}
\vspace{-0.5em}
\end{table*}

\begin{table*}[ht]
    \centering
        \caption{BD-rate (\%) of different quantization surrogates on Charm and Charm-zero. The anchor is BPG. The best performance of each column is underlined. SUA uses $\alpha_m = 8$.}
    \vspace{-0.8em}
    \tabcolsep=5.5pt
    \resizebox{\linewidth}{!}{ 
    \begin{tabular}{ cc|c||cc|cc|cc||cc|cc|cc } 
    \hline
    \multicolumn{3}{c||}{}&\multicolumn{6}{c||}{Charm}&\multicolumn{6}{c}{Charm-zero}\\
    \hline
    \multirow{2}{*}{rate} &  \multirow{2}{*}{distortion}&\multirow{2}{*}{$\sigma_0$}& \multicolumn{2}{c|}{Kodak} &\multicolumn{2}{c|}{Tecnick}& \multicolumn{2}{c||}{Avg} & \multicolumn{2}{c|}{Kodak} &\multicolumn{2}{c|}{Tecnick}& \multicolumn{2}{c}{Avg} \\

    &&&joint&post&joint&post&joint&post&joint&post&joint&post&joint&post\\
    \hline
    AUN+PGE&AUN+PGE&1e-6&-4.31&-10.51&2.02&-11.74&-1.14&-11.13&-5.92&-12.20&-3.66&-14.64&-4.79&-13.42\\
    AUN+PGE&AUN+PGE&0.11&-2.22&-15.45&3.24&-13.26&0.51&-14.35&-6.65&-18.16&-0.37&-17.76&-3.51&-17.96\\
    AUN+PGE&round+STE&0.11&-14.13&-15.68&-12.26&-14.34&-13.20&-15.01&-18.36&-18.92&\underline{-19.19}&-19.51&\underline{-18.77}&-19.22\\
    AUN+PGE&UQ-s+STE&0.11&-11.02&-14.81&-7.70&-12.69&-9.36&-13.75&-11.46&-15.51&-9.78&-13.89&-10.62&-14.70\\
    UQ-s+STE&UQ-s+STE&0.11&-10.86&-14.79&-7.18&-12.59&-9.02&-13.69&-12.47&-16.16&-11.50&-15.58&-11.99&-15.87\\
    SUA+EP&SUA+STE&0.11&\underline{-17.74}&\underline{-19.15}&\underline{-17.64}&\underline{-19.24}&\underline{-17.69}&\underline{-19.20}&\underline{-18.63}&\underline{-20.34}&-18.64&\underline{-20.65}&-18.63&\underline{-20.50}\\
    SUA+EP&SUA+PGE&0.11&-17.51&-19.05&-17.17&-19.19&-17.34&-19.12&-18.04&-19.72&-17.12&-19.32&-17.58&-19.52\\

    \hline
    \end{tabular}
    }
    \label{tab:charm}
\vspace{-0.5em}
\end{table*}

\begin{table*}
    \centering
        \caption{BD-rate (\%) of different quantization surrogates on ELIC-sm-zero and ELIC-zero. The anchor is BPG. The best performance of each column is underlined. SUA uses $\alpha_m = 4$ and $\alpha_m = 2$ ($\alpha_m=2$ for some high-bitrate models where $\alpha_m = 4$ cannot converge).}
    \vspace{-0.8em}
        \tabcolsep=5.5pt
    \resizebox{\linewidth}{!}{ 
    \begin{tabular}{ cc|c||cc|cc|cc||cc|cc|cc } 
    \hline
    \multicolumn{3}{c||}{}&\multicolumn{6}{c||}{ELIC-sm-zero}&\multicolumn{6}{c}{ELIC-zero}\\
    \hline
    \multirow{2}{*}{rate} &  \multirow{2}{*}{distortion}&\multirow{2}{*}{$\sigma_0$}& \multicolumn{2}{c|}{Kodak} &\multicolumn{2}{c|}{Tecnick}& \multicolumn{2}{c||}{Avg} & \multicolumn{2}{c|}{Kodak} &\multicolumn{2}{c|}{Tecnick}& \multicolumn{2}{c}{Avg} \\

    &&&joint&post&joint&post&joint&post&joint&post&joint&post&joint&post\\
    \hline
    AUN+PGE&AUN+PGE&0.11&-13.70&-26.39&-7.06&-29.02&-10.38&-27.70&-16.15&-28.26&-8.13&-31.94&-12.14&-30.10\\
    AUN+PGE&round+STE&0.11&\underline{-25.61}&-26.23&\underline{-28.85}&-29.41&\underline{-27.23}&-27.82&\underline{-28.37}&-28.89&\underline{-34.09}&\underline{-34.58}&\underline{-31.23}&-31.74\\
    SUA+EP&SUA+STE&0.11&-20.94&-25.80&-19.60&-27.63&-20.27&-26.72&-26.52&-30.05&-27.60&-34.23&-27.06&-32.14\\
    SUA+EP&SUA+PGE&0.11&-22.35&\underline{-27.11}&-19.50&\underline{-29.51}&-20.92&\underline{-28.31}&-25.61&\underline{-30.49}&-24.74&-34.51&-25.18&\underline{-32.50}\\

    \hline
    \end{tabular}
    }
\vspace{-0.5em}
    \label{tab:elic}
\end{table*}

\vspace{-0.8em}
\section{Conclusion}
In this paper, we present systematic theoretical analyses for evaluating quantization surrogates. In our study, the quantization surrogate is decomposed into forward and backward calculations. We examine the train-test mismatch and gradient estimation risk of existing methods, respectively. Our analyses show that there is a tradeoff between train-test mismatch and the gradient estimation risk, which varies across different network structures. To flexibly control this tradeoff, we introduce stochastic uniform annealing-based methods.
In addition, we highlight two subtle tricks that greatly affect performance: the lower bound for the estimated variance parameter and zero-center quantization. We propose setting appropriate lower bounds for the estimated distribution to reduce train-test mismatch. For zero-center quantization, we propose a partial stop-gradient method to reduce gradient estimation variance when annealing-based methods are applied.
 Our method with the tricks
outperforms the existing quantization surrogates on a variety of representative image compression models.
\vspace{-0.5em}
\ifCLASSOPTIONcaptionsoff
  \newpage
\fi

\bibliographystyle{IEEEtran}
\bibliography{main}

\begin{thebibliography}{10}
\providecommand{\url}[1]{#1}
\csname url@samestyle\endcsname
\providecommand{\newblock}{\relax}
\providecommand{\bibinfo}[2]{#2}
\providecommand{\BIBentrySTDinterwordspacing}{\spaceskip=0pt\relax}
\providecommand{\BIBentryALTinterwordstretchfactor}{4}
\providecommand{\BIBentryALTinterwordspacing}{\spaceskip=\fontdimen2\font plus
\BIBentryALTinterwordstretchfactor\fontdimen3\font minus \fontdimen4\font\relax}
\providecommand{\BIBforeignlanguage}[2]{{%
\expandafter\ifx\csname l@#1\endcsname\relax
\typeout{** WARNING: IEEEtran.bst: No hyphenation pattern has been}%
\typeout{** loaded for the language `#1'. Using the pattern for}%
\typeout{** the default language instead.}%
\else
\language=\csname l@#1\endcsname
\fi
#2}}
\providecommand{\BIBdecl}{\relax}
\BIBdecl

\bibitem{wallace1992jpeg}
G.~K. Wallace, ``The jpeg still picture compression standard,'' \emph{IEEE transactions on consumer electronics}, vol.~38, no.~1, pp. xviii--xxxiv, 1992.

\bibitem{skodras2001jpeg}
A.~Skodras, C.~Christopoulos, and T.~Ebrahimi, ``The jpeg 2000 still image compression standard,'' \emph{IEEE Signal processing magazine}, vol.~18, no.~5, pp. 36--58, 2001.

\bibitem{bpg}
F.~Bellard, ``Bpg image format,'' \url{http://bellard.org/bpg/}, 2014.

\bibitem{balle2016end}
J.~Ball{\'e}, V.~Laparra, and E.~P. Simoncelli, ``End-to-end optimized image compression,'' \emph{arXiv preprint arXiv:1611.01704}, 2016.

\bibitem{agustsson2020universally}
E.~Agustsson and L.~Theis, ``Universally quantized neural compression,'' \emph{Advances in neural information processing systems}, vol.~33, pp. 12\,367--12\,376, 2020.

\bibitem{theis2022lossy}
L.~Theis, T.~Salimans, M.~D. Hoffman, and F.~Mentzer, ``Lossy compression with gaussian diffusion,'' \emph{arXiv preprint arXiv:2206.08889}, 2022.

\bibitem{guo2021soft}
Z.~Guo, Z.~Zhang, R.~Feng, and Z.~Chen, ``Soft then hard: Rethinking the quantization in neural image compression,'' in \emph{International Conference on Machine Learning}.\hskip 1em plus 0.5em minus 0.4em\relax PMLR, 2021, pp. 3920--3929.

\bibitem{theis2017lossy}
L.~Theis, W.~Shi, A.~Cunningham, and F.~Husz{\'a}r, ``Lossy image compression with compressive autoencoders,'' \emph{arXiv preprint arXiv:1703.00395}, 2017.

\bibitem{bengio2013estimating}
Y.~Bengio, N.~L{\'e}onard, and A.~Courville, ``Estimating or propagating gradients through stochastic neurons for conditional computation,'' \emph{arXiv preprint arXiv:1308.3432}, 2013.

\bibitem{minnen2020channel}
D.~Minnen and S.~Singh, ``Channel-wise autoregressive entropy models for learned image compression,'' in \emph{2020 IEEE International Conference on Image Processing (ICIP)}.\hskip 1em plus 0.5em minus 0.4em\relax IEEE, 2020, pp. 3339--3343.

\bibitem{tsubota2023comprehensive}
K.~Tsubota and K.~Aizawa, ``Comprehensive comparisons of uniform quantization in deep image compression,'' \emph{IEEE Access}, 2023.

\bibitem{mohamed2020monte}
S.~Mohamed, M.~Rosca, M.~Figurnov, and A.~Mnih, ``Monte carlo gradient estimation in machine learning,'' \emph{The Journal of Machine Learning Research}, vol.~21, no.~1, pp. 5183--5244, 2020.

\bibitem{yang2020improving}
Y.~Yang, R.~Bamler, and S.~Mandt, ``Improving inference for neural image compression,'' \emph{Advances in Neural Information Processing Systems}, vol.~33, pp. 573--584, 2020.

\bibitem{agustsson2017soft}
E.~Agustsson, F.~Mentzer, M.~Tschannen, L.~Cavigelli, R.~Timofte, L.~Benini, and L.~V. Gool, ``Soft-to-hard vector quantization for end-to-end learning compressible representations,'' \emph{Advances in neural information processing systems}, vol.~30, 2017.

\bibitem{ziv1985universal}
J.~Ziv, ``On universal quantization,'' \emph{IEEE Transactions on Information Theory}, vol.~31, no.~3, pp. 344--347, 1985.

\bibitem{choi2019variable}
Y.~Choi, M.~El-Khamy, and J.~Lee, ``Variable rate deep image compression with a conditional autoencoder,'' in \emph{Proceedings of the IEEE/CVF International Conference on Computer Vision}, 2019, pp. 3146--3154.

\bibitem{kingma2013auto}
D.~P. Kingma and M.~Welling, ``Auto-encoding variational bayes,'' \emph{arXiv preprint arXiv:1312.6114}, 2013.

\bibitem{toderici2015variable}
G.~Toderici, S.~M. O'Malley, S.~J. Hwang, D.~Vincent, D.~Minnen, S.~Baluja, M.~Covell, and R.~Sukthankar, ``Variable rate image compression with recurrent neural networks,'' \emph{arXiv preprint arXiv:1511.06085}, 2015.

\bibitem{toderici2017full}
G.~Toderici, D.~Vincent, N.~Johnston, S.~J. Hwang, D.~Minnen, J.~Shor, and M.~Covell, ``Full resolution image compression with recurrent neural networks,'' in \emph{CVPR}, 2017, pp. 5306--5314.

\bibitem{johnston2018improved}
N.~Johnston, D.~Vincent, D.~Minnen, M.~Covell, S.~Singh, T.~Chinen, S.~J. Hwang, J.~Shor, and G.~Toderici, ``Improved lossy image compression with priming and spatially adaptive bit rates for recurrent networks,'' in \emph{Proceedings of the IEEE Conference on Computer Vision and Pattern Recognition}, 2018, pp. 4385--4393.

\bibitem{balle2018variational}
J.~Ball{\'e}, D.~Minnen, S.~Singh, S.~J. Hwang, and N.~Johnston, ``Variational image compression with a scale hyperprior,'' \emph{arXiv preprint arXiv:1802.01436}, 2018.

\bibitem{minnen2018joint}
D.~Minnen, J.~Ball{\'e}, and G.~D. Toderici, ``Joint autoregressive and hierarchical priors for learned image compression,'' \emph{Advances in neural information processing systems}, vol.~31, 2018.

\bibitem{cheng2020learned}
Z.~Cheng, H.~Sun, M.~Takeuchi, and J.~Katto, ``Learned image compression with discretized gaussian mixture likelihoods and attention modules,'' in \emph{Proceedings of the IEEE/CVF Conference on Computer Vision and Pattern Recognition}, 2020, pp. 7939--7948.

\bibitem{He_2021_CVPR}
D.~He, Y.~Zheng, B.~Sun, Y.~Wang, and H.~Qin, ``Checkerboard context model for efficient learned image compression,'' in \emph{Proceedings of the IEEE/CVF Conference on Computer Vision and Pattern Recognition (CVPR)}, June 2021, pp. 14\,771--14\,780.

\bibitem{qian2022entroformer}
Y.~Qian, M.~Lin, X.~Sun, Z.~Tan, and R.~Jin, ``Entroformer: A transformer-based entropy model for learned image compression,'' \emph{arXiv preprint arXiv:2202.05492}, 2022.

\bibitem{lee2018context}
J.~Lee, S.~Cho, and S.-K. Beack, ``Context-adaptive entropy model for end-to-end optimized image compression,'' \emph{arXiv preprint arXiv:1809.10452}, 2018.

\bibitem{chen2021end}
T.~Chen, H.~Liu, Z.~Ma, Q.~Shen, X.~Cao, and Y.~Wang, ``End-to-end learnt image compression via non-local attention optimization and improved context modeling,'' \emph{IEEE Transactions on Image Processing}, vol.~30, pp. 3179--3191, 2021.

\bibitem{guo2021causal}
Z.~Guo, Z.~Zhang, R.~Feng, and Z.~Chen, ``Causal contextual prediction for learned image compression,'' \emph{IEEE Transactions on Circuits and Systems for Video Technology}, vol.~32, no.~4, pp. 2329--2341, 2021.

\bibitem{hu2021learning}
Y.~Hu, W.~Yang, Z.~Ma, and J.~Liu, ``Learning end-to-end lossy image compression: A benchmark,'' \emph{IEEE Transactions on Pattern Analysis and Machine Intelligence}, vol.~44, no.~8, pp. 4194--4211, 2021.

\bibitem{he2022elic}
D.~He, Z.~Yang, W.~Peng, R.~Ma, H.~Qin, and Y.~Wang, ``Elic: Efficient learned image compression with unevenly grouped space-channel contextual adaptive coding,'' in \emph{Proceedings of the IEEE/CVF Conference on Computer Vision and Pattern Recognition}, 2022, pp. 5718--5727.

\bibitem{lin2020spatial}
C.~Lin, J.~Yao, F.~Chen, and L.~Wang, ``A spatial rnn codec for end-to-end image compression,'' in \emph{Proceedings of the IEEE/CVF Conference on Computer Vision and Pattern Recognition}, 2020, pp. 13\,269--13\,277.

\bibitem{zou2022devil}
R.~Zou, C.~Song, and Z.~Zhang, ``The devil is in the details: Window-based attention for image compression,'' in \emph{Proceedings of the IEEE/CVF Conference on Computer Vision and Pattern Recognition}, 2022, pp. 17\,492--17\,501.

\bibitem{zhu2022transformer}
Y.~Zhu, Y.~Yang, and T.~Cohen, ``Transformer-based transform coding,'' in \emph{International Conference on Learning Representations}, 2022.

\bibitem{ma2020end}
H.~Ma, D.~Liu, N.~Yan, H.~Li, and F.~Wu, ``End-to-end optimized versatile image compression with wavelet-like transform,'' \emph{IEEE Transactions on Pattern Analysis and Machine Intelligence}, vol.~44, no.~3, pp. 1247--1263, 2020.

\bibitem{helminger2020lossy}
L.~Helminger, A.~Djelouah, M.~Gross, and C.~Schroers, ``Lossy image compression with normalizing flows,'' \emph{arXiv preprint arXiv:2008.10486}, 2020.

\bibitem{xie2021enhanced}
Y.~Xie, K.~L. Cheng, and Q.~Chen, ``Enhanced invertible encoding for learned image compression,'' in \emph{Proceedings of the 29th ACM international conference on multimedia}, 2021, pp. 162--170.

\bibitem{10008823}
H.~Sun, L.~Yu, and J.~Katto, ``Improving latent quantization of learned image compression with gradient scaling,'' in \emph{2022 IEEE International Conference on Visual Communications and Image Processing (VCIP)}, 2022, pp. 1--5.

\bibitem{cai2018deep}
J.~Cai and L.~Zhang, ``Deep image compression with iterative non-uniform quantization,'' in \emph{2018 25th IEEE International Conference on Image Processing (ICIP)}.\hskip 1em plus 0.5em minus 0.4em\relax IEEE, 2018, pp. 451--455.

\bibitem{mentzer2018conditional}
F.~Mentzer, E.~Agustsson, M.~Tschannen, R.~Timofte, and L.~Van~Gool, ``Conditional probability models for deep image compression,'' in \emph{Proceedings of the IEEE Conference on Computer Vision and Pattern Recognition}, 2018, pp. 4394--4402.

\bibitem{zhong2020channel}
Z.~Zhong, H.~Akutsu, and K.~Aizawa, ``Channel-level variable quantization network for deep image compression,'' \emph{arXiv preprint arXiv:2007.12619}, 2020.

\bibitem{zhu2022unified}
X.~Zhu, J.~Song, L.~Gao, F.~Zheng, and H.~T. Shen, ``Unified multivariate gaussian mixture for efficient neural image compression,'' in \emph{Proceedings of the IEEE/CVF Conference on Computer Vision and Pattern Recognition}, 2022, pp. 17\,612--17\,621.

\bibitem{feng2023nvtc}
R.~Feng, Z.~Guo, W.~Li, and Z.~Chen, ``Nvtc: Nonlinear vector transform coding,'' in \emph{Proceedings of the IEEE/CVF Conference on Computer Vision and Pattern Recognition}, 2023, pp. 6101--6110.

\bibitem{zhang2023lvqac}
X.~Zhang and X.~Wu, ``Lvqac: Lattice vector quantization coupled with spatially adaptive companding for efficient learned image compression,'' \emph{arXiv preprint arXiv:2304.12319}, 2023.

\bibitem{theis2022algorithms}
L.~Theis and N.~Yosri, ``Algorithms for the communication of samples,'' 2022.

\bibitem{shannon1948mathematical}
C.~E. Shannon, ``A mathematical theory of communication,'' \emph{The Bell system technical journal}, vol.~27, no.~3, pp. 379--423, 1948.

\bibitem{shannon1959coding}
C.~E. Shannon \emph{et~al.}, ``Coding theorems for a discrete source with a fidelity criterion,'' \emph{IRE Nat. Conv. Rec}, vol.~4, no. 142-163, p.~1, 1959.

\bibitem{blahut1972computation}
R.~Blahut, ``Computation of channel capacity and rate-distortion functions,'' \emph{IEEE transactions on Information Theory}, vol.~18, no.~4, pp. 460--473, 1972.

\bibitem{balle2020nonlinear}
J.~Ball{\'e}, P.~A. Chou, D.~Minnen, S.~Singh, N.~Johnston, E.~Agustsson, S.~J. Hwang, and G.~Toderici, ``Nonlinear transform coding,'' \emph{IEEE Journal of Selected Topics in Signal Processing}, vol.~15, no.~2, pp. 339--353, 2020.

\bibitem{jang2016categorical}
E.~Jang, S.~Gu, and B.~Poole, ``Categorical reparameterization with gumbel-softmax,'' \emph{arXiv preprint arXiv:1611.01144}, 2016.

\bibitem{williams1992simple}
R.~J. Williams, ``Simple statistical gradient-following algorithms for connectionist reinforcement learning,'' \emph{Reinforcement learning}, pp. 5--32, 1992.

\bibitem{ghosh2023gradient}
A.~Ghosh, Y.~H. Liu, G.~Lajoie, K.~Kording, and B.~A. Richards, ``How gradient estimator variance and bias impact learning in neural networks,'' in \emph{The Eleventh International Conference on Learning Representations}, 2023.

\bibitem{deng2012mnist}
L.~Deng, ``The mnist database of handwritten digit images for machine learning research [best of the web],'' \emph{IEEE signal processing magazine}, vol.~29, no.~6, pp. 141--142, 2012.

\bibitem{liu2023learned}
J.~Liu, H.~Sun, and J.~Katto, ``Learned image compression with mixed transformer-cnn architectures,'' in \emph{Proceedings of the IEEE/CVF Conference on Computer Vision and Pattern Recognition}, 2023, pp. 14\,388--14\,397.

\bibitem{liu2020unified}
J.~Liu, G.~Lu, Z.~Hu, and D.~Xu, ``A unified end-to-end framework for efficient deep image compression,'' \emph{arXiv preprint arXiv:2002.03370}, 2020.

\bibitem{kingma2014adam}
D.~P. Kingma and J.~Ba, ``Adam: A method for stochastic optimization,'' \emph{arXiv preprint arXiv:1412.6980}, 2014.

\bibitem{kodak}
K.~Eastman, ``Kodak lossless true color image suite (photocd pcd0992),'' \url{http://r0k.us/graphics/kodak/}, 1993.

\bibitem{asuni2014testimages}
N.~Asuni and A.~Giachetti, ``Testimages: a large-scale archive for testing visual devices and basic image processing algorithms.'' in \emph{STAG}, 2014, pp. 63--70.

\bibitem{bjontegaard2001calculation}
G.~Bjontegaard, ``Calculation of average psnr differences between rd-curves,'' \emph{ITU SG16 Doc. VCEG-M33}, 2001.

\end{thebibliography}


\begin{thebibliography}{1}
\providecommand{\url}[1]{#1}
\csname url@samestyle\endcsname
\providecommand{\newblock}{\relax}
\providecommand{\bibinfo}[2]{#2}
\providecommand{\BIBentrySTDinterwordspacing}{\spaceskip=0pt\relax}
\providecommand{\BIBentryALTinterwordstretchfactor}{4}
\providecommand{\BIBentryALTinterwordspacing}{\spaceskip=\fontdimen2\font plus
\BIBentryALTinterwordstretchfactor\fontdimen3\font minus \fontdimen4\font\relax}
\providecommand{\BIBforeignlanguage}[2]{{%
\expandafter\ifx\csname l@#1\endcsname\relax
\typeout{** WARNING: IEEEtran.bst: No hyphenation pattern has been}%
\typeout{** loaded for the language `#1'. Using the pattern for}%
\typeout{** the default language instead.}%
\else
\language=\csname l@#1\endcsname
\fi
#2}}
\providecommand{\BIBdecl}{\relax}
\BIBdecl

\bibitem{begaint2020compressai}
J.~B{\'e}gaint, F.~Racap{\'e}, S.~Feltman, and A.~Pushparaja, ``Compressai: a pytorch library and evaluation platform for end-to-end compression research,'' \emph{arXiv preprint arXiv:2011.03029}, 2020.

\end{thebibliography}

\end{document}


\maketitle
\section{\large Derivation of Eq. (3) in Sec. 3.2.}
The mutual information $I(X;\hat{Y})$ and $I(Y;\hat{Y})$ can be represented as 
\begin{equation}
\begin{aligned}
   I(X;\hat{Y})=H(\hat{Y})-H(\hat{Y}|X),\\
   I(Y;\hat{Y})=H(\hat{Y})-H(\hat{Y}|Y).
\end{aligned}
\end{equation}
With the assumption that the analysis transform is a deterministic function, we have
\begin{equation}
\begin{aligned}
H(\hat{Y}|X)=H(\hat{Y}|g_a(X),X)=H(\hat{Y}|g_a(X))=H(\hat{Y}|Y).
\end{aligned}
\end{equation}
Since $H(\hat{Y}|X)=H(\hat{Y}|Y)$, then $I(X;\hat{Y})=I(Y;\hat{Y})$.
If the quantizer is rounding, which is a deterministic function, then $H(\hat{Y}|X)=H(\hat{Y}|Y)=0$. Thus, 
\begin{equation}
\begin{aligned}
I(X;\hat{Y})=I(Y;\hat{Y})=H(\hat{Y})
\end{aligned}
\end{equation}
\section{\large Visualization of the Soft Function in Sec. 4.1.1.}
\begin{figure}[h]
  \centering
\includegraphics[width=0.8\linewidth]{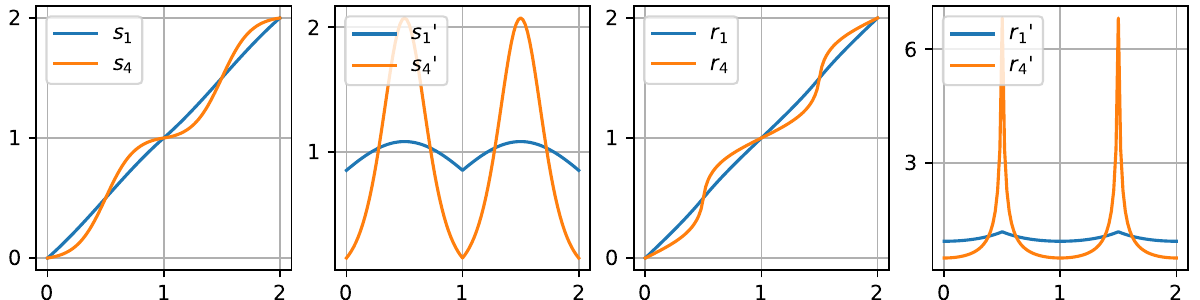}
\caption{
Visualization of function and derivative of $s_{\alpha}$ and $r_{\alpha}$.
}
\end{figure}
\section{\large Derivation of Eq. (21) in Sec. 4.1.2.}
With the assumption that $y$ is a scalar and the forward calculation is adding uniform noise, we have
\begin{equation}
\begin{aligned}
\mathbb{E}_{p_U} L(\tilde{y})=\int_{u}L(y+u)du=\int_{y-0.5}^{y+0.5}L(t)dt
\end{aligned}
\end{equation}
Then, the gradient of the expected loss function is 
\begin{equation}
\begin{aligned}
       &\frac{\partial \mathbb{E}_{p_U} L(\tilde{y})}{\partial y}= \frac{\partial \int_{y-0.5}^{y+0.5}L(t)dt}{\partial y}=L(y+0.5)-L(y-0.5).
\end{aligned}
\end{equation}
\section{\large Derivation of the Mutual Information of UQ-s in Sec. 4.2.2.}
We consider a two-dimensional Gaussian source $X$, $X\sim \mathcal{N}(0,\Sigma),~Y=\sigma X,~Y\sim\mathcal{N}(0,\sigma^2 \Sigma),~Y=[Y_1,Y_2]$,
 \begin{equation}
 \Sigma=
 \left[
 \begin{array}{cc}
     1 & \rho \\
     \rho & 1 
 \end{array}
 \right],    
  \sigma^2 \Sigma=
 \left[
 \begin{array}{cc}
     \sigma^2 & \sigma^2\rho \\
     \sigma^2\rho & \sigma^2
 \end{array}
 \right].
 \end{equation}
 We then study the situation where $\rho=1$, $Y_1=Y_2$. Note that $\tilde{Y}_1=Y_1+U_1,~\tilde{Y}_2=Y_2+U_2$, and $U_1$ is independent from $U_2$. Obviously, rounding has $I(Y;\lfloor Y\rfloor)=I(Y_1;\lfloor Y_1 \rfloor)$; UQ-s has $I(Y;\tilde{Y})=I(Y_1; \tilde{Y_1})$. For UQ-i and AUN,
\begin{equation}
\begin{aligned}
&I(Y_1,Y_2;\tilde{Y}_1,\tilde{Y}_2)\\
=&H(Y_1,Y_2)+H(\tilde{Y}_1,\tilde{Y}_2)-H(Y_1,Y_2,\tilde{Y}_1,\tilde{Y}_2)\\
=&H(Y_1)+H(Y_2|Y_1)+H(\tilde{Y}_1)+H(\tilde{Y}_2|\tilde{Y}_1)-H(Y_1,\tilde{Y}_1)-H(Y_2,\tilde{Y}_2|Y_1,\tilde{Y}_1)\\
\end{aligned}
\end{equation}
Since $H(Y_2|Y_1)=0,~H(\tilde{Y}_1)=H(\tilde{Y}_2),~H(\tilde{Y}_1|Y_1)=H(\tilde{Y}_2|Y_2),~H(Y_1,\tilde{Y}_1)=H(\tilde{Y}_1|Y_1)+H(Y_1),~H(Y_2,\tilde{Y}_2|Y_1,\tilde{Y}_1)=H(\tilde{Y}_1|Y_1)$, then we have 
\begin{equation}
\begin{aligned}
&I(Y_1,Y_2;\tilde{Y}_1,\tilde{Y}_2)\\
=&H(\tilde{Y}_1)+H(\tilde{Y}_2|\tilde{Y}_1)-2H(\tilde{Y}_1|Y_1)\\
=&H(\tilde{Y}_1)-H(\tilde{Y}_1|Y_1)+H(\tilde{Y}_2)-H(\tilde{Y}_2|Y_2)-(H(\tilde{Y}_2)-H(\tilde{Y}_2|\tilde{Y}_1))\\
=& 2I(Y_1;\tilde{Y}_1)-I(\tilde{Y}_2;\tilde{Y}_1)
\end{aligned}
\end{equation}
Futhermore, since $H(\tilde{Y}_1|Y_1)\leq H(\tilde{Y}_2|\tilde{Y}_1)$, then we have $2I(Y_1;\tilde{Y}_1)-I(\tilde{Y}_2;\tilde{Y}_1)\geq I(Y_1;\tilde{Y}_1)$.
 
\section{\large Calculation of the Estimated Rate of UQ-s in Sec. 4.3.}
We set $Y$ as a two-dimensional Gaussian variable, $Y\sim\mathcal{N}(0,\Sigma_p),~Y=[Y_1,Y_2]$. The estimated distribution is also a two-dimensional Gaussian $\mathcal{N}(0,\Sigma_q)$.
 \begin{equation}
 \Sigma_p=
 \left[
 \begin{array}{cc}
     1 & \rho_p \\
     \rho_p & 1 
 \end{array}
 \right],    
\Sigma_q=
 \left[
 \begin{array}{cc}
     1 & \rho_q \\
     \rho_q & 1 
 \end{array}
 \right].
 \end{equation}
We then calculate the rate through the conditional probability formula.
First, we calculate the rate of $Y_1$ using its marginal distribution, $Y_1\sim\mathcal{N}(0,1)$. Then, we calculate the rate of $Y_2$ using the distribution conditioned on $y_1$. Here, we conditioned on $y_1$ instead of $\tilde{y}_1$ for simpler calculation. The conditional distribution is $\mathcal{N}(\rho_q y_1, 1-\rho_q^2)$. The rate is the average rate of $Y_1$ and $Y_2$.
\section{\large Derivation of the Expected Gradient in Sec. 4.4.1.}
\subsection{SUA}
We assume $y\in\mathbb{R}^{n},~u\sim\mathcal{U}(-0.5,0.5)^n$. The expected loss of using SUA is
\begin{equation}\label{eq_univesalsoft}
\begin{aligned}
\mathbb{E}_{p_{U}}L(\tilde{y})&=\int_{u} L(\tilde{y})du\\
&=\int_{u_i}\int_{u_{j\neq i}}L(\tilde{y_i},\tilde{y}_{j\neq i})du_{j\neq i}du_i\\
&=\int_{u_i}f(\tilde{y_i})du_i\\
\text{where }& f(\tilde{y_i})=\int_{u_{j\neq i}}L(\tilde{y_i},\tilde{y}_{j\neq i})du_{j\neq i}.
\end{aligned}
\end{equation}
Since $\tilde{y}_i=r_{\alpha}(s_{\alpha}(y_i)+u_i)$, the expected gradient is
\begin{equation}\label{eq_univesalsoft}
\begin{aligned}
\frac{\partial \mathbb{E}_{p_{U}}L(\tilde{y})}{\partial y_i}&=\frac{\partial \int_{u_i}f(\tilde{y_i})du_i}{\partial y_i}\\
&=\frac{\partial \int_{s_{\alpha}(y_i)-0.5}^{s_{\alpha}(y_i)+0.5}f(r_{\alpha}(t))dt}{\partial y_i}\\
&=\frac{\partial \int_{s_{\alpha}(y_i)-0.5}^{s_{\alpha}(y_i)+0.5}f(r_{\alpha}(t))dt}{\partial s_{\alpha}(y_i)}\frac{\partial s_{\alpha}(y_i)}{\partial y_i}\\
&=\left[f(r_{\alpha}(s_{\alpha}(y_i)+0.5))-f(r_{\alpha}(s_{\alpha}(y_i)-0.5))\right]\frac{\partial s_{\alpha}(y_i)}{\partial y_i}\\
&=\left[f(\tilde{y_i})|_{u_i=0.5}-f(\tilde{y_i})|_{u_i=-0.5}\right]\frac{\partial s_{\alpha}(y_i)}{\partial y_i}
\end{aligned}
\end{equation}
For rate term, $R(\tilde{y})=\sum^{n}_{i=1}R(\tilde{y}_i)$, we can directly calculate the expected gradient,
\begin{equation}\label{eq_suarep}
\begin{aligned}
\frac{\partial \mathbb{E}_{p_{U}}R(\tilde{y})}{\partial y_i}&=\frac{\partial \mathbb{E}_{p_{U_i}}R(\tilde{y}_i)}{\partial y_i}\\
&=\frac{\partial s_{\alpha}(y_i)}{\partial y_i}[R(\tilde{y}_i)|_{u_i=0.5}-R(\tilde{y}_i)|_{u_i=-0.5}].
\end{aligned}
\end{equation}
\subsection{SRA}
The expected objective function is
\begin{equation}\label{eq_univesalsoft}
\begin{aligned}
\mathbb{E}_{p_{\tilde{Y}}}L(\tilde{y})&=\sum_{\tilde{y}} L(\tilde{y})p_{\tilde{Y}}(\tilde{y})\\
&=\sum_{\tilde{y}_i}p_{\tilde{Y}_{i}}(\tilde{y}_i)\sum_{\tilde{y}_{j\neq i}}L(\tilde{y_i},\tilde{y}_{j\neq i})p_{\tilde{Y}_{j\neq i}}(\tilde{y}_{j\neq i})\\
&=\sum_{\tilde{y}_i}p_{\tilde{Y}_{i}}(\tilde{y}_i)f(\tilde{y}_i)\\
\text{where }&f(\tilde{y}_i)=\sum_{\tilde{y}_{j\neq i}}L(\tilde{y_i},\tilde{y}_{j\neq i})p_{\tilde{Y}_{j\neq i}}(\tilde{y}_{j\neq i}).\\
\end{aligned}
\end{equation}
Since for SRA,
\begin{equation}\label{eq_univesalsoft}
\begin{aligned}
       & P(\tilde{y_i}=\lfloor y_i \rfloor)= \lceil s_{\alpha}(y_i) \rceil -s_{\alpha}(y_i)=\lceil y_i \rceil -s_{\alpha}(y_i),\\
       & P(\tilde{y_i}=\lceil y_i \rceil)= s_{\alpha}(y_i)-\lfloor s_{\alpha}(y_i) \rfloor=s_{\alpha}(y_i)-\lfloor y_i \rfloor.\\
\end{aligned}
\end{equation}
Then, the expected gradient is 
\begin{equation}\label{eq_univesalsoft}
\begin{aligned}
\frac{\partial \mathbb{E}_{p_{\tilde{Y}}}L(\tilde{y})}{\partial y_i}&=\sum_{\tilde{y}_i} \frac{\partial p_{\tilde{Y}_{i}}(\tilde{y}_i)}{\partial y_i}f(\tilde{y}_i)\\
&=\frac{\partial P(\tilde{y_i}=\lfloor y_i \rfloor)}{\partial y_i}f(\lfloor y_i \rfloor)+\frac{\partial P(\tilde{y_i}=\lceil y_i \rceil)}{\partial y_i}f(\lceil y_i \rceil)\\
&=[f(\lceil y_i \rceil)-f(\lfloor y_i \rfloor)]\frac{\partial s_{\alpha}(y_i)}{\partial y_i}
\end{aligned}
\end{equation}
For rate term, the expected gradient can be easily computed, 
\begin{equation}\label{eq_univesalsoft}
\begin{aligned}
\frac{\partial \mathbb{E}_{p_{\tilde{Y}}}R(\tilde{y})}{\partial y_i}=\frac{\partial s_{\alpha}(y_i)}{\partial y_i}[R(\lceil y_i \rceil)-R(\lfloor y_i \rfloor)]
\end{aligned}
\end{equation}
\section{\large Derivation of Eq. (32) and (33) in Sec. 4.4.2.}
\subsection{SUA}
When applying the generalized STE to SUA, which views $r_{\alpha}$ as an identity function in the backward pass, we have 
\begin{equation}\label{eq_univesalsoft}
\begin{aligned}
\frac{\partial r_{\alpha}(s_{\alpha}(y_i)+u_i)}{\partial y_i}&=\frac{\partial r_{\alpha}(s_{\alpha}(y_i)+u_i)}{\partial s_{\alpha}(y_i)} \frac{\partial s_{\alpha}(y_i)}{\partial y_i},\\
&\approx \frac{\partial s_{\alpha}(y_i)}{\partial y_i}.
\end{aligned}
\end{equation}
Then the gradient can be estimated through  
\begin{equation}\label{eq_suaste}
\begin{aligned}
\frac{\partial \mathbb{E}_{p_{U}}L(\tilde{y})}{\partial y_i}
&\approx\left[\mathbb{E}_{p_{U}} \frac{\partial L(\tilde{y})}{\partial \tilde{y}_i} \right]\frac{\partial s_{\alpha}(y_i)}{\partial y_i}. \\
\end{aligned}
\end{equation}
\subsection{SRA}
When applying STE to SRA, which views $\lfloor\cdot\rfloor$ and $\lceil\cdot\rceil$ as identity functions in the backward pass, we have
\begin{equation}\label{eq_univesalsoft}
\begin{aligned}
\frac{\partial \lfloor y_i\rfloor}{\partial y_i}&\approx 1,~\frac{\partial \lceil y_i\rceil}{\partial y_i}&\approx 1.
\end{aligned}
\end{equation}
Then the gradient can be estimated through  
\begin{equation}\label{eq_srabiased}
\begin{aligned}
\frac{\partial \mathbb{E}_{p_{\tilde{Y}}}L(\tilde{y})}{\partial y_i}&=\mathbb{E}_{p_{\tilde{Y}}}\left[\frac{\partial L(\tilde{y})}{\partial \tilde{y}_i}\frac{\partial \tilde{y}_i}{\partial s_{\alpha}(y_i)}\frac{\partial s_{\alpha}(y_i)}{\partial y_i} \right]\\
&\approx \mathbb{E}_{p_{\tilde{Y}}}\left[\frac{\partial L(\tilde{y})}{\partial \tilde{y}_i}\frac{\partial s_{\alpha}(y_i)}{\partial y_i} \right]\\
&\approx\left[\sum_{\tilde{y}} \frac{\partial L(\tilde{y})}{\partial \tilde{y}_i} p_{\tilde{Y}}(\tilde{y}) \right]\frac{\partial s_{\alpha}(y_i)}{\partial y_i}. \\
\end{aligned}
\end{equation}
\section{\large Experiment Setting on MNIST in Sec. 4.4.2.}
The structure of the analysis and synthesis transform, which are used to illustrate the effect of network complexity on gradient bias, is shown below. We adjust the size of networks by changing N.
\begin{figure}[ht]
    \centering
    \subfloat[analysis transform]{ 
    \includegraphics[width=0.45\linewidth]{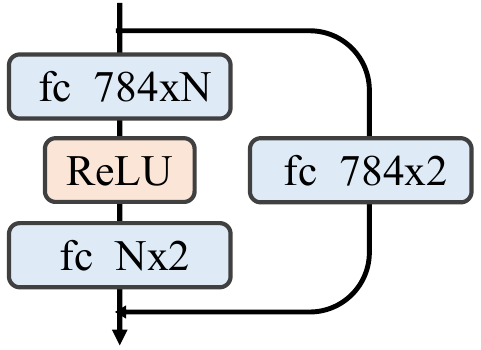}
    \label{fig:mse_extra}
    }
    \subfloat[synthesis transform]{ 
    \includegraphics[width=0.45\linewidth]{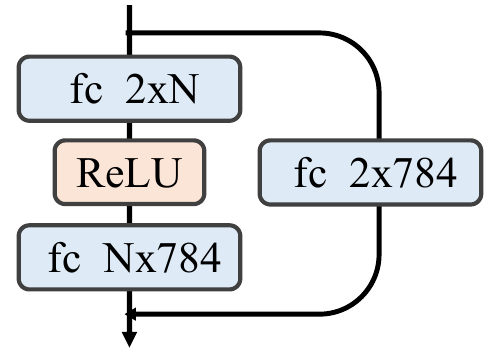}
    \label{fig:func_extra}
    }
    \caption{
    Structure of the analysis transform and synthesis transform.
    }
    \label{fig:extra}
\end{figure}
\section{\large Experiment Setting in Sec. 4.5.}
\begin{figure}[ht]

    \centering
    \subfloat[$g_a$: res]{ 
    \includegraphics[width=0.95\linewidth]{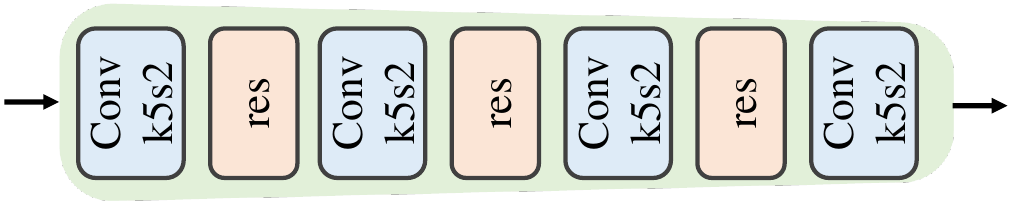}
    \label{fig:mse_extra}
    }

    \subfloat[$g_a$: res+attn]{ 
    \includegraphics[width=0.95\linewidth]{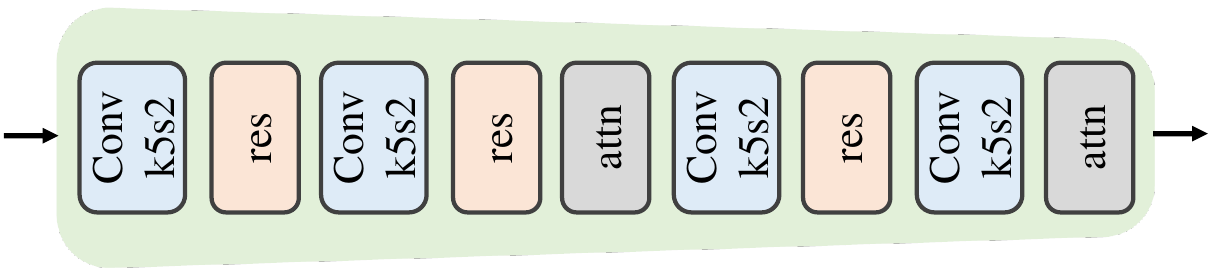}
    \label{fig:func_extra}
    }

    \subfloat[$g_s$: res]{ 
    \includegraphics[width=0.95\linewidth]{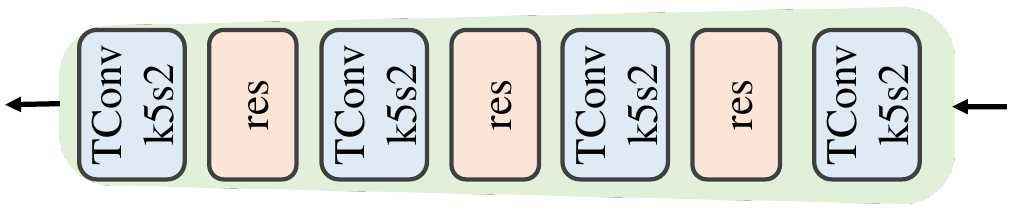}
    \label{fig:mse_extra}
    }

    \subfloat[$g_s$: res+attn]{ 
    \includegraphics[width=0.95\linewidth]{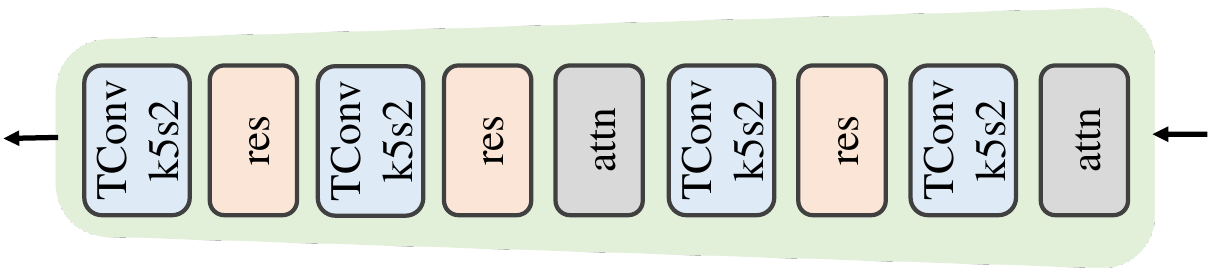}
    \label{fig:func_extra}
    }

    \caption{
    Structure of the analysis transform and synthesis transform.
    }
    \label{fig:tab4}

\end{figure}

The structure of the analysis and the synthesis transform with various settings is shown in Figure \ref{fig:tab4}. For 'res' and 'attn', we refer to the implementation of 
ResidualBlock and AttentionBlock provided py CompressAI \cite{begaint2020compressai}.

\section{\large Derivation in Sec. 4.6.2.}
\begin{figure}[ht]
  \centering
\includegraphics[width=0.6\linewidth]{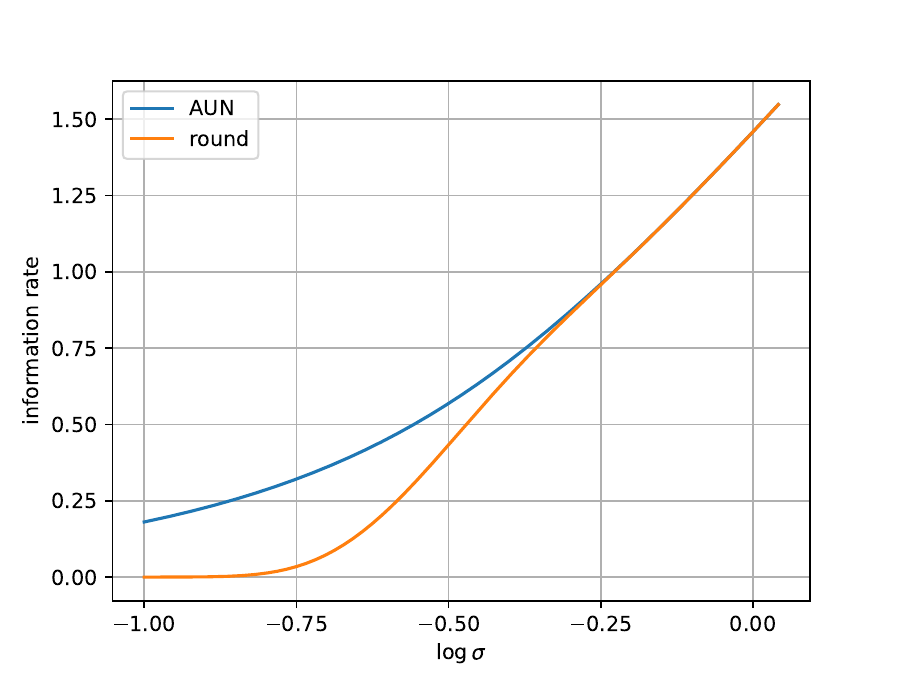}

\caption{
Comparison between $H(\tilde{Y})$ and $H(\lfloor Y -\mu\rceil)$.
}
\label{fig}
\end{figure}

For adding uniform noise, 
\begin{equation}\label{eq_univesalsoft}
\begin{aligned}
&R(\tilde{Y}) = \mathbb{E}_{p_Y,p_U} [-\log q_{\tilde{Y}}(y+u)],\\
    q_{\tilde{Y}}(y+u)&=\Phi(\frac{y+u-\mu+0.5}{\sigma_q})-\Phi(\frac{y+u-\mu-0.5}{\sigma_q})\\
&R(\tilde{Y}-\mu) = \mathbb{E}_{p_Y,p_U} [-\log q_{\tilde{Y}-\mu}(y-\mu+u)],\\
    q_{\tilde{Y}-\mu}(y-\mu+u)&=\Phi(\frac{y+u-\mu+0.5}{\sigma_q})-\Phi(\frac{y+u-\mu-0.5}{\sigma_q})
\end{aligned}
\end{equation}
Therefore, we have $R(\tilde{Y})=R(\tilde{Y}-\mu)$. Then, we consider $H(\tilde{Y})$ and $H(\lfloor Y -\mu\rceil)$. 
\begin{equation}\label{eq_univesalsoft}
\begin{aligned}
p_{\tilde{y}}(\tilde{y})&=\int_{-\infty}^{\infty}p_{u}(u)p_y(\tilde{y}-u)du,\\
&=\int_{-0.5}^{0.5}p_y(\tilde{y}-u)du,\\
&=\int_{-0.5}^{0.5}\frac{1}{\sqrt{2\pi}\sigma}\exp{-\frac{(\tilde{y}-u-\mu)^2}{2\sigma^2}}du,\\
&=\Phi(\frac{\tilde{y}-\mu+0.5}{\sigma})-\Phi(\frac{\tilde{y}-\mu-0.5}{\sigma}).\\
H(\tilde{Y})&=-\int_{\tilde{y}}p_{\tilde{y}}(\tilde{y})\log{p_{\tilde{y}}(\tilde{y})}
\end{aligned}
\end{equation}
\begin{equation}
\begin{aligned}
&P_{\lfloor Y -\mu\rceil}(\lfloor y -\mu\rceil)=\Phi(\frac{\lfloor y -\mu\rceil+0.5}{\sigma})-\Phi(\frac{\lfloor y -\mu\rceil-0.5}{\sigma}).\\
&H(\lfloor Y -\mu\rceil)=-\sum_{\lfloor y -\mu\rceil}P_{\lfloor Y -\mu\rceil}(\lfloor y -\mu\rceil)\log{P_{\lfloor Y -\mu\rceil}(\lfloor y -\mu\rceil)}
\end{aligned}
\end{equation}
It is not easy to compare $H(\tilde{Y})$ and $H(\lfloor Y -\mu\rceil)$ directly from their analytical expressions. Thus, we provide the numerical forms of their magnitudes for comparison. As shown in Figure \ref{fig}, $H(\tilde{Y})\geq H(\lfloor Y -\mu\rceil)$. Therefore, we have $R(\tilde{Y}-\mu)=R(\tilde{Y})\geq H(\tilde{Y})\geq H(\lfloor Y-\mu\rceil)=I(Y-\mu;\lfloor Y-\mu\rceil)$.

\section{\large RD Curves of Tested Models in Sec. 6.2.}
\begin{figure}[ht]
    \vspace{-0.35cm}
    \centering
    \subfloat{ 
    \includegraphics[width=0.8\linewidth]{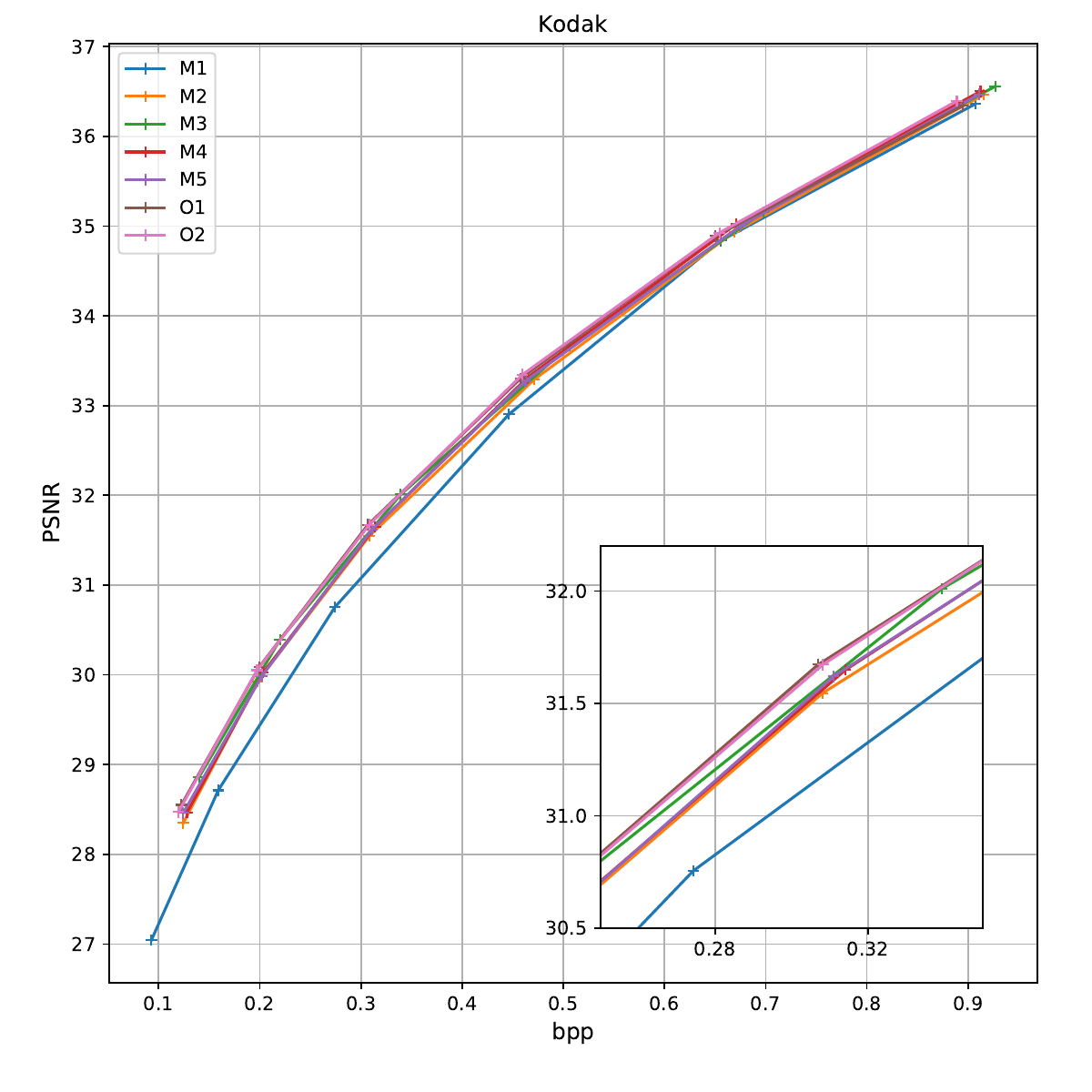}
    \label{fig:mse_extra}
    }

    \vspace{-0.8cm}
    \subfloat{ 
    \includegraphics[width=0.8\linewidth]{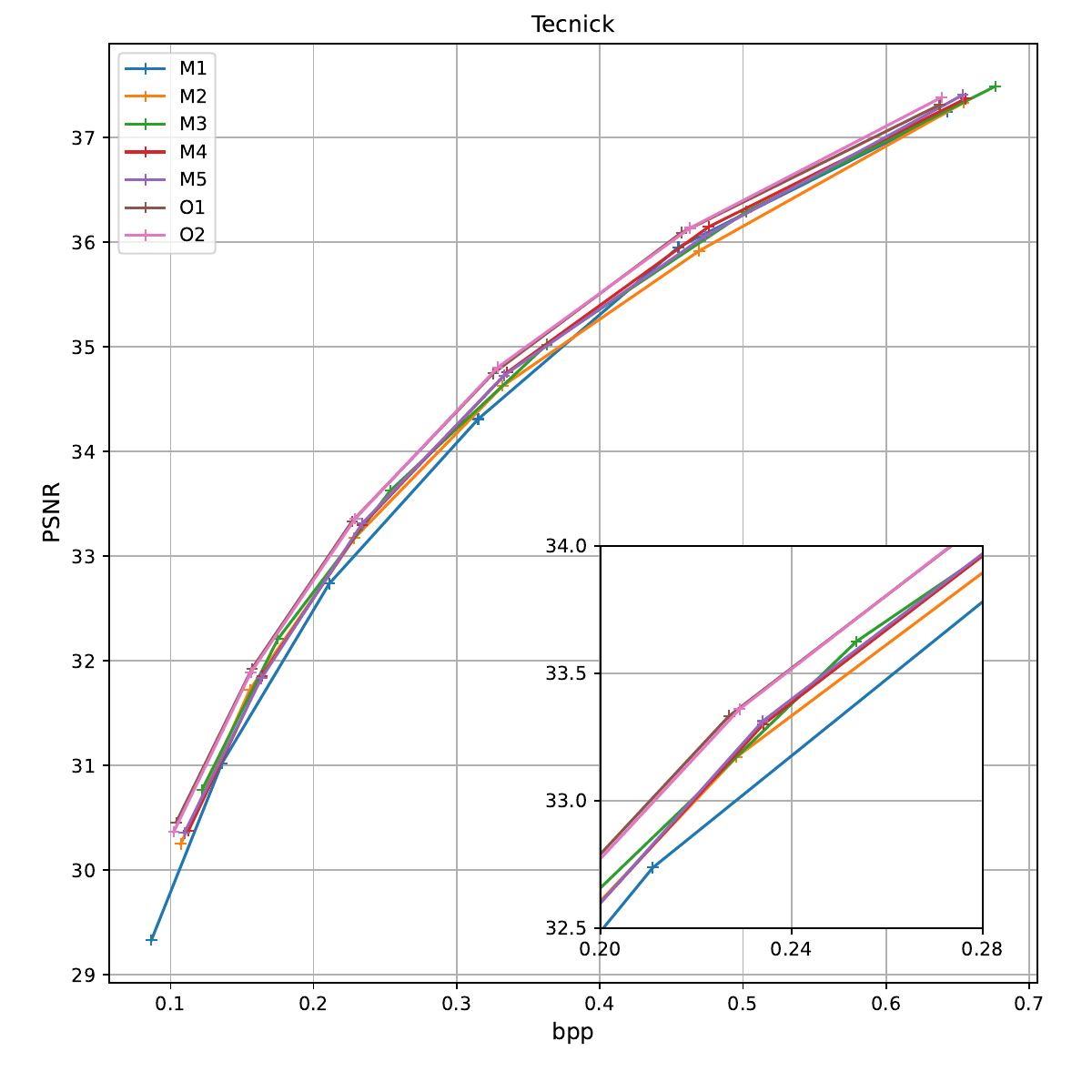}
    \label{fig:func_extra}
    }
    \caption{
    MS-context.
    }
    \label{fig:MS-context}
\end{figure}

\begin{figure}[ht]
    \vspace{-0.35cm}
    \centering
    \subfloat{ 
    \includegraphics[width=0.8\linewidth]{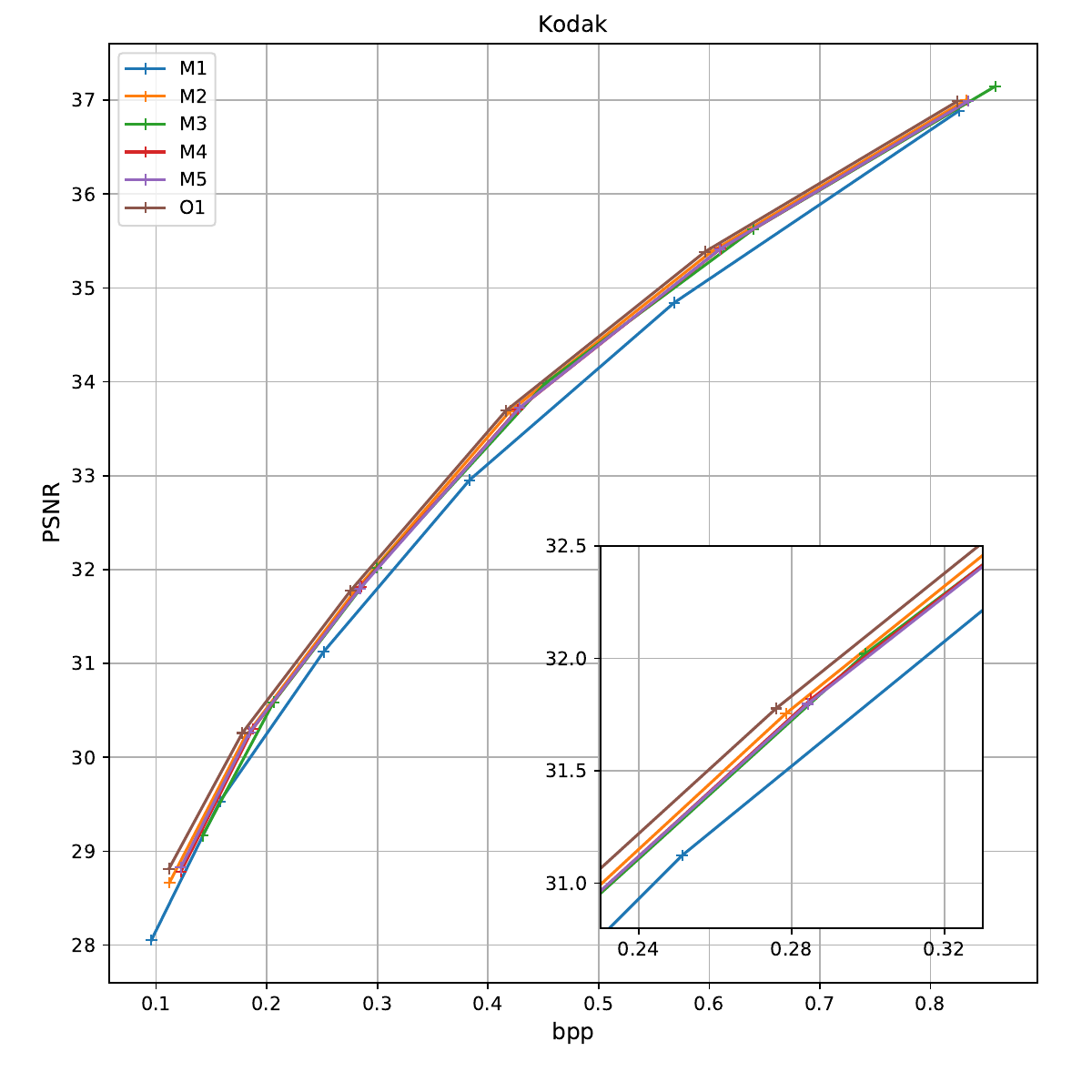}
    \label{fig:mse_extra}
    }

    \vspace{-0.8cm}
    \subfloat{ 
    \includegraphics[width=0.8\linewidth]{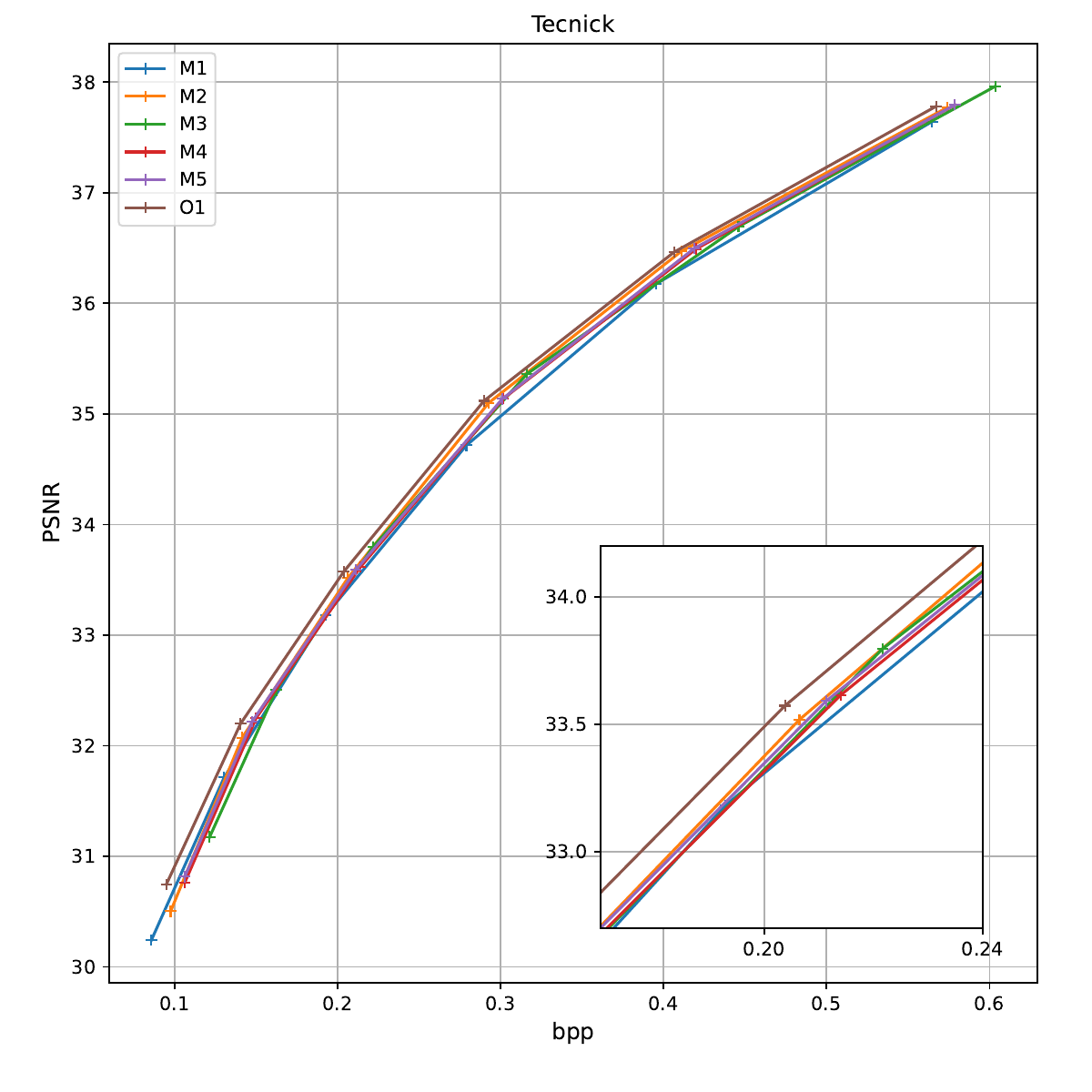}
    \label{fig:func_extra}
    }
    \caption{
    Cheng-anchor.
    }
    \label{fig:Cheng-anchor}
\end{figure}

\begin{figure}[ht]
    \vspace{-0.35cm}
    \centering
    \subfloat{ 
    \includegraphics[width=0.8\linewidth]{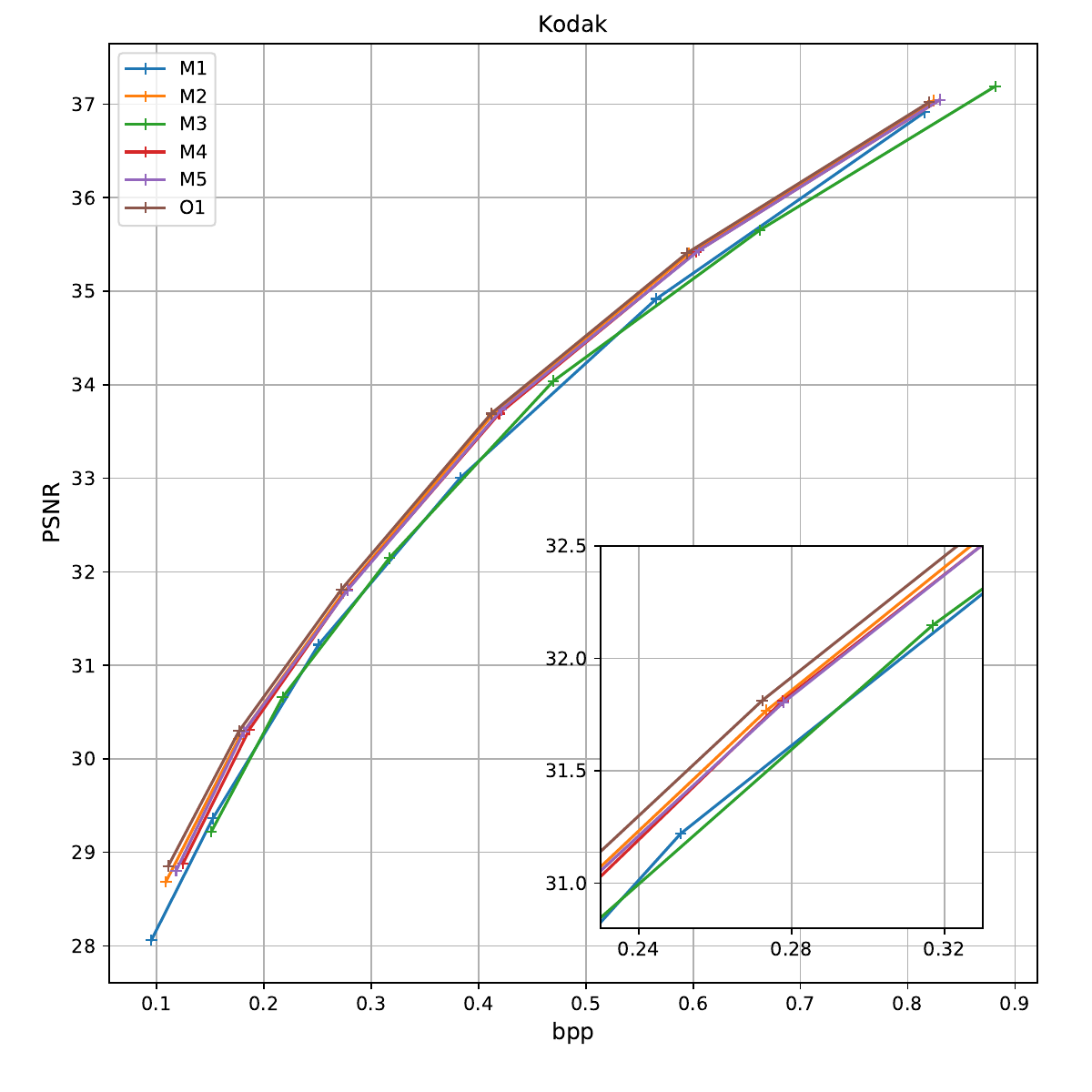}
    \label{fig:mse_extra}
    }

    \vspace{-0.8cm}
    \subfloat{ 
    \includegraphics[width=0.8\linewidth]{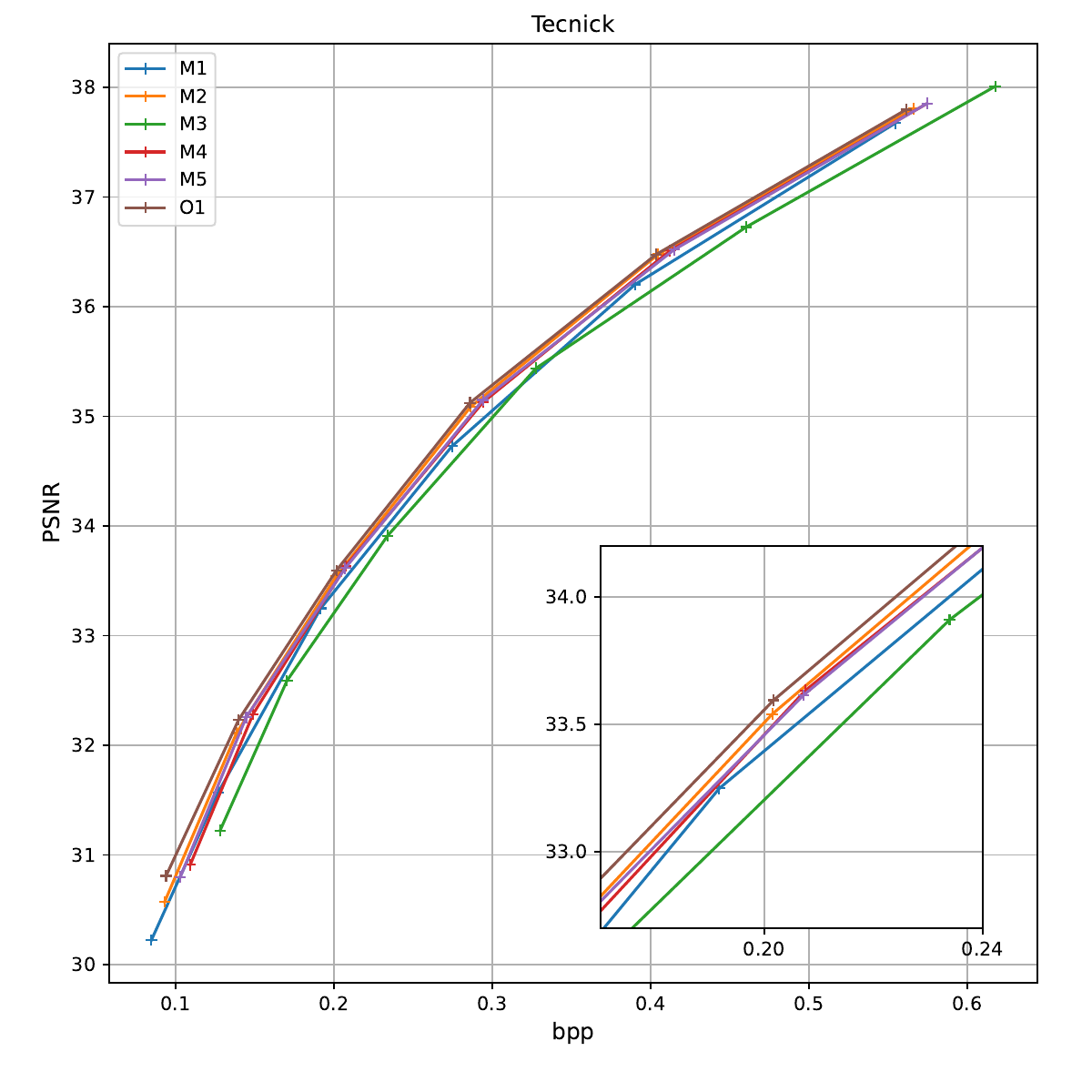}
    \label{fig:func_extra}
    }
    \caption{
    Cheng-attn.
    }
    \label{fig:Cheng-attn}
\end{figure}

\begin{figure}[ht]
    \vspace{-0.35cm}
    \centering
    \subfloat{ 
    \includegraphics[width=0.8\linewidth]{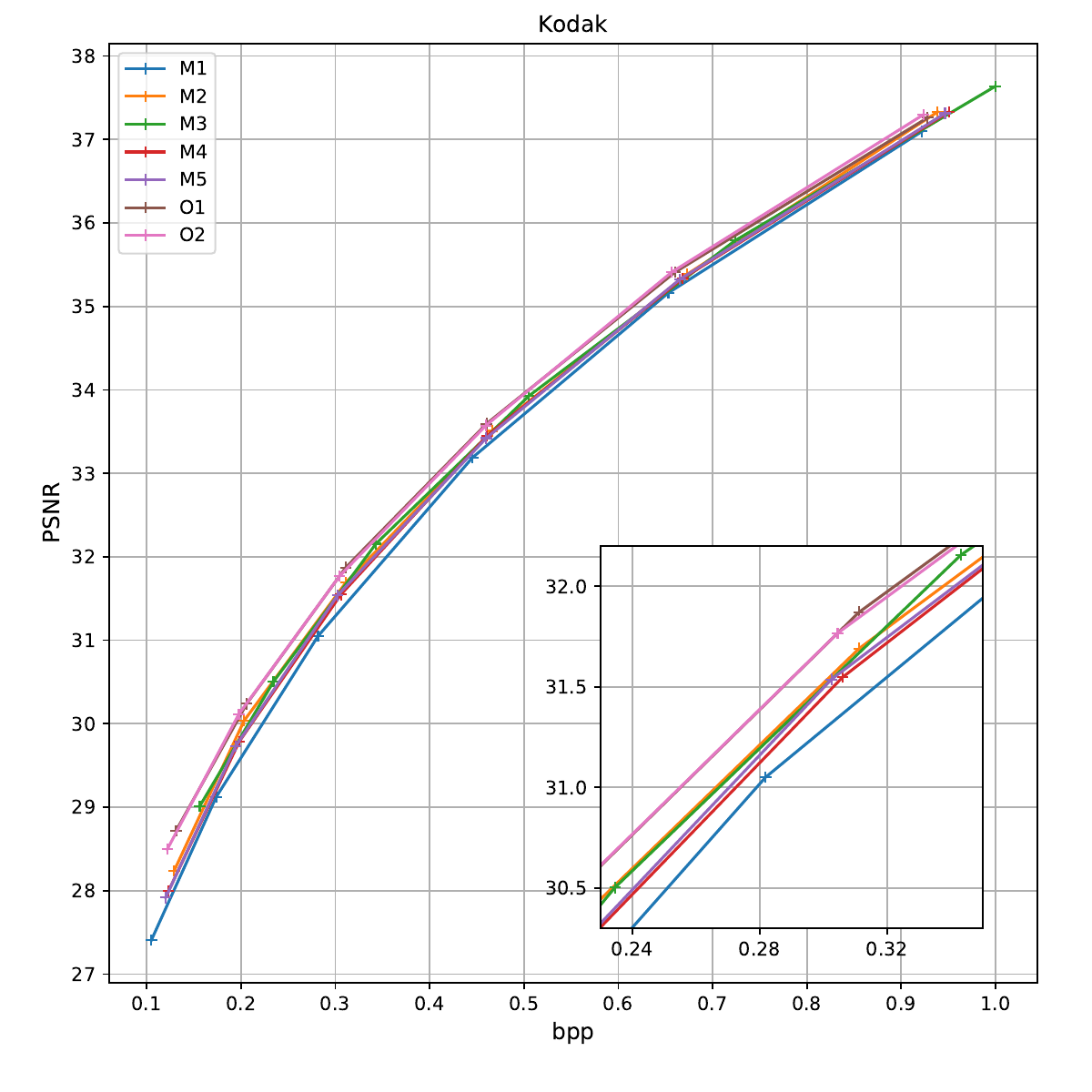}
    \label{fig:mse_extra}
    }

    \vspace{-0.8cm}
    \subfloat{ 
    \includegraphics[width=0.8\linewidth]{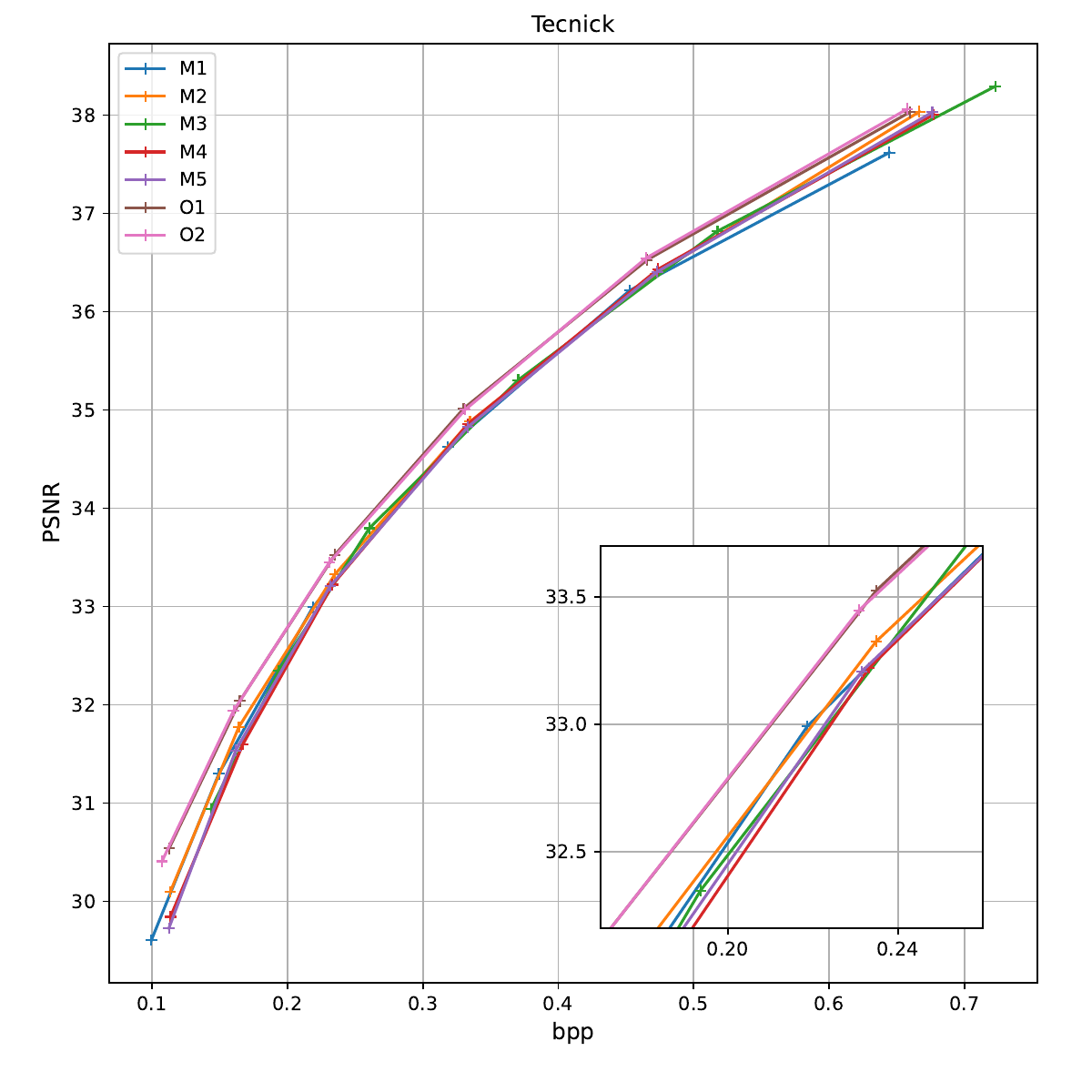}
    \label{fig:func_extra}
    }
    \caption{
    Charm.
    }
    \label{fig:Charm}
\end{figure}

\begin{figure}[ht]
    \vspace{-0.35cm}
    \centering
    \subfloat{ 
    \includegraphics[width=0.8\linewidth]{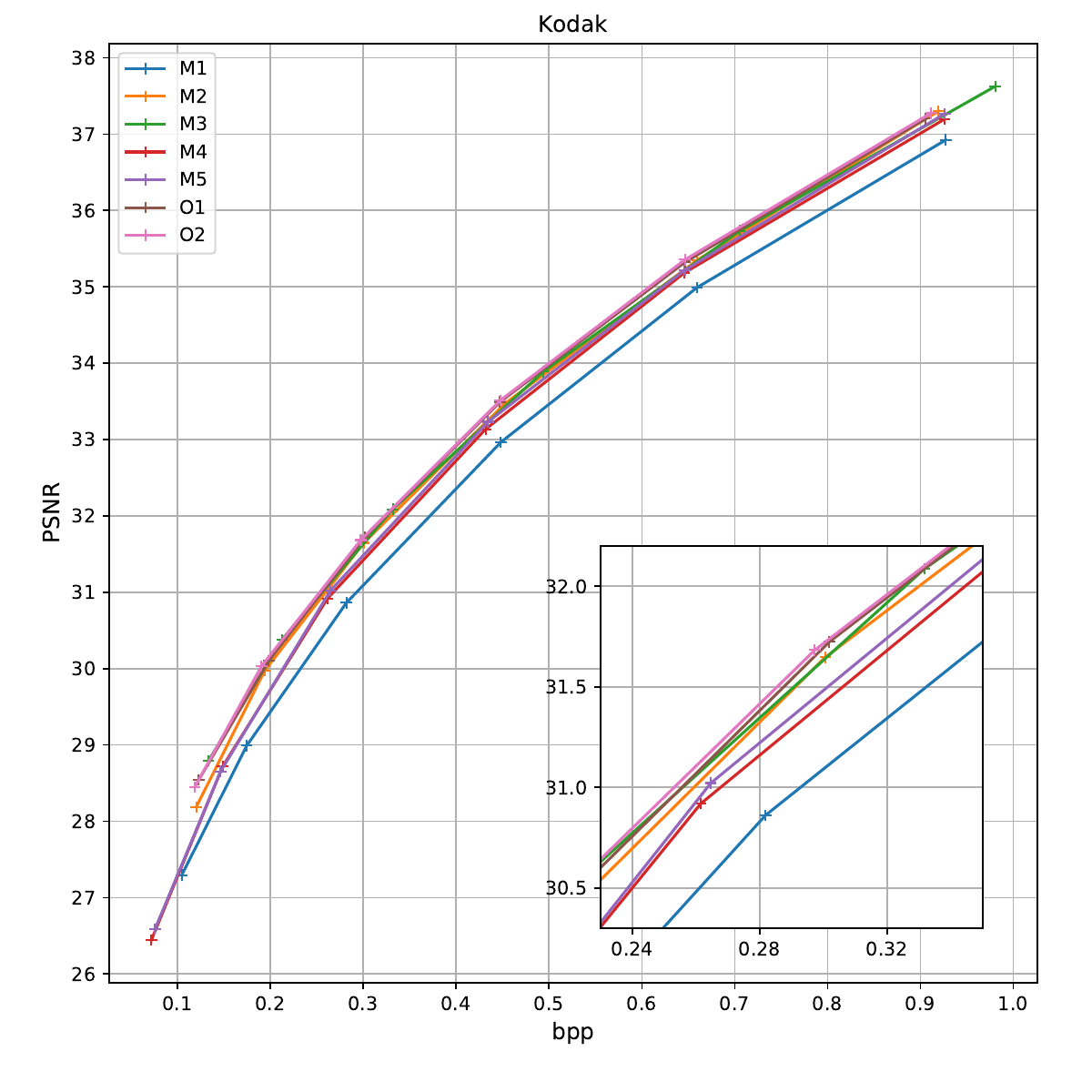}
    \label{fig:mse_extra}
    }

    \vspace{-0.8cm}
    \subfloat{ 
    \includegraphics[width=0.8\linewidth]{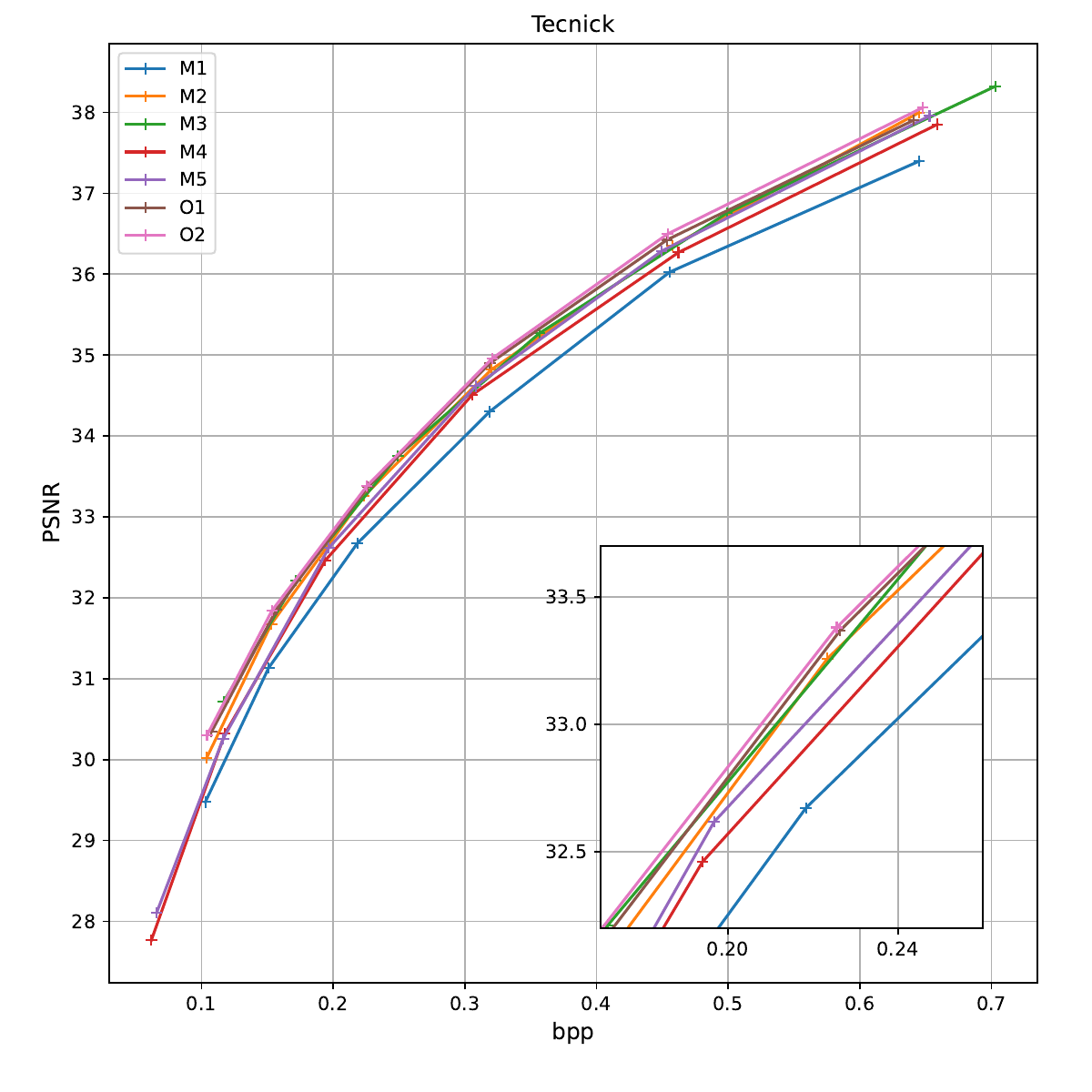}
    \label{fig:func_extra}
    }
    \caption{
    Charm-zero.
    }
    \label{fig:Charm-zero}
\end{figure}

\begin{figure}[ht]
    \vspace{-0.35cm}
    \centering
    \subfloat{ 
    \includegraphics[width=0.8\linewidth]{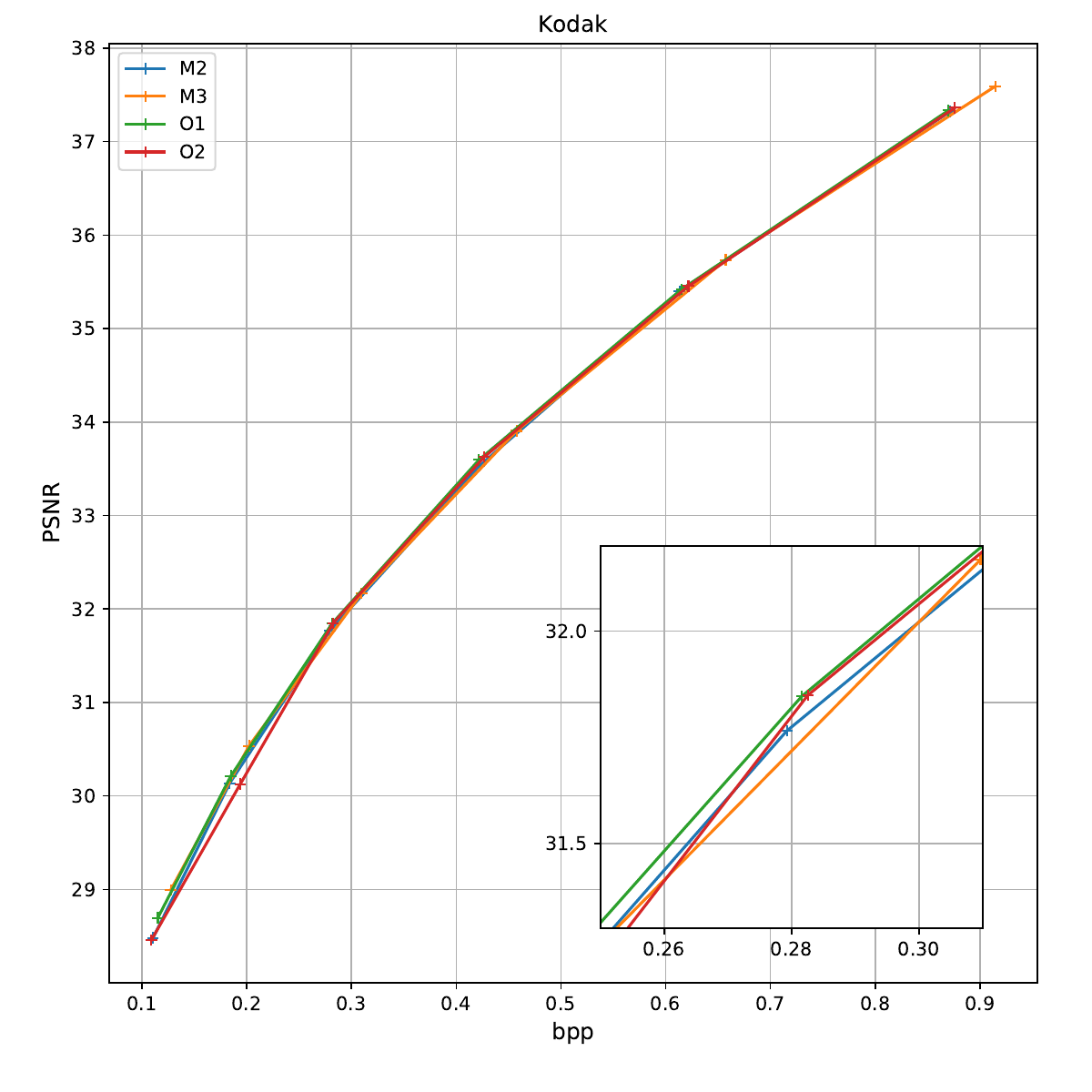}
    \label{fig:mse_extra}
    }

    \vspace{-0.8cm}
    \subfloat{ 
    \includegraphics[width=0.8\linewidth]{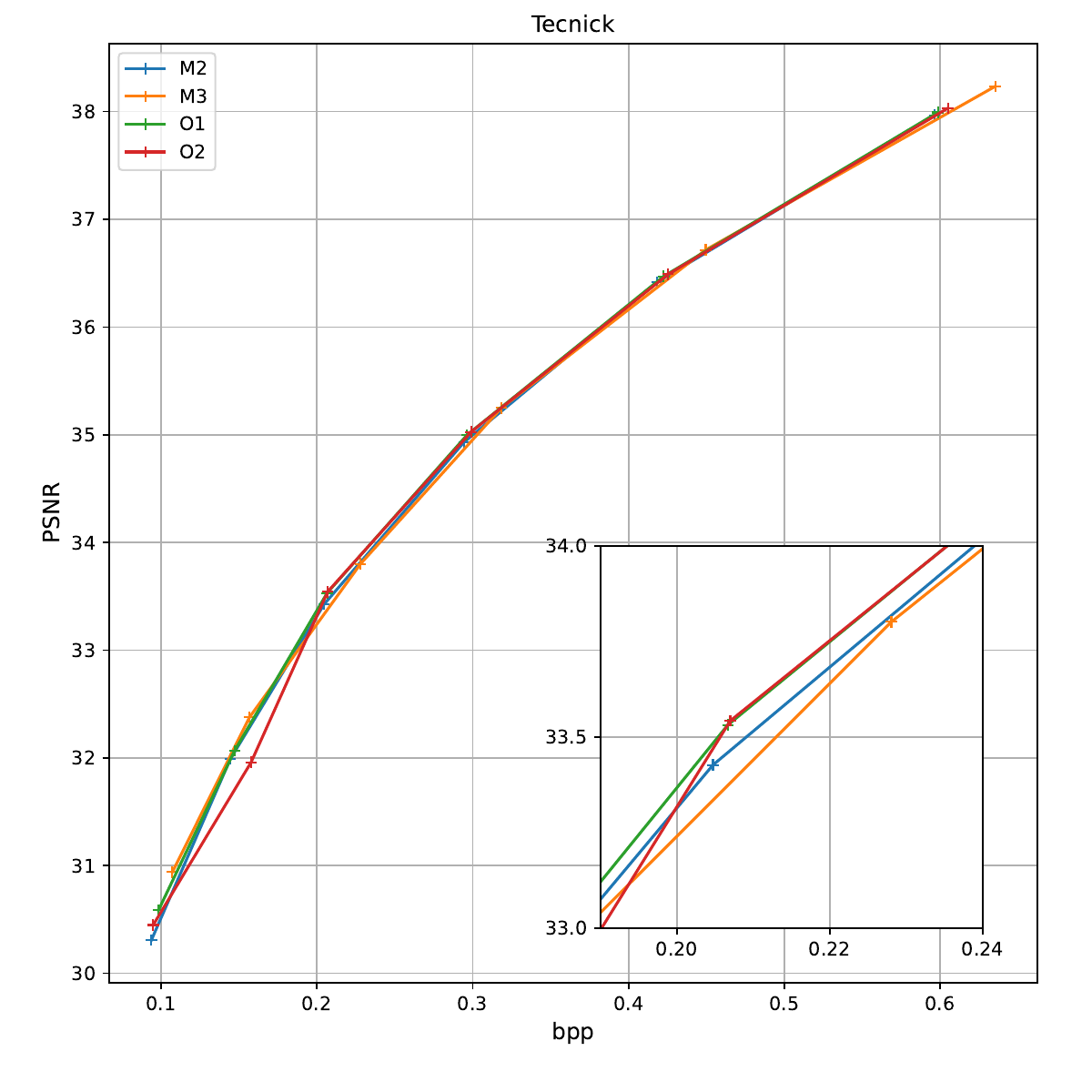}
    \label{fig:func_extra}
    }
    \caption{
    ELIC-sm-zero.
    }
    \label{fig:ELIC-sm-zero}
\end{figure}

\begin{figure}[ht]
    \vspace{-0.35cm}
    \centering
    \subfloat{ 
    \includegraphics[width=0.8\linewidth]{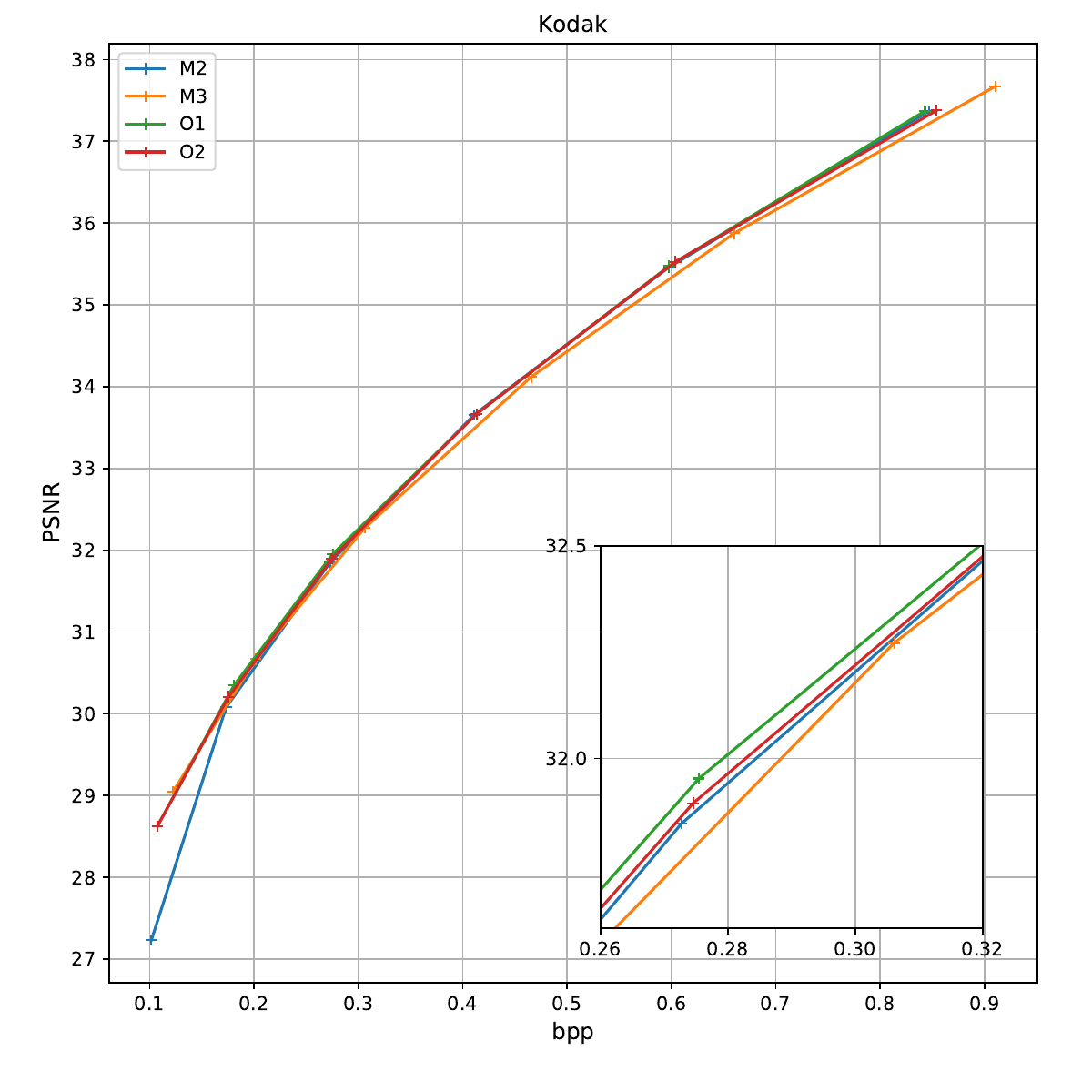}
    \label{fig:mse_extra}
    }

    \vspace{-0.8cm}
    \subfloat{ 
    \includegraphics[width=0.8\linewidth]{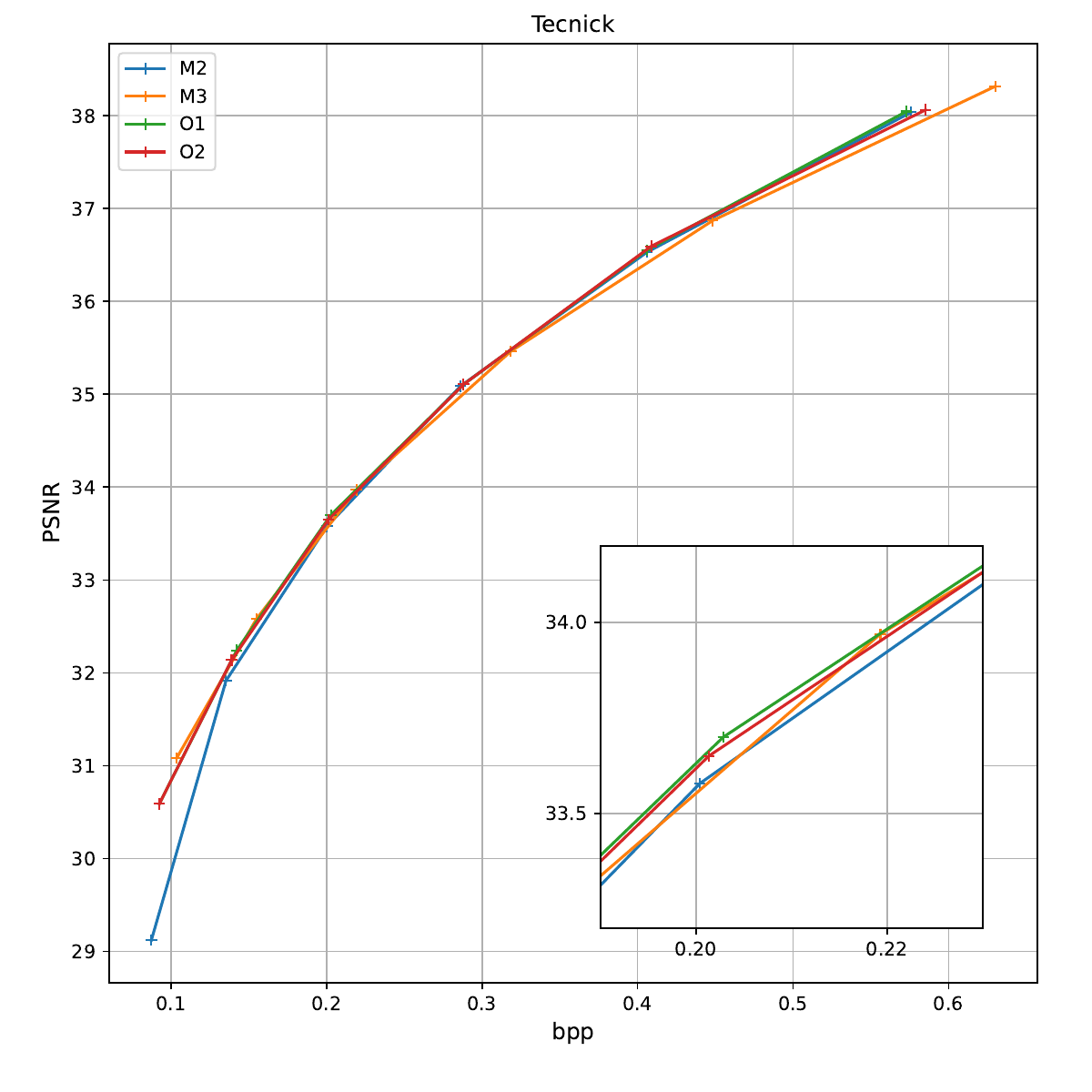}
    \label{fig:func_extra}
    }
    \caption{
    ELIC-zero.
    }
\label{fig:ELIC-zero}
\end{figure}

We show the rate-distortion (RD) curves of MS-context, Cheng-anchor, Cheng-attn, Charm, Charm-zero, ELIC-sm-zero, and ELIC-zero, respectively in Figure (\ref{fig:MS-context}, \ref{fig:Cheng-anchor}, \ref{fig:Cheng-attn}, \ref{fig:Charm}, \ref{fig:Charm-zero}, \ref{fig:ELIC-sm-zero}, and \ref{fig:ELIC-zero}). The results are obtained after post-training. The methods applied during joint-training are shown in Table \ref{tab:suas}.

\begin{table}[ht]
    \centering
    \caption{Explanations of notations.}
    \begin{tabular}{ c|c|c|c } 
    \hline
    Notation&$\sigma_0$&rate& reconstruction\\
    \hline
    M1&1e-6&AUN+PGE&AUN+PGE\\
    M2&0.11&AUN+PGE&AUN+PGE\\
    M3&0.11&AUN+PGE&round+STE\\
    M4&0.11&AUN+PGE&UQ-s+STE\\
    M5&0.11&UQ-s+STE&UQ-s+STE\\
    O1&0.11&SUA+EP&SUA+PGE\\
    O2&0.11&SUA+EP&SUA+STE\\
    \hline
    \end{tabular}
    \label{tab:suas}
\end{table}  


\bibliographystyle{IEEEtran}
\bibliography{supp}